\documentclass[%
 reprint,
 amsmath,amssymb,
 aps,
]{revtex4-2}

\usepackage{graphicx}
\usepackage{dcolumn}
\usepackage{bm}
\usepackage{hyperref}       
\usepackage{url}            
\usepackage{booktabs}       
\usepackage{amsfonts}       
\usepackage{nicefrac}       
\usepackage{microtype}      
\usepackage{xcolor}         
\usepackage{graphicx}
\usepackage{subcaption}
\usepackage{todonotes}

\begin{document}

\preprint{APS/123-QED}

\title{Variational Integrator Graph Networks for Learning Energy Conserving Dynamical Systems}

\author{Shaan A. Desai}
 \altaffiliation[Also at ]{John A. Paulson School of Engineering and Applied Sciences, Harvard University}
 \email{shaan@robots.ox.ac.uk}
\author{Stephen J. Roberts}%
 
\affiliation{%
 Machine Learning Research Group, University of Oxford \\
 Eagle House, Oxford OX26ED, United Kingdom
}%


\author{Marios Mattheakis}
\affiliation{
John A. Paulson School of Engineering and Applied Sciences, Harvard University \\
Cambridge, Massachusetts 02138, United States
}

\date{\today}

\begin{abstract}
Recent advances show that neural networks embedded with physics-informed priors significantly outperform vanilla neural networks in learning and predicting the long term dynamics of complex physical systems from noisy data.  Despite this success, there has only been a limited study on how to optimally combine physics priors to improve predictive performance. To tackle this problem we unpack and generalize recent innovations into individual inductive bias segments.  As such, we are able to systematically investigate all possible combinations of inductive biases of which existing methods are a natural subset.   Using this framework we introduce \emph{Variational Integrator Graph Networks} - a novel method that unifies the strengths of existing approaches by combining an energy constraint, high-order symplectic variational integrators, and graph neural networks.   We demonstrate, across an extensive ablation,  that the proposed unifying framework outperforms existing methods, for data-efficient learning and in predictive accuracy, across both single and many-body problems studied in recent literature.  We empirically show that the improvements arise because high order variational integrators combined with a potential energy constraint induce coupled learning of generalized position and momentum updates  which can be formalized via the Partitioned Runge-Kutta method.


 \end{abstract}

\maketitle

\section{Introduction \label{sec:intro}}
Accurately and efficiently learning the time evolution of energy conserving dynamical systems from limited, noisy data is a crucial challenge in numerous domains including robotics \cite{lutter2018deep}, spatiotemporal dynamical systems \cite{PLA_marios2020}, interacting particle systems \cite{li_conformation-guided_2021}, and materials \cite{zhai_inferring_2021}. To address this challenge, researchers have shown that enriching neural networks with well-chosen inductive biases such as Hamiltonians \cite{greydanus_hamiltonian_2019}, integrators \cite{saemundsson_variational_2019,chang_reversible_2017, chen_neural_2018} and graphs \cite{battaglia_relational_2018, sanchez-gonzalez_graph_2018, sanchez-gonzalez_learning_2020} can significantly improve the learning of complex dynamical systems over vanilla neural networks. Fundamentally, physics-informed learning biases constrain neural networks to uncover and preserve the underlying physical process of a system under investigation. Most methods in this space typically combine multiple individual inductive biases to improve overall predictive performance. However, no study extensively quantifies the performance uplift induced by an individual bias within these combinations. In addition, it remains an open challenge to identify the best combination. 

In this paper, we unpack recent innovations by grouping their inductive biases into generalized segments. We then systematically investigate all possible combinations of these biases. In doing so, existing methods are naturally explored and generalized as they form a subset of the entire ablation. Using this we identify and develop \emph{Variational Integrator Graph Networks} (VIGNs) - a novel method that brings together the core benefits of multiple inductive biases and unifies existing approaches which bring integrative, symplectic and structural form to modeling energy conserving physical systems. We show that higher order variational integrators, formalized via the Partioned Runge-Kutta (PRK) method, can be used to couple position and momentum updates for more precise learning. To benchmark our method we conduct an extensive ablation study across recent developments in physics-informed learning biases and show that VIGNs consistently outperform existing baselines including Hamiltonian Graph Networks (HOGNs) \cite{sanchez-gonzalez_hamiltonian_2019}, ODE Graph Networks (OGNs) \cite{sanchez-gonzalez_hamiltonian_2019}, Hamiltonian Neural Networks (HNNs) \cite{greydanus_hamiltonian_2019}, and Variational Integrator Networks (VINs) \cite{saemundsson_variational_2019} across energy conserved noisy many-body dynamical systems. 

In section~\ref{sec:bgnd} we describe the individual learning biases that comprise VIGNs. We then outline the details of the proposed architecture in section~\ref{sec:mthd}. In Section~\ref{sec:expt} we demonstrate the performance of VIGNs across numerous well known energy conserving physical systems such as the simple pendulum and the many-body interacting spring particle system. Finally, in section~\ref{sec:disc} we unpack the performance uplift and highlight some of the limitations of the existing method.
\section{Background\label{sec:bgnd}}

Numerous recent approaches tackle learning from physical data, but of them three methods stand out; Graph Networks \cite{sanchez-gonzalez_graph_2018}, Hamiltonian Neural Networks \cite{greydanus_hamiltonian_2019} and networks with embedded integrators \cite{chen_neural_2018,saemundsson_variational_2019}. VIGNs allow us to combine the major strengths of each approach and hence form a simple, unifying framework for learning the temporal behaviour of dynamical systems. We briefly review the methods in the following sections.

\subsection{Graph Neural Networks}

The state of a physical system can be represented by a graph $G = (u,V,E)$ \cite{battaglia_relational_2018}. For example, a node ($V$) can be a particle in a many-body problem. These nodes can be used to represent the core features of the particle like its position, momentum, mass, and other particle constants. Edges ($E$) can represent forces between the particles, and the `Globals' ($u$) can represent universal constants such as air density, the gravitational constant etc. In representing physical systems this way, we are able to preserve the structure of our data and find solutions that conform to this prior structure using graph neural networks\cite{battaglia_interaction_2016, battaglia_relational_2018, sanchez-gonzalez_graph_2018,seo_differentiable_2019,cranmer_learning_2019, seo_physics-aware_2020, sanchez-gonzalez_learning_2020,lamb_graph_2020,cranmer_lagrangian_2020}.
Graph neural networks carry out a sequence of transformations to the graph nodes and edges to update the graph parameters. The representation is therefore powerful for many-body systems primarily because the graph networks can operate within the known constraints of physical systems. 

\subsection{Hamiltonian Neural Networks}

In designing a neural network, the typical operation of interest for many physical systems is one which accurately models the time evolution of the system. Recently, the work of \cite{greydanus_hamiltonian_2019} demonstrated that predictions through time can be improved using Hamiltonian Neural Networks (HNNs) which endow models with a Hamiltonian constraint. Given a system with $N$ particles, the Hamiltonian $\mathcal{H}$ is a scalar function of canonical position $\mathbf{q} = (q_1,q_2,....,q_N)$ and momentum $\mathbf{p} = (p_1,p_2,....,p_N)$. In representing physical systems with a Hamiltonian, one can simply use Hamilton's equations to extract the time derivatives of the inputs by differentiating the Hamiltonian with respect to its variables as:
\begin{equation}
\dot{\mathbf{q}} = \frac{\partial \mathcal{H}}{\partial \mathbf{p}}, ~~~
\dot{\mathbf{p}} = -\frac{\partial \mathcal{H}}{\partial \mathbf{q}},
\label{eqn.hamiltonian}
\end{equation}

\noindent where $\dot{a}=\frac{da}{dt}$ $\forall a(t)$. As a consequence, it is noted in \cite{greydanus_hamiltonian_2019} that by training a network to learn $\mathcal{H}$ given inputs $[\mathbf{q},\mathbf{p}]$, the system's state-time derivatives can be naturally extracted through auto-differentiation of the predicted Hamiltonian with respect to the inputs. 
The Hamiltonian in most systems represents the total mechanical energy of the system and is therefore a powerful inductive bias that can be utilized to evolve a physical state while maintaining energy conservation.

\subsection{Potential Neural Networks}

Separable Hamiltonians found in many dynamical systems can be written as $\mathcal{H}(\mathbf{q},\mathbf{p}) = E_{\mathrm{kinetic}}(\mathbf{p}) + E_{\mathrm{potential}}(\mathbf{q}).$ Typically, for rigid body systems the form of the kinetic energy is $E_{\mathrm{kinetic}}=\frac{M^{-1}}{2}\mathbf{p}^2$ where $M$ is an inertial mass matrix that connects the generalized momenta $\dot{\mathbf{q}}$ to the canonical momenta $\mathbf{p}$ such that $\mathbf{\dot{q}} =  \frac{\partial E_{\mathrm{kinetic}}}{\partial \mathbf{p}} = M^{-1} \mathbf{p}$. The authors of \cite{yu_onsagernet_2020} and \cite{saemundsson_variational_2019} exploit this simplification when dealing with generalized position $\mathbf{q}$ and velocity $\dot{\mathbf{q}}$ to collapse Eqn.~\ref{eqn.hamiltonian} into:
\begin{equation}
\frac{\mathrm{d}\mathbf{q}}{\mathrm{d}t} = \dot{\mathbf{q}}, ~~~
\frac{\mathrm{d}\mathbf{\dot{q}}}{\mathrm{d}t} = -M^{-1}\frac{\partial E_{\mathrm{potential}}(\mathbf{q})}{\partial \mathbf{q}}.
\label{eqn.hamiltonian1}
\end{equation}

Equation~\ref{eqn.hamiltonian1} allows us to learn a single function $E_{\mathrm{potential}}$ with fewer network weights needed to learn a Hamiltonian and a single backpropagation with respect to $\mathbf{q}$ as opposed to $[\mathbf{q},\mathbf{p}]$ for HNN. Further, it gives us the flexibility to learn the inertial mass matrix by explicitly learning $M_{\theta}$ rather than nesting it in $\tilde{\mathcal{U}}_{\theta} = M^{-1} \mathcal{U}$. As this type of network has not been introduced formally as an individual inductive bias, we coin the term Potential Neural Networks (PNNs) in reference to them.

In the case where we only have canonical coordinates, potential networks can still be used but a separate neural network needs to be designed to learn the inertial matrix $M$ \cite{saemundsson_variational_2019}.

\subsection{Embedded Integrators}

Dynamical systems can be represented by systems of differential equations in the form: 

\begin{equation}
    \dot{\mathbf{s}} = f(\mathbf{s},t),
    \label{eqn.diff}
\end{equation}

\noindent where $\mathbf{s} = (\mathbf{q},\mathbf{p})^T$ is a state vector, $t$ is time, and $f$ is an arbitrary function of time and the state vector. One approach to solving Eqn.~\ref{eqn.diff} is to parametrize the function $f$ by a neural network and minimize the euclidean distance between the predicted state time derivatives $\hat{\dot{\mathbf{s}}}$ and the ground truth $\dot{\mathbf{s}}_{\mathrm{gt}}$ derivatives. One challenge in doing this is it assumes access to the ground truth state time derivatives, which can be hard to extract. To circumvent this problem, researchers embed a numerical integrator into the training process \cite{chen_neural_2018, zhong_symplectic_2019}. Formally this equates to integrating both sides of Eqn.~\ref{eqn.diff} such that:

Short Range Integration:
\begin{equation}
\mathbf{s}_{t+1} = \mathbf{s}_t + \int_t^{t+\Delta t} f_{\theta}(\mathbf{s},t) \mathrm{d}t.
\label{eqn.action_int1}
\end{equation}

Long Range Integration:
\begin{equation}
\mathbf{s}_{t+n} = \mathbf{s}_t + \int_t^{T_{\max}} f_{\theta}(\mathbf{s},t) \mathrm{d}t,
\label{eqn.action_int2}
\end{equation}

\noindent where $\theta$ are the weights of the neural network. The short range integration involves a single discrete step $\Delta t$ whereas the long range integration involves a sequence of discrete steps to the final time $T_{\max} = n\Delta t$. It can be shown that if we integrate the system from $t$ to $T_{\max}$, the network above ends up being a composition of multiple transformations as would be found in recurrent neural networks and residual networks \cite{chen_neural_2018}.


\subsubsection{Symplecticity}

While the embedded integrator resolves the challenge of having to obtain state time derivatives, it introduces a new complexity - the choice of integrator. Numerical integrators are chosen based on their precision and truncation error, however, when dealing with Hamiltonian systems an additional factor to consider is whether the integrator preserves the energy of the system. 
It has been shown that symplectic integrators can preserve the energy during integration making them versatile for long range integrations of conserved quantities \cite{marsden_discrete_2001}. The performance of these integrators on a range of different systems is outlined in the Appendix, where we see that for long range integrations even low order symplectic integrators are as performant as high order Runge-Kutta (RK) methods in terms of energy conservation.

While Variational Integrator Networks (VINs) \cite{saemundsson_variational_2019} and Symplectic Recurrent Neural Networks  \cite{chen_symplectic_2020} both illustrate how an embedded symplectic integrator improves learning over traditional RK methods, they only do so for low order methods. To extend our investigation to higher order symplectic integrators, we need to study Partitioned Runge-Kutta methods.

Typically, RK methods can be described by Butcher tables (see Appendix) and if the coefficients satisfy certain conditions then they can be made symplectic \cite{marsden_discrete_2001}. However, the additional symplecticity constraint on the table of coefficients forces the integration scheme to be implicit. While implicit integrators are powerful, they require a root finding approach. Introducing such complexity into an embedded NN is possible but significantly complicates the backpropagation technique. As such, it is of importance to establish whether explicit symplectic methods can be developed. Fortunately, by creating separate Butcher tables for position and momentum it is indeed possible to describe a Partitioned Runge-Kutta (PRK) method with coefficients that result in an explicit symplectic integration scheme. Note that variational integrators, derived through variational calculus, can be described by PRK methods \cite{marsden_discrete_2001}. As a consequence, it is possible to generalize the result of Variational Integrator Networks (VINs) to higher order methods (see Appendix for details).



\subsection{Related work}
The notion of embedding physically-informed inductive biases in neural networks can be found in numerous early work aimed at modeling materials \cite{witkoskie_neural_2005, pukrittayakamee_simultaneous_2009, smith_ani-1_2017, rupp_fast_2012, yao_tensormol-01_2018}. For example, early efforts by Witkoskie and Doren \cite{witkoskie_neural_2005} demonstrate that in contrast to directly learning a potential energy surface, the inclusion of gradients in the learning process can drive a network to accurately model the forces. However, most materials modeling frameworks are task-specific and usually do not generalize well.

More general approaches that capture physical laws include search algorithms \cite{hills_algorithm_2015}, symbolic learning \cite{cranmer_learning_2019}, as well as regressive techniques \cite{iten_discovering_2018,schmidt_distilling_2009,de_silva_discovery_2019}. In addition, graphs have also been presented as natural inductive biases in modeling physics \cite{battaglia_interaction_2016, battaglia_relational_2018}.

NeuralODE \cite{chen_neural_2018} has also re-sparked an interest in inductive biases for differential equations. Inspired by this work, \cite{greydanus_hamiltonian_2019} and \cite{toth_hamiltonian_2019} show that a neural network can be used to predict a Hamiltonian which can be differentiated with respect to the input ($\bf {p}$ and $\bf {q}$) to obtain the time derivatives of the system. With these derivatives accurately learnt, a NeuralODE-type integration scheme can be used to evolve a system. This general approach has formed the basis for many advancements within physical learning \cite{saemundsson_variational_2019,sanchez-gonzalez_hamiltonian_2019,zhong_symplectic_2019,choudhary_physics_2019,sanchez-gonzalez_learning_2020}.

Although HNNs predict dynamics for few body systems well (e.g. a swinging pendulum or mass spring system) they are not readily adaptable to large N-body problems when the input dimension grows. The work in \cite{sanchez-gonzalez_hamiltonian_2019} shows that graph networks are ideal for resolving this type of system because they can operate on structured data i.e. the system does not need to be vectorized as would be the case for multi-layer feed forward neural networks.

Inspired by NeuralODEs, variational integrator networks \cite{saemundsson_variational_2019} propose a neural network whose architecture matches the discrete equation of motion governing the dynamical system, as derived by applying the Euler-Lagrange equations to a discretized action integral. The paper indicates major benefits when using the method for noisy data, as well as providing precise energy and momentum conservation. 

While it is clear that the constrained HNN \cite{finzi_simplifying_2020} is capable of solving Hamiltonian systems more efficiently, it assumes we have access to Cartesian coordinates for all systems and requires explicit rigid body constraints.

Our method brings together the inductive biases presented in all these papers and leverages them to solve large many-body problems in noisy data settings.

\begin{figure*}
\includegraphics[width = \textwidth]{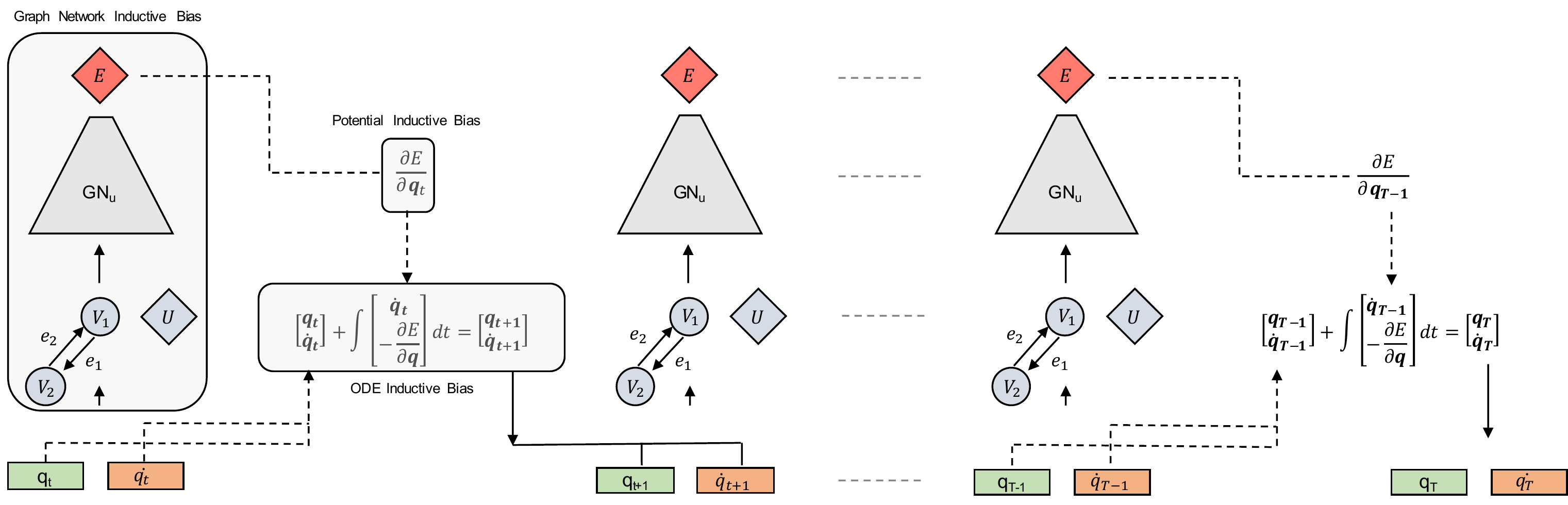}
\caption{The architecture for our method takes as input the position vector $[q]$ and feeds it through a graph network to compute the potential energy $E(q)$. Using backpropagation, the update for $\dot{q}$ is computed and the input state is integrated one step. Continuing this sequence yields an multi-step integration scheme.}
\label{fig.arch}
\end{figure*}

\section{Method\label{sec:mthd}}

The architecture for our method is shown in Fig. \ref{fig.arch}. The network takes as input state vectors $\mathbf{s} = (\mathbf{q})$ representing generalized coordinates and learns to predict the potential function and its derivatives with respect to the inputs. Note, we adopt the potential neural network so we require generalized coordinates. However, the transition to a Hamiltonian NN is straightforward.  Since the input training data can be described by a graph, we show vertices $V_i$, edges $E_{ij}$ and globals $u$ 
as input to the graph network $GN_u$ to predict the potential energy $E_{\mathrm{potential}}$. The key difference between our graph approach and HOGN \cite{sanchez-gonzalez_hamiltonian_2019} is that our network only takes the position $\mathbf{q}$ as input i.e. the nodes only have position data.
In the noiseless setting, the training loss is defined as the mean-squared error (MSE) across all time steps and across all state vectors. In the noisy setting, we follow a similar approach to \cite{saemundsson_variational_2019} and compute the full log-likelihood of the predicted state vector $\mathbf{s}_{\mathrm{pred}}$ as:
\begin{equation}
P(\mathbf{s}_{\mathrm{pred}}|\mathbf{s},\sigma^2) = \prod_{t=1}^{T_{\mathrm{max}}} \mathcal{N}(\mathbf{s}_{\mathrm{pred}}(t)|\mathbf{s}(t),\sigma^2I),
\label{eqn.noisyloss}
\end{equation}
\noindent where $\mathcal{N}$ is a Gaussian distribution, $\sigma^2$ reflects the variance and $I$ is an identity matrix.

To benchmark the performance of our method we conduct an extensive ablation study. Our ablation iterates across all combinations of the inductive biases described in the  section \ref{sec:bgnd}. Namely, it includes both graph and non-graph methods that either learn the state derivatives directly   \cite{greydanus_hamiltonian_2019,sanchez-gonzalez_hamiltonian_2019},
the Hamiltonian (Hamiltonian networks) or the potential function (potential networks). We use 1st through 4th order integrators that are symplectic and non-symplectic. We also iterate over a multi-step integration scheme with step sizes of 1, 5 and 10 to account for both short, mid and long-range integrations during training. Note that we can indeed integrate for more than 10 steps but this increases the memory requirement. In addition, since the ablation iterates over all possible combinations of inductive biases, existing methods in the literature are naturally covered. For example, VINs can be described as low-order, long-range symplectic integrators coupled with potential networks. While HOGNs couple low and high order, short-range integrators with Hamiltonians and graphs. 

Since our work iterates across all these methods we adopt a new naming convention for convenience. We refer to networks that combine graphs with potential networks as Potential Graph Networks (PGNs). Note that VIGNs are PGNs under symplectic integration.

\section{Experiments\label{sec:expt}}

We carry out our experiments on numerous datasets used in recent literature and describe their configurations here (see Appendix for full training/testing configurations). 

\noindent \textit{Training:} For all the systems we investigate the training data is generated using an 8th order Runge-Kutta method with $r_{\mathrm{tol}}=10^{-12}$ so that the ground truth is precise and conserves energy. The noise model for all systems is chosen to maintain a noise-to-signal ratio of less than 30$\%$ which allows us to investigate which architecture is the most robust to noisy data. For all noisy training configurations, the noise source is a Gaussian $\mathcal{N}(0,\sigma)$. The noise is added to each state vector similar to the approach taken in \cite{greydanus_hamiltonian_2019} and \cite{saemundsson_variational_2019}.

\noindent \textit{Testing:} To evaluate the performance of our models, we sample 50 initial conditions and integrate these systems to 3 times the training time horizon $3T_{\max}$. In other words, the true performance of the model is tested by evaluating points beyond the training regime. For each set of initial conditions we compute the MSE across the entire trajectory between the prediction and the ground truth states. Note that some of our 50 sampled initial conditions can be slightly outside the training regime which can lead to a few poorly predicted trajectories by all the models. To prevent this skewing our final reported results, we compute the geometric mean, a measure of central tendency, of the MSEs computed across the 50 initial conditions.

Here, we describe the systems investigated. A list of all the experimental results can be found in the appendix. In the following discussion, we only report the results of our ablation with 4th order methods for clarity. The results for these systems are summarized in Fig.~\ref{fig.f1} and the others are summarized in the appendix.

\begin{figure*}[htb]
\centering
\includegraphics[width=\textwidth]{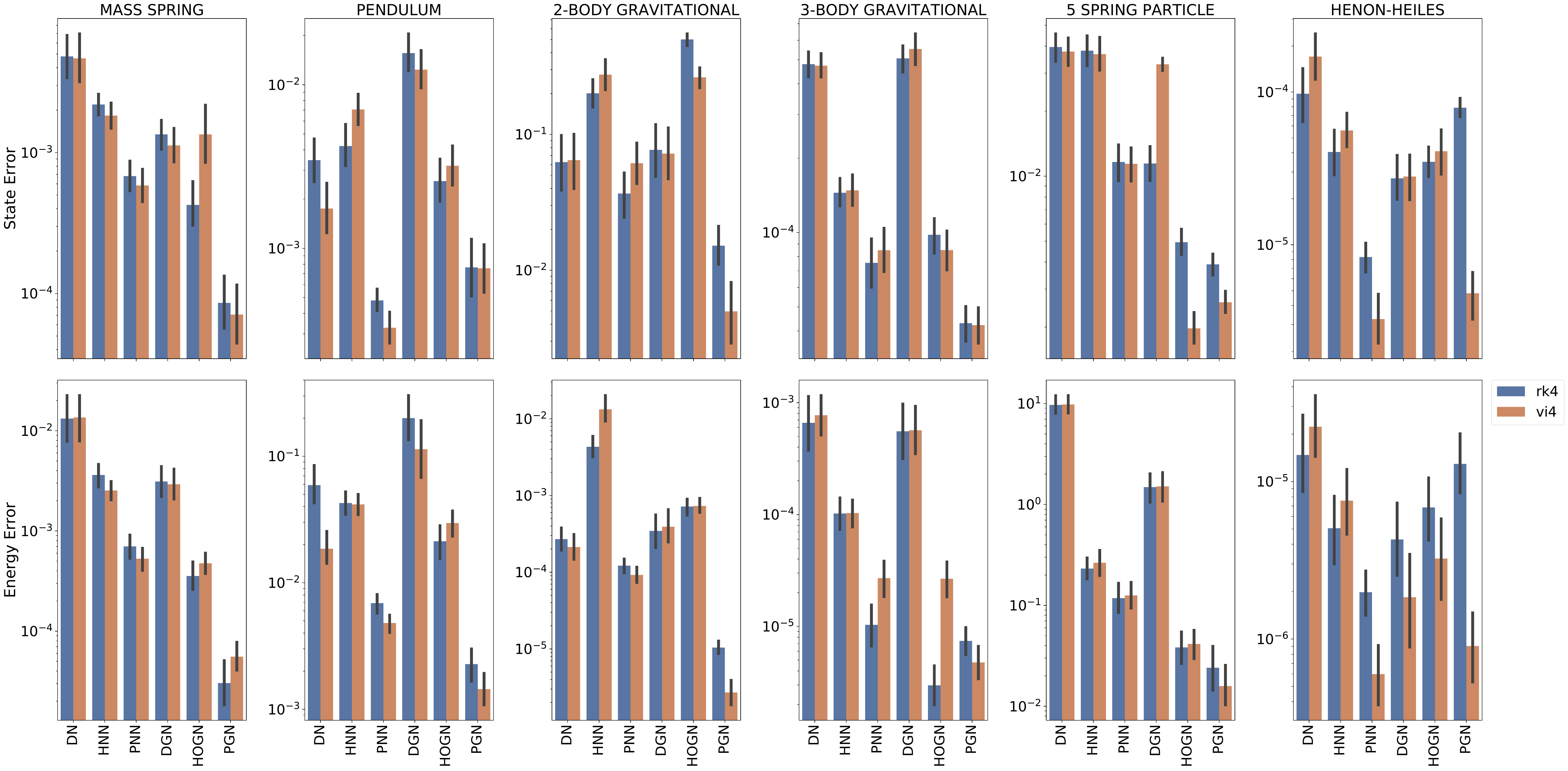}
\caption{State and energy geometric mean (and $\pm  \sigma$ standard errors) of the rollout MSE for 50 initial conditions. The results are reported for models trained on noisy data. We see that high order, long-range integrators coupled with potential networks perform the best with their graph variants showing added versatility in both single and many-body systems.}
\label{fig.f1}
\end{figure*}	

\subsection*{Mass-Spring system}
We start by considering a simple frictionless 1-D mass-spring system modeled by the Hamiltonian as:
\begin{equation}
\mathcal{H} = \frac{p^2}{2m} + k \frac{q^2}{2}.
\label{eqn.mspring}
\end{equation}
\noindent For simplicity we set the mass $m$ and spring constants to $1$ without loss of generality. As is done in \cite{greydanus_hamiltonian_2019}, the training data is sampled uniformly in an energy range of 0.5 to 4.5.

\subsection*{Pendulum system}

We carry out testing on a 1-D pendulum, which is more complex than the simple mass spring because it is a non-linear system. The Hamiltonian is modeled as:
\begin{equation}
\mathcal{H} = \frac{p^2}{2ml^2} + m\mathrm{g}l \left ( 1-\cos(q) \right ),
\label{eqn.pendulum}
\end{equation}
\noindent where the mass and lengths are set to 1, $g$ is set to $9.81$. We use 25 initial conditions which satisfy the condition that the total energy lies in $[1.3,2.3]$ for training. Note that this energy yields strong non-linear behaviour.

\subsection*{2-body gravitational system}

\begin{figure*}[htb]
\centering
\includegraphics[width=0.9\textwidth]{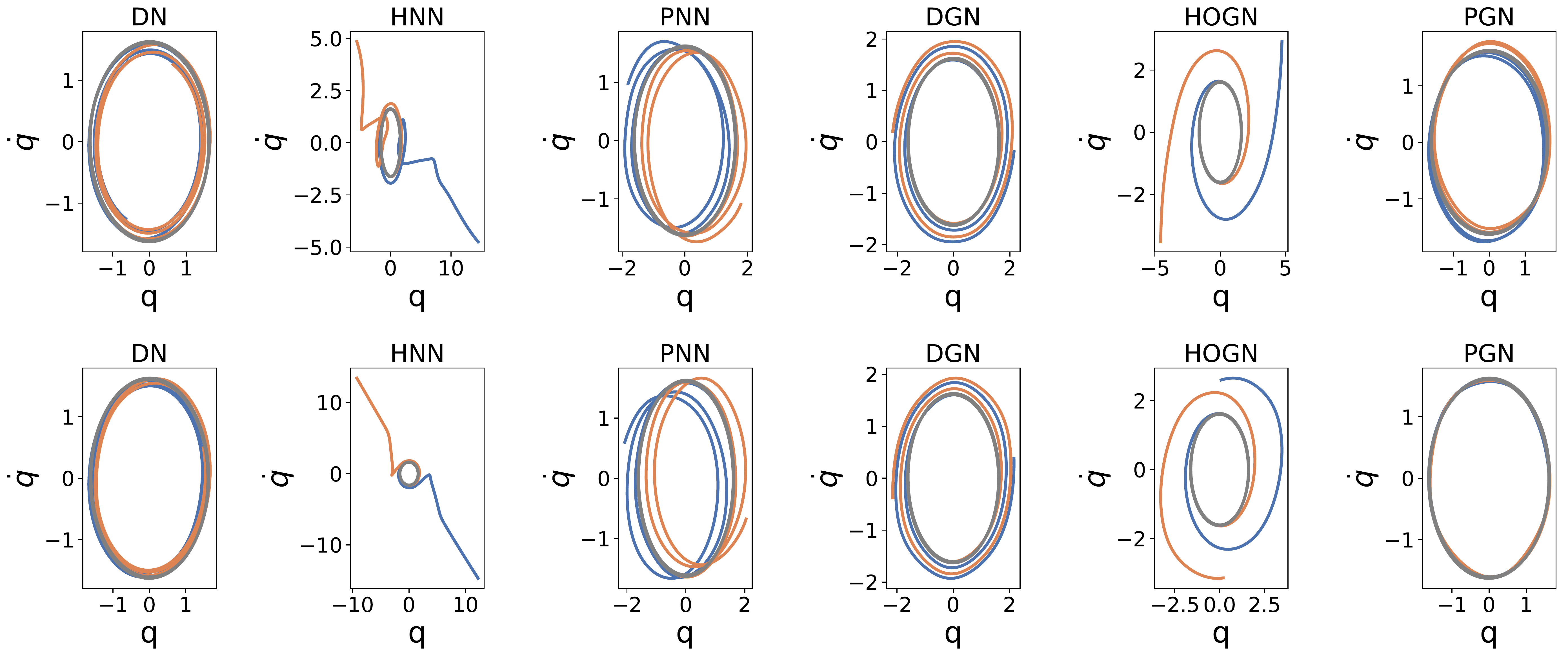}
\caption{Qualitative evolution of a single test state of the 2-body gravitational problem with each model trained on noisy data. The top row shows each method integrated with a RK4 integrator. The bottom is a 4th order symplectic integrator. We see that PGN is the most performant method as it stays close to the ground truth lines (marked in black) with the variational integrator variant of PGN (VIGN) doing the best (bottom right).}
\label{fig.ngravtrial}
\end{figure*}

The 2-body system represents a particle system in which the forces between particles is modelled by a gravitational force. The system can be represented by the Hamiltonian:

\begin{equation}
\mathcal{H} = \sum_{i=1}^{2}\frac{|p_i|^2}{2m_i} - \sum_{1\leq i \leq j \leq 2}\mathrm{g} \frac{m_i m_j}{|q_j-q_i|^2},
\label{eqn.twobody}
\end{equation}
\noindent where we set masses to 1 and $g$ to 1 without loss of generality. The coordinates are assumed to be scaled by the reduced mass $\mu$, in addition, the center of mass is assumed to be fixed at 0. We use 20 initial conditions which satisfy the condition that the radius of a particle's trajectory is uniformly sampled between $[0.5,1.5]$ as is done in \cite{greydanus_hamiltonian_2019}. We visualize the rollout of a single test point in Fig.~\ref{fig.ngravtrial}.

\subsection*{3-body gravitational system}

The 3-body system represents a particle system in which the forces between particles is modeled by a gravitational force. The system can be represented by:  
\begin{equation}
\mathcal{H} = \sum_{i=1}^{3}\frac{|p_i|^2}{2m_i} - \sum_{1\leq i \leq j \leq 3}\mathrm{g} \frac{m_i m_j}{|q_j-q_i|^2},
\label{eqn.twobody}
\end{equation}

\noindent where we set masses to 1 and $g$ to 1.

\subsection*{N-body spring force system}

We also carry out our experiments on a dataset similar to that found in \cite{sanchez-gonzalez_hamiltonian_2019}. We develop a N-body dataset, where the interaction force between particles is modeled by $ \mathbf{F}_{ij} = -k_ik_j(\mathbf{q}_i - \mathbf{q}_j) $ following the same sampling procedure in \cite{sanchez-gonzalez_hamiltonian_2019}, leading to a Hamiltonian as:

\begin{equation}
\mathcal{H} = \frac{1}{2}\sum_i^N \frac{|\mathbf{p}_i|^2}{2m_i}  + \sum_i^N \sum_{i<j}^N \frac{1}{2}k_ik_j(\mathbf{q}_i-\mathbf{q}_j)^2.
\label{eqn.nbody}
\end{equation}

\begin{figure*}[htb]
\centering
\includegraphics[width=.9\textwidth]{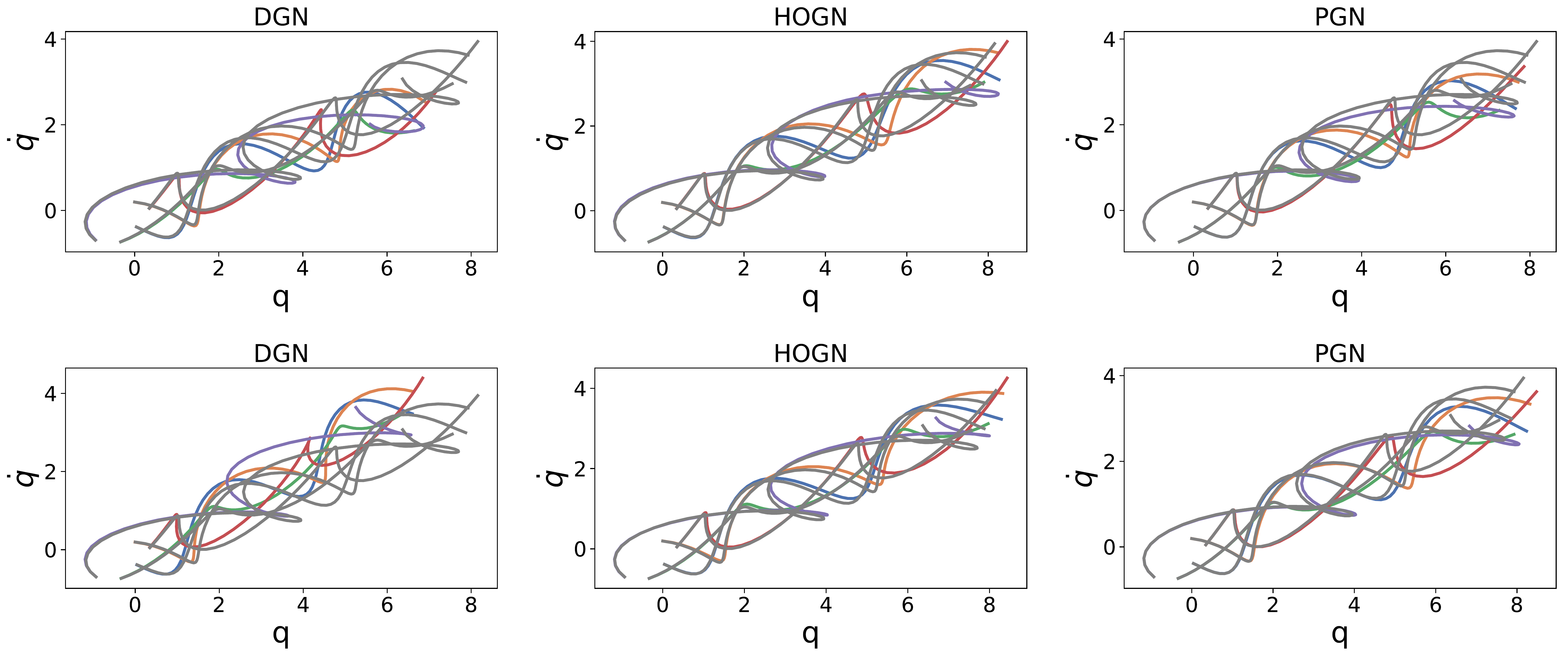}
\caption{Rollout of a single test point of the 5 body spring particle system with each model trained on noisy data. Each color represents a different particle. The ground truth is highlighted by black lines. The top row shows each method integrated with a RK4 integrator and the bottom 4th order symplectic integration. Unlike OGN, PGNs perform very well under symplectic integration and typically outperform HOGN in energy conservation.}
\label{fig.nspringtrial}
\end{figure*}

\noindent The overall mechanism closely aligns with important problems in N-particle systems used to model complex materials in solid state physics. Qualitative results for the 5-body problem are presented in Fig. \ref{fig.nspringtrial}.

\subsection*{H\'{e}non-Heiles system}

The systems investigate so far do not exhibit chaotic motion. H\'{e}non-Heiles is a system used to describe the nonlinear motion of a star around a galactic center and defined by the Hamiltonian:
\begin{equation}
\mathcal{H} = \frac{1}{2} \mathbf{p}^2 + \frac{1}{2} \mathbf{q}^2 +\lambda\left(q_x^2 q_y -\frac{q_y^3}{3}\right)
\label{eq:hh},
\end{equation}
which exhibits chaotic motion, i.e. small perturbations on initial conditions lead to drastically different trajectories. It has been shown that Hamiltonian Neural Networks can be used to capture dynamics in this setting \cite{choudhary_physics_2019, mattheakis_hamiltonian_2020}. 

We show the test results of all systems in Figure.~\ref{fig.f1} with models trained on noisy data using 10-step integration during training. Most notable is that potential-based networks consistently perform the best with potential graph networks doing the best in terms of state and energy MSE for most systems. In addition, the performance of RK-4 and the symplectic 4th order integrator are relatively comparable across most systems, indicating 4th order symplecticity constraints are more relevant for very large integration time steps such as in our 2-body problem or chaotic trajectories like H\'{e}non-Heiles. 

\section{Conclusion\label{sec:disc}}

From our extensive ablation across both noisy and non-noisy training data we find that VIGN is consistently the most performant in the noisy data setting. We believe that the reason inductive biases are not as successful with noiseless data is that they overfit to the training set in addition to the networks attempting to compensate for the error induced by numerical integration. Noise naturally reduces the overfitting and thus allows VIGN/PGN to do well. Once we identified VIGN as the most performant we needed to establish which biases were most useful. From \ref{fig.f1} we see that a potential network bias drives the largest performance increase against other approaches. We also see that graph based methods are more performant for larger many-body systems as is expected but remain robust in the single body settings too. We also find that using a long-range integration scheme in the noisy data setting tends to help the overall performance of all methods as it encourages the network to learn the underlying dynamics using multiple noisy points rather than one.

Although we do note that symplectic integrators are good for long range integration and energy preservation, their performance in many of the systems we investigate is only marginally better than Runge-Kutta. In preserving the energy, symplectic integrators are capable of drifting from the ground truth state while ensuring energy conservation which explains why we occasionally see RK methods doing much better at state and energy conservation. However, we do note that symplectic integrators of low order are much better at preserving the dynamics over low order Runge Kutta methods. This result is consistent with the theory of symplectic integrators.

We have shown that learning dynamics from data strongly benefits from well-chosen inductive biases. We present VIGNs as one such method capable of learning from scarce, noisy data across a diverse array of domains. We highlight that VIGNs are able to (1) unify graph networks, ODEs, potential networks, and symplectic inductive biases for learning precise trajectories in large many-body systems, (2) make learning data-efficient, (3) maintain flexibility in learning from generalized momenta and easily extended to canonincal coordinates, and (4) build higher order variational integrators through partitioned Runge-Kutta methods. 

\pagebreak

\pagebreak


\begin{thebibliography}{35}%
\makeatletter
\providecommand \@ifxundefined [1]{%
 \@ifx{#1\undefined}
}%
\providecommand \@ifnum [1]{%
 \ifnum #1\expandafter \@firstoftwo
 \else \expandafter \@secondoftwo
 \fi
}%
\providecommand \@ifx [1]{%
 \ifx #1\expandafter \@firstoftwo
 \else \expandafter \@secondoftwo
 \fi
}%
\providecommand \natexlab [1]{#1}%
\providecommand \enquote  [1]{``#1''}%
\providecommand \bibnamefont  [1]{#1}%
\providecommand \bibfnamefont [1]{#1}%
\providecommand \citenamefont [1]{#1}%
\providecommand \href@noop [0]{\@secondoftwo}%
\providecommand \href [0]{\begingroup \@sanitize@url \@href}%
\providecommand \@href[1]{\@@startlink{#1}\@@href}%
\providecommand \@@href[1]{\endgroup#1\@@endlink}%
\providecommand \@sanitize@url [0]{\catcode `\\12\catcode `\$12\catcode
  `\&12\catcode `\#12\catcode `\^12\catcode `\_12\catcode `\%12\relax}%
\providecommand \@@startlink[1]{}%
\providecommand \@@endlink[0]{}%
\providecommand \url  [0]{\begingroup\@sanitize@url \@url }%
\providecommand \@url [1]{\endgroup\@href {#1}{\urlprefix }}%
\providecommand \urlprefix  [0]{URL }%
\providecommand \Eprint [0]{\href }%
\providecommand \doibase [0]{https://doi.org/}%
\providecommand \selectlanguage [0]{\@gobble}%
\providecommand \bibinfo  [0]{\@secondoftwo}%
\providecommand \bibfield  [0]{\@secondoftwo}%
\providecommand \translation [1]{[#1]}%
\providecommand \BibitemOpen [0]{}%
\providecommand \bibitemStop [0]{}%
\providecommand \bibitemNoStop [0]{.\EOS\space}%
\providecommand \EOS [0]{\spacefactor3000\relax}%
\providecommand \BibitemShut  [1]{\csname bibitem#1\endcsname}%
\let\auto@bib@innerbib\@empty
\bibitem [{\citenamefont {Lutter}\ \emph {et~al.}(2019)\citenamefont {Lutter},
  \citenamefont {Ritter},\ and\ \citenamefont {Peters}}]{lutter2018deep}%
  \BibitemOpen
  \bibfield  {author} {\bibinfo {author} {\bibfnamefont {M.}~\bibnamefont
  {Lutter}}, \bibinfo {author} {\bibfnamefont {C.}~\bibnamefont {Ritter}},\
  and\ \bibinfo {author} {\bibfnamefont {J.}~\bibnamefont {Peters}},\
  }\bibfield  {title} {\bibinfo {title} {Deep lagrangian networks: Using
  physics as model prior for deep learning},\ }in\ \href
  {https://openreview.net/forum?id=BklHpjCqKm} {\emph {\bibinfo {booktitle}
  {International Conference on Learning Representations}}}\ (\bibinfo {year}
  {2019})\BibitemShut {NoStop}%
\bibitem [{\citenamefont {Barmparis}\ \emph {et~al.}(2020)\citenamefont
  {Barmparis}, \citenamefont {Neofotistos}, \citenamefont {Mattheakis},
  \citenamefont {Hizanidis}, \citenamefont {Tsironis},\ and\ \citenamefont
  {Kaxiras}}]{PLA_marios2020}%
  \BibitemOpen
  \bibfield  {author} {\bibinfo {author} {\bibfnamefont {G.}~\bibnamefont
  {Barmparis}}, \bibinfo {author} {\bibfnamefont {G.}~\bibnamefont
  {Neofotistos}}, \bibinfo {author} {\bibfnamefont {M.}~\bibnamefont
  {Mattheakis}}, \bibinfo {author} {\bibfnamefont {J.}~\bibnamefont
  {Hizanidis}}, \bibinfo {author} {\bibfnamefont {G.}~\bibnamefont
  {Tsironis}},\ and\ \bibinfo {author} {\bibfnamefont {E.}~\bibnamefont
  {Kaxiras}},\ }\bibfield  {title} {\bibinfo {title} {Robust prediction of
  complex spatiotemporal states through machine learning with sparse sensing},\
  }\href {https://doi.org/https://doi.org/10.1016/j.physleta.2020.126300}
  {\bibfield  {journal} {\bibinfo  {journal} {Physics Letters A}\ }\textbf
  {\bibinfo {volume} {384}},\ \bibinfo {pages} {126300} (\bibinfo {year}
  {2020})}\BibitemShut {NoStop}%
\bibitem [{\citenamefont {Li}\ \emph {et~al.}(2021)\citenamefont {Li},
  \citenamefont {Yang}, \citenamefont {Song},\ and\ \citenamefont
  {Cai}}]{li_conformation-guided_2021}%
  \BibitemOpen
  \bibfield  {author} {\bibinfo {author} {\bibfnamefont {Z.}~\bibnamefont
  {Li}}, \bibinfo {author} {\bibfnamefont {S.}~\bibnamefont {Yang}}, \bibinfo
  {author} {\bibfnamefont {G.}~\bibnamefont {Song}},\ and\ \bibinfo {author}
  {\bibfnamefont {L.}~\bibnamefont {Cai}},\ }\bibfield  {title} {\bibinfo
  {title} {{conformation}-{guided} {molecular} {representation} {with}
  {Hamiltonian} {neural} {networks}},\ }\href@noop {} {\ ,\ \bibinfo {pages}
  {11} (\bibinfo {year} {2021})}\BibitemShut {NoStop}%
\bibitem [{\citenamefont {Zhai}\ and\ \citenamefont
  {Hu}(2021)}]{zhai_inferring_2021}%
  \BibitemOpen
  \bibfield  {author} {\bibinfo {author} {\bibfnamefont {H.}~\bibnamefont
  {Zhai}}\ and\ \bibinfo {author} {\bibfnamefont {G.}~\bibnamefont {Hu}},\
  }\bibfield  {title} {\bibinfo {title} {Inferring micro-bubble dynamics with
  physics-informed deep learning},\ }\href {http://arxiv.org/abs/2105.07179}
  {\bibfield  {journal} {\bibinfo  {journal} {arXiv:2105.07179 [physics]}\ }
  (\bibinfo {year} {2021})},\ \bibinfo {note} {arXiv: 2105.07179}\BibitemShut
  {NoStop}%
\bibitem [{\citenamefont {Greydanus}\ \emph {et~al.}(2019)\citenamefont
  {Greydanus}, \citenamefont {Dzamba},\ and\ \citenamefont
  {Yosinski}}]{greydanus_hamiltonian_2019}%
  \BibitemOpen
  \bibfield  {author} {\bibinfo {author} {\bibfnamefont {S.}~\bibnamefont
  {Greydanus}}, \bibinfo {author} {\bibfnamefont {M.}~\bibnamefont {Dzamba}},\
  and\ \bibinfo {author} {\bibfnamefont {J.}~\bibnamefont {Yosinski}},\
  }\bibfield  {title} {\bibinfo {title} {Hamiltonian {Neural} {Networks}},\
  }in\ \href {http://papers.nips.cc/paper/9672-hamiltonian-neural-networks.pdf}
  {\emph {\bibinfo {booktitle} {Advances in {Neural} {Information} {Processing}
  {Systems} 32}}},\ \bibinfo {editor} {edited by\ \bibinfo {editor}
  {\bibfnamefont {H.}~\bibnamefont {Wallach}}, \bibinfo {editor} {\bibfnamefont
  {H.}~\bibnamefont {Larochelle}}, \bibinfo {editor} {\bibfnamefont
  {A.}~\bibnamefont {Beygelzimer}}, \bibinfo {editor} {\bibfnamefont {F.~d.}\
  \bibnamefont {Alché-Buc}}, \bibinfo {editor} {\bibfnamefont
  {E.}~\bibnamefont {Fox}},\ and\ \bibinfo {editor} {\bibfnamefont
  {R.}~\bibnamefont {Garnett}}}\ (\bibinfo  {publisher} {Curran Associates,
  Inc.},\ \bibinfo {year} {2019})\ pp.\ \bibinfo {pages}
  {15379--15389}\BibitemShut {NoStop}%
\bibitem [{\citenamefont {Saemundsson}\ \emph {et~al.}(2019)\citenamefont
  {Saemundsson}, \citenamefont {Terenin}, \citenamefont {Hofmann},\ and\
  \citenamefont {Deisenroth}}]{saemundsson_variational_2019}%
  \BibitemOpen
  \bibfield  {author} {\bibinfo {author} {\bibfnamefont {S.}~\bibnamefont
  {Saemundsson}}, \bibinfo {author} {\bibfnamefont {A.}~\bibnamefont
  {Terenin}}, \bibinfo {author} {\bibfnamefont {K.}~\bibnamefont {Hofmann}},\
  and\ \bibinfo {author} {\bibfnamefont {M.~P.}\ \bibnamefont {Deisenroth}},\
  }\bibfield  {title} {\bibinfo {title} {Variational {Integrator} {Networks}
  for {Physically} {Meaningful} {Embeddings}},\ }\href
  {http://arxiv.org/abs/1910.09349} {\bibfield  {journal} {\bibinfo  {journal}
  {arXiv:1910.09349 [cs, stat]}\ } (\bibinfo {year} {2019})},\ \bibinfo {note}
  {arXiv: 1910.09349}\BibitemShut {NoStop}%
\bibitem [{\citenamefont {Chang}\ \emph {et~al.}(2017)\citenamefont {Chang},
  \citenamefont {Meng}, \citenamefont {Haber}, \citenamefont {Ruthotto},
  \citenamefont {Begert},\ and\ \citenamefont
  {Holtham}}]{chang_reversible_2017}%
  \BibitemOpen
  \bibfield  {author} {\bibinfo {author} {\bibfnamefont {B.}~\bibnamefont
  {Chang}}, \bibinfo {author} {\bibfnamefont {L.}~\bibnamefont {Meng}},
  \bibinfo {author} {\bibfnamefont {E.}~\bibnamefont {Haber}}, \bibinfo
  {author} {\bibfnamefont {L.}~\bibnamefont {Ruthotto}}, \bibinfo {author}
  {\bibfnamefont {D.}~\bibnamefont {Begert}},\ and\ \bibinfo {author}
  {\bibfnamefont {E.}~\bibnamefont {Holtham}},\ }\bibfield  {title} {\bibinfo
  {title} {Reversible {Architectures} for {Arbitrarily} {Deep} {Residual}
  {Neural} {Networks}},\ }\href {http://arxiv.org/abs/1709.03698} {\bibfield
  {journal} {\bibinfo  {journal} {arXiv:1709.03698 [cs, stat]}\ } (\bibinfo
  {year} {2017})},\ \bibinfo {note} {arXiv: 1709.03698}\BibitemShut {NoStop}%
\bibitem [{\citenamefont {Chen}\ \emph {et~al.}(2018)\citenamefont {Chen},
  \citenamefont {Rubanova}, \citenamefont {Bettencourt},\ and\ \citenamefont
  {Duvenaud}}]{chen_neural_2018}%
  \BibitemOpen
  \bibfield  {author} {\bibinfo {author} {\bibfnamefont {R.~T.~Q.}\
  \bibnamefont {Chen}}, \bibinfo {author} {\bibfnamefont {Y.}~\bibnamefont
  {Rubanova}}, \bibinfo {author} {\bibfnamefont {J.}~\bibnamefont
  {Bettencourt}},\ and\ \bibinfo {author} {\bibfnamefont {D.~K.}\ \bibnamefont
  {Duvenaud}},\ }\bibfield  {title} {\bibinfo {title} {Neural {Ordinary}
  {Differential} {Equations}},\ }in\ \href
  {http://papers.nips.cc/paper/7892-neural-ordinary-differential-equations.pdf}
  {\emph {\bibinfo {booktitle} {Advances in {Neural} {Information} {Processing}
  {Systems} 31}}},\ \bibinfo {editor} {edited by\ \bibinfo {editor}
  {\bibfnamefont {S.}~\bibnamefont {Bengio}}, \bibinfo {editor} {\bibfnamefont
  {H.}~\bibnamefont {Wallach}}, \bibinfo {editor} {\bibfnamefont
  {H.}~\bibnamefont {Larochelle}}, \bibinfo {editor} {\bibfnamefont
  {K.}~\bibnamefont {Grauman}}, \bibinfo {editor} {\bibfnamefont
  {N.}~\bibnamefont {Cesa-Bianchi}},\ and\ \bibinfo {editor} {\bibfnamefont
  {R.}~\bibnamefont {Garnett}}}\ (\bibinfo  {publisher} {Curran Associates,
  Inc.},\ \bibinfo {year} {2018})\ pp.\ \bibinfo {pages}
  {6571--6583}\BibitemShut {NoStop}%
\bibitem [{\citenamefont {Battaglia}\ \emph {et~al.}(2018)\citenamefont
  {Battaglia}, \citenamefont {Hamrick}, \citenamefont {Bapst}, \citenamefont
  {Sanchez-Gonzalez}, \citenamefont {Zambaldi}, \citenamefont {Malinowski},
  \citenamefont {Tacchetti}, \citenamefont {Raposo}, \citenamefont {Santoro},
  \citenamefont {Faulkner}, \citenamefont {Gulcehre}, \citenamefont {Song},
  \citenamefont {Ballard}, \citenamefont {Gilmer}, \citenamefont {Dahl},
  \citenamefont {Vaswani}, \citenamefont {Allen}, \citenamefont {Nash},
  \citenamefont {Langston}, \citenamefont {Dyer}, \citenamefont {Heess},
  \citenamefont {Wierstra}, \citenamefont {Kohli}, \citenamefont {Botvinick},
  \citenamefont {Vinyals}, \citenamefont {Li},\ and\ \citenamefont
  {Pascanu}}]{battaglia_relational_2018}%
  \BibitemOpen
  \bibfield  {author} {\bibinfo {author} {\bibfnamefont {P.~W.}\ \bibnamefont
  {Battaglia}}, \bibinfo {author} {\bibfnamefont {J.~B.}\ \bibnamefont
  {Hamrick}}, \bibinfo {author} {\bibfnamefont {V.}~\bibnamefont {Bapst}},
  \bibinfo {author} {\bibfnamefont {A.}~\bibnamefont {Sanchez-Gonzalez}},
  \bibinfo {author} {\bibfnamefont {V.}~\bibnamefont {Zambaldi}}, \bibinfo
  {author} {\bibfnamefont {M.}~\bibnamefont {Malinowski}}, \bibinfo {author}
  {\bibfnamefont {A.}~\bibnamefont {Tacchetti}}, \bibinfo {author}
  {\bibfnamefont {D.}~\bibnamefont {Raposo}}, \bibinfo {author} {\bibfnamefont
  {A.}~\bibnamefont {Santoro}}, \bibinfo {author} {\bibfnamefont
  {R.}~\bibnamefont {Faulkner}}, \bibinfo {author} {\bibfnamefont
  {C.}~\bibnamefont {Gulcehre}}, \bibinfo {author} {\bibfnamefont
  {F.}~\bibnamefont {Song}}, \bibinfo {author} {\bibfnamefont {A.}~\bibnamefont
  {Ballard}}, \bibinfo {author} {\bibfnamefont {J.}~\bibnamefont {Gilmer}},
  \bibinfo {author} {\bibfnamefont {G.}~\bibnamefont {Dahl}}, \bibinfo {author}
  {\bibfnamefont {A.}~\bibnamefont {Vaswani}}, \bibinfo {author} {\bibfnamefont
  {K.}~\bibnamefont {Allen}}, \bibinfo {author} {\bibfnamefont
  {C.}~\bibnamefont {Nash}}, \bibinfo {author} {\bibfnamefont {V.}~\bibnamefont
  {Langston}}, \bibinfo {author} {\bibfnamefont {C.}~\bibnamefont {Dyer}},
  \bibinfo {author} {\bibfnamefont {N.}~\bibnamefont {Heess}}, \bibinfo
  {author} {\bibfnamefont {D.}~\bibnamefont {Wierstra}}, \bibinfo {author}
  {\bibfnamefont {P.}~\bibnamefont {Kohli}}, \bibinfo {author} {\bibfnamefont
  {M.}~\bibnamefont {Botvinick}}, \bibinfo {author} {\bibfnamefont
  {O.}~\bibnamefont {Vinyals}}, \bibinfo {author} {\bibfnamefont
  {Y.}~\bibnamefont {Li}},\ and\ \bibinfo {author} {\bibfnamefont
  {R.}~\bibnamefont {Pascanu}},\ }\bibfield  {title} {\bibinfo {title}
  {Relational inductive biases, deep learning, and graph networks},\ }\href
  {http://arxiv.org/abs/1806.01261} {\bibfield  {journal} {\bibinfo  {journal}
  {arXiv:1806.01261 [cs, stat]}\ } (\bibinfo {year} {2018})},\ \bibinfo {note}
  {arXiv: 1806.01261}\BibitemShut {NoStop}%
\bibitem [{\citenamefont {Sanchez-Gonzalez}\ \emph {et~al.}(2018)\citenamefont
  {Sanchez-Gonzalez}, \citenamefont {Heess}, \citenamefont {Springenberg},
  \citenamefont {Merel}, \citenamefont {Riedmiller}, \citenamefont {Hadsell},\
  and\ \citenamefont {Battaglia}}]{sanchez-gonzalez_graph_2018}%
  \BibitemOpen
  \bibfield  {author} {\bibinfo {author} {\bibfnamefont {A.}~\bibnamefont
  {Sanchez-Gonzalez}}, \bibinfo {author} {\bibfnamefont {N.}~\bibnamefont
  {Heess}}, \bibinfo {author} {\bibfnamefont {J.~T.}\ \bibnamefont
  {Springenberg}}, \bibinfo {author} {\bibfnamefont {J.}~\bibnamefont {Merel}},
  \bibinfo {author} {\bibfnamefont {M.}~\bibnamefont {Riedmiller}}, \bibinfo
  {author} {\bibfnamefont {R.}~\bibnamefont {Hadsell}},\ and\ \bibinfo {author}
  {\bibfnamefont {P.}~\bibnamefont {Battaglia}},\ }\bibfield  {title} {\bibinfo
  {title} {Graph networks as learnable physics engines for inference and
  control},\ }\href {http://arxiv.org/abs/1806.01242} {\bibfield  {journal}
  {\bibinfo  {journal} {arXiv:1806.01242 [cs, stat]}\ } (\bibinfo {year}
  {2018})},\ \bibinfo {note} {arXiv: 1806.01242}\BibitemShut {NoStop}%
\bibitem [{\citenamefont {Sanchez-Gonzalez}\ \emph {et~al.}(2020)\citenamefont
  {Sanchez-Gonzalez}, \citenamefont {Godwin}, \citenamefont {Pfaff},
  \citenamefont {Ying}, \citenamefont {Leskovec},\ and\ \citenamefont
  {Battaglia}}]{sanchez-gonzalez_learning_2020}%
  \BibitemOpen
  \bibfield  {author} {\bibinfo {author} {\bibfnamefont {A.}~\bibnamefont
  {Sanchez-Gonzalez}}, \bibinfo {author} {\bibfnamefont {J.}~\bibnamefont
  {Godwin}}, \bibinfo {author} {\bibfnamefont {T.}~\bibnamefont {Pfaff}},
  \bibinfo {author} {\bibfnamefont {R.}~\bibnamefont {Ying}}, \bibinfo {author}
  {\bibfnamefont {J.}~\bibnamefont {Leskovec}},\ and\ \bibinfo {author}
  {\bibfnamefont {P.~W.}\ \bibnamefont {Battaglia}},\ }\bibfield  {title}
  {\bibinfo {title} {Learning to {Simulate} {Complex} {Physics} with {Graph}
  {Networks}},\ }\href {http://arxiv.org/abs/2002.09405} {\bibfield  {journal}
  {\bibinfo  {journal} {arXiv:2002.09405 [physics, stat]}\ } (\bibinfo {year}
  {2020})},\ \bibinfo {note} {arXiv: 2002.09405}\BibitemShut {NoStop}%
\bibitem [{\citenamefont {Sanchez-Gonzalez}\ \emph {et~al.}(2019)\citenamefont
  {Sanchez-Gonzalez}, \citenamefont {Bapst}, \citenamefont {Cranmer},\ and\
  \citenamefont {Battaglia}}]{sanchez-gonzalez_hamiltonian_2019}%
  \BibitemOpen
  \bibfield  {author} {\bibinfo {author} {\bibfnamefont {A.}~\bibnamefont
  {Sanchez-Gonzalez}}, \bibinfo {author} {\bibfnamefont {V.}~\bibnamefont
  {Bapst}}, \bibinfo {author} {\bibfnamefont {K.}~\bibnamefont {Cranmer}},\
  and\ \bibinfo {author} {\bibfnamefont {P.}~\bibnamefont {Battaglia}},\
  }\bibfield  {title} {\bibinfo {title} {Hamiltonian {Graph} {Networks} with
  {ODE} {Integrators}},\ }\href {http://arxiv.org/abs/1909.12790} {\bibfield
  {journal} {\bibinfo  {journal} {arXiv:1909.12790 [physics]}\ } (\bibinfo
  {year} {2019})},\ \bibinfo {note} {arXiv: 1909.12790}\BibitemShut {NoStop}%
\bibitem [{\citenamefont {Battaglia}\ \emph {et~al.}(2016)\citenamefont
  {Battaglia}, \citenamefont {Pascanu}, \citenamefont {Lai}, \citenamefont
  {Rezende},\ and\ \citenamefont {Kavukcuoglu}}]{battaglia_interaction_2016}%
  \BibitemOpen
  \bibfield  {author} {\bibinfo {author} {\bibfnamefont {P.~W.}\ \bibnamefont
  {Battaglia}}, \bibinfo {author} {\bibfnamefont {R.}~\bibnamefont {Pascanu}},
  \bibinfo {author} {\bibfnamefont {M.}~\bibnamefont {Lai}}, \bibinfo {author}
  {\bibfnamefont {D.}~\bibnamefont {Rezende}},\ and\ \bibinfo {author}
  {\bibfnamefont {K.}~\bibnamefont {Kavukcuoglu}},\ }\bibfield  {title}
  {\bibinfo {title} {Interaction {Networks} for {Learning} about {Objects},
  {Relations} and {Physics}},\ }\href {http://arxiv.org/abs/1612.00222}
  {\bibfield  {journal} {\bibinfo  {journal} {arXiv:1612.00222 [cs]}\ }
  (\bibinfo {year} {2016})},\ \bibinfo {note} {arXiv: 1612.00222}\BibitemShut
  {NoStop}%
\bibitem [{\citenamefont {Seo}\ and\ \citenamefont
  {Liu}(2019)}]{seo_differentiable_2019}%
  \BibitemOpen
  \bibfield  {author} {\bibinfo {author} {\bibfnamefont {S.}~\bibnamefont
  {Seo}}\ and\ \bibinfo {author} {\bibfnamefont {Y.}~\bibnamefont {Liu}},\
  }\bibfield  {title} {\bibinfo {title} {Differentiable {Physics}-informed
  {Graph} {Networks}},\ }\href {http://arxiv.org/abs/1902.02950} {\bibfield
  {journal} {\bibinfo  {journal} {arXiv:1902.02950 [cs, stat]}\ } (\bibinfo
  {year} {2019})},\ \bibinfo {note} {arXiv: 1902.02950}\BibitemShut {NoStop}%
\bibitem [{\citenamefont {Cranmer}\ \emph {et~al.}(2019)\citenamefont
  {Cranmer}, \citenamefont {Xu}, \citenamefont {Battaglia},\ and\ \citenamefont
  {Ho}}]{cranmer_learning_2019}%
  \BibitemOpen
  \bibfield  {author} {\bibinfo {author} {\bibfnamefont {M.~D.}\ \bibnamefont
  {Cranmer}}, \bibinfo {author} {\bibfnamefont {R.}~\bibnamefont {Xu}},
  \bibinfo {author} {\bibfnamefont {P.}~\bibnamefont {Battaglia}},\ and\
  \bibinfo {author} {\bibfnamefont {S.}~\bibnamefont {Ho}},\ }\bibfield
  {title} {\bibinfo {title} {Learning {Symbolic} {Physics} with {Graph}
  {Networks}},\ }\href {http://arxiv.org/abs/1909.05862} {\bibfield  {journal}
  {\bibinfo  {journal} {arXiv:1909.05862 [astro-ph, physics:physics, stat]}\ }
  (\bibinfo {year} {2019})},\ \bibinfo {note} {arXiv: 1909.05862}\BibitemShut
  {NoStop}%
\bibitem [{\citenamefont {Seo}\ \emph {et~al.}(2020)\citenamefont {Seo},
  \citenamefont {Meng},\ and\ \citenamefont {Liu}}]{seo_physics-aware_2020}%
  \BibitemOpen
  \bibfield  {author} {\bibinfo {author} {\bibfnamefont {S.}~\bibnamefont
  {Seo}}, \bibinfo {author} {\bibfnamefont {C.}~\bibnamefont {Meng}},\ and\
  \bibinfo {author} {\bibfnamefont {Y.}~\bibnamefont {Liu}},\ }\bibfield
  {title} {\bibinfo {title} {{physics}-{aware} {difference} {graph} {networks}
  {for} {sparsely}-{observed} {dynamics}},\ }\href@noop {} {\ ,\ \bibinfo
  {pages} {15} (\bibinfo {year} {2020})}\BibitemShut {NoStop}%
\bibitem [{\citenamefont {Lamb}\ \emph {et~al.}(2020)\citenamefont {Lamb},
  \citenamefont {Garcez}, \citenamefont {Gori}, \citenamefont {Prates},
  \citenamefont {Avelar},\ and\ \citenamefont {Vardi}}]{lamb_graph_2020}%
  \BibitemOpen
  \bibfield  {author} {\bibinfo {author} {\bibfnamefont {L.}~\bibnamefont
  {Lamb}}, \bibinfo {author} {\bibfnamefont {A.}~\bibnamefont {Garcez}},
  \bibinfo {author} {\bibfnamefont {M.}~\bibnamefont {Gori}}, \bibinfo {author}
  {\bibfnamefont {M.}~\bibnamefont {Prates}}, \bibinfo {author} {\bibfnamefont
  {P.}~\bibnamefont {Avelar}},\ and\ \bibinfo {author} {\bibfnamefont
  {M.}~\bibnamefont {Vardi}},\ }\bibfield  {title} {\bibinfo {title} {Graph
  {Neural} {Networks} {Meet} {Neural}-{Symbolic} {Computing}: {A} {Survey} and
  {Perspective}},\ }\href {http://arxiv.org/abs/2003.00330} {\bibfield
  {journal} {\bibinfo  {journal} {arXiv:2003.00330 [cs]}\ } (\bibinfo {year}
  {2020})},\ \bibinfo {note} {arXiv: 2003.00330}\BibitemShut {NoStop}%
\bibitem [{\citenamefont {Cranmer}\ \emph {et~al.}(2020)\citenamefont
  {Cranmer}, \citenamefont {Greydanus}, \citenamefont {Hoyer}, \citenamefont
  {Battaglia}, \citenamefont {Spergel},\ and\ \citenamefont
  {Ho}}]{cranmer_lagrangian_2020}%
  \BibitemOpen
  \bibfield  {author} {\bibinfo {author} {\bibfnamefont {M.}~\bibnamefont
  {Cranmer}}, \bibinfo {author} {\bibfnamefont {S.}~\bibnamefont {Greydanus}},
  \bibinfo {author} {\bibfnamefont {S.}~\bibnamefont {Hoyer}}, \bibinfo
  {author} {\bibfnamefont {P.}~\bibnamefont {Battaglia}}, \bibinfo {author}
  {\bibfnamefont {D.}~\bibnamefont {Spergel}},\ and\ \bibinfo {author}
  {\bibfnamefont {S.}~\bibnamefont {Ho}},\ }\bibfield  {title} {\bibinfo
  {title} {Lagrangian {Neural} {Networks}},\ }\href
  {http://arxiv.org/abs/2003.04630} {\bibfield  {journal} {\bibinfo  {journal}
  {arXiv:2003.04630 [physics, stat]}\ } (\bibinfo {year} {2020})},\ \bibinfo
  {note} {arXiv: 2003.04630}\BibitemShut {NoStop}%
\bibitem [{\citenamefont {Yu}\ \emph {et~al.}(2020)\citenamefont {Yu},
  \citenamefont {Tian}, \citenamefont {E},\ and\ \citenamefont
  {Li}}]{yu_onsagernet_2020}%
  \BibitemOpen
  \bibfield  {author} {\bibinfo {author} {\bibfnamefont {H.}~\bibnamefont
  {Yu}}, \bibinfo {author} {\bibfnamefont {X.}~\bibnamefont {Tian}}, \bibinfo
  {author} {\bibfnamefont {W.}~\bibnamefont {E}},\ and\ \bibinfo {author}
  {\bibfnamefont {Q.}~\bibnamefont {Li}},\ }\bibfield  {title} {\bibinfo
  {title} {{OnsagerNet}: {Learning} {Stable} and {Interpretable} {Dynamics}
  using a {Generalized} {Onsager} {Principle}},\ }\href
  {http://arxiv.org/abs/2009.02327} {\bibfield  {journal} {\bibinfo  {journal}
  {arXiv:2009.02327 [physics]}\ } (\bibinfo {year} {2020})},\ \bibinfo {note}
  {arXiv: 2009.02327}\BibitemShut {NoStop}%
\bibitem [{\citenamefont {Zhong}\ \emph {et~al.}(2019)\citenamefont {Zhong},
  \citenamefont {Dey},\ and\ \citenamefont
  {Chakraborty}}]{zhong_symplectic_2019}%
  \BibitemOpen
  \bibfield  {author} {\bibinfo {author} {\bibfnamefont {Y.~D.}\ \bibnamefont
  {Zhong}}, \bibinfo {author} {\bibfnamefont {B.}~\bibnamefont {Dey}},\ and\
  \bibinfo {author} {\bibfnamefont {A.}~\bibnamefont {Chakraborty}},\
  }\bibfield  {title} {\bibinfo {title} {Symplectic {ODE}-{Net}: {Learning}
  {Hamiltonian} {Dynamics} with {Control}},\ }\href
  {http://arxiv.org/abs/1909.12077} {\bibfield  {journal} {\bibinfo  {journal}
  {arXiv:1909.12077 [physics, stat]}\ } (\bibinfo {year} {2019})},\ \bibinfo
  {note} {arXiv: 1909.12077}\BibitemShut {NoStop}%
\bibitem [{\citenamefont {Marsden}\ and\ \citenamefont
  {West}(2001)}]{marsden_discrete_2001}%
  \BibitemOpen
  \bibfield  {author} {\bibinfo {author} {\bibfnamefont {J.~E.}\ \bibnamefont
  {Marsden}}\ and\ \bibinfo {author} {\bibfnamefont {M.}~\bibnamefont {West}},\
  }\bibfield  {title} {\bibinfo {title} {Discrete mechanics and variational
  integrators},\ }\href {https://doi.org/10.1017/S096249290100006X} {\bibfield
  {journal} {\bibinfo  {journal} {Acta Numerica}\ }\textbf {\bibinfo {volume}
  {10}},\ \bibinfo {pages} {357} (\bibinfo {year} {2001})}\BibitemShut
  {NoStop}%
\bibitem [{\citenamefont {Chen}\ \emph {et~al.}(2020)\citenamefont {Chen},
  \citenamefont {Zhang}, \citenamefont {Arjovsky},\ and\ \citenamefont
  {Bottou}}]{chen_symplectic_2020}%
  \BibitemOpen
  \bibfield  {author} {\bibinfo {author} {\bibfnamefont {Z.}~\bibnamefont
  {Chen}}, \bibinfo {author} {\bibfnamefont {J.}~\bibnamefont {Zhang}},
  \bibinfo {author} {\bibfnamefont {M.}~\bibnamefont {Arjovsky}},\ and\
  \bibinfo {author} {\bibfnamefont {L.}~\bibnamefont {Bottou}},\ }\bibfield
  {title} {\bibinfo {title} {Symplectic {Recurrent} {Neural} {Networks}},\
  }\href {http://arxiv.org/abs/1909.13334} {\bibfield  {journal} {\bibinfo
  {journal} {arXiv:1909.13334 [cs, stat]}\ } (\bibinfo {year} {2020})},\
  \bibinfo {note} {arXiv: 1909.13334}\BibitemShut {NoStop}%
\bibitem [{\citenamefont {Witkoskie}\ and\ \citenamefont
  {Doren}(2005)}]{witkoskie_neural_2005}%
  \BibitemOpen
  \bibfield  {author} {\bibinfo {author} {\bibfnamefont {J.~B.}\ \bibnamefont
  {Witkoskie}}\ and\ \bibinfo {author} {\bibfnamefont {D.~J.}\ \bibnamefont
  {Doren}},\ }\bibfield  {title} {\bibinfo {title} {Neural {Network} {Models}
  of {Potential} {Energy} {Surfaces}: {Prototypical} {Examples}},\ }\href
  {https://doi.org/10.1021/ct049976i} {\bibfield  {journal} {\bibinfo
  {journal} {Journal of Chemical Theory and Computation}\ }\textbf {\bibinfo
  {volume} {1}},\ \bibinfo {pages} {14} (\bibinfo {year} {2005})}\BibitemShut
  {NoStop}%
\bibitem [{\citenamefont {Pukrittayakamee}\ \emph {et~al.}(2009)\citenamefont
  {Pukrittayakamee}, \citenamefont {Malshe}, \citenamefont {Hagan},
  \citenamefont {Raff}, \citenamefont {Narulkar}, \citenamefont {Bukkapatnum},\
  and\ \citenamefont {Komanduri}}]{pukrittayakamee_simultaneous_2009}%
  \BibitemOpen
  \bibfield  {author} {\bibinfo {author} {\bibfnamefont {A.}~\bibnamefont
  {Pukrittayakamee}}, \bibinfo {author} {\bibfnamefont {M.}~\bibnamefont
  {Malshe}}, \bibinfo {author} {\bibfnamefont {M.}~\bibnamefont {Hagan}},
  \bibinfo {author} {\bibfnamefont {L.~M.}\ \bibnamefont {Raff}}, \bibinfo
  {author} {\bibfnamefont {R.}~\bibnamefont {Narulkar}}, \bibinfo {author}
  {\bibfnamefont {S.}~\bibnamefont {Bukkapatnum}},\ and\ \bibinfo {author}
  {\bibfnamefont {R.}~\bibnamefont {Komanduri}},\ }\bibfield  {title} {\bibinfo
  {title} {Simultaneous fitting of a potential-energy surface and its
  corresponding force fields using feedforward neural networks},\ }\href
  {https://doi.org/10.1063/1.3095491} {\bibfield  {journal} {\bibinfo
  {journal} {The Journal of Chemical Physics}\ }\textbf {\bibinfo {volume}
  {130}},\ \bibinfo {pages} {134101} (\bibinfo {year} {2009})}\BibitemShut
  {NoStop}%
\bibitem [{\citenamefont {Smith}\ \emph {et~al.}(2017)\citenamefont {Smith},
  \citenamefont {Isayev},\ and\ \citenamefont {Roitberg}}]{smith_ani-1_2017}%
  \BibitemOpen
  \bibfield  {author} {\bibinfo {author} {\bibfnamefont {J.~S.}\ \bibnamefont
  {Smith}}, \bibinfo {author} {\bibfnamefont {O.}~\bibnamefont {Isayev}},\ and\
  \bibinfo {author} {\bibfnamefont {A.~E.}\ \bibnamefont {Roitberg}},\
  }\bibfield  {title} {\bibinfo {title} {{ANI}-1: an extensible neural network
  potential with {DFT} accuracy at force field computational cost},\ }\href
  {https://doi.org/10.1039/C6SC05720A} {\bibfield  {journal} {\bibinfo
  {journal} {Chemical Science}\ }\textbf {\bibinfo {volume} {8}},\ \bibinfo
  {pages} {3192} (\bibinfo {year} {2017})}\BibitemShut {NoStop}%
\bibitem [{\citenamefont {Rupp}\ \emph {et~al.}(2012)\citenamefont {Rupp},
  \citenamefont {Tkatchenko}, \citenamefont {Müller},\ and\ \citenamefont {von
  Lilienfeld}}]{rupp_fast_2012}%
  \BibitemOpen
  \bibfield  {author} {\bibinfo {author} {\bibfnamefont {M.}~\bibnamefont
  {Rupp}}, \bibinfo {author} {\bibfnamefont {A.}~\bibnamefont {Tkatchenko}},
  \bibinfo {author} {\bibfnamefont {K.-R.}\ \bibnamefont {Müller}},\ and\
  \bibinfo {author} {\bibfnamefont {O.~A.}\ \bibnamefont {von Lilienfeld}},\
  }\bibfield  {title} {\bibinfo {title} {Fast and {Accurate} {Modeling} of
  {Molecular} {Atomization} {Energies} with {Machine} {Learning}},\ }\href
  {https://doi.org/10.1103/PhysRevLett.108.058301} {\bibfield  {journal}
  {\bibinfo  {journal} {Physical Review Letters}\ }\textbf {\bibinfo {volume}
  {108}},\ \bibinfo {pages} {058301} (\bibinfo {year} {2012})},\ \bibinfo
  {note} {publisher: American Physical Society}\BibitemShut {NoStop}%
\bibitem [{\citenamefont {Yao}\ \emph {et~al.}(2018)\citenamefont {Yao},
  \citenamefont {Herr}, \citenamefont {Toth}, \citenamefont {Mckintyre},\ and\
  \citenamefont {Parkhill}}]{yao_tensormol-01_2018}%
  \BibitemOpen
  \bibfield  {author} {\bibinfo {author} {\bibfnamefont {K.}~\bibnamefont
  {Yao}}, \bibinfo {author} {\bibfnamefont {J.~E.}\ \bibnamefont {Herr}},
  \bibinfo {author} {\bibfnamefont {D.}~\bibnamefont {Toth}}, \bibinfo {author}
  {\bibfnamefont {R.}~\bibnamefont {Mckintyre}},\ and\ \bibinfo {author}
  {\bibfnamefont {J.}~\bibnamefont {Parkhill}},\ }\bibfield  {title} {\bibinfo
  {title} {The {TensorMol}-0.1 model chemistry: a neural network augmented with
  long-range physics},\ }\href {https://doi.org/10.1039/C7SC04934J} {\bibfield
  {journal} {\bibinfo  {journal} {Chemical Science}\ }\textbf {\bibinfo
  {volume} {9}},\ \bibinfo {pages} {2261} (\bibinfo {year} {2018})}\BibitemShut
  {NoStop}%
\bibitem [{\citenamefont {Hills}\ \emph {et~al.}(2015)\citenamefont {Hills},
  \citenamefont {Grütter},\ and\ \citenamefont
  {Hudson}}]{hills_algorithm_2015}%
  \BibitemOpen
  \bibfield  {author} {\bibinfo {author} {\bibfnamefont {D.~J.}\ \bibnamefont
  {Hills}}, \bibinfo {author} {\bibfnamefont {A.~M.}\ \bibnamefont
  {Grütter}},\ and\ \bibinfo {author} {\bibfnamefont {J.~J.}\ \bibnamefont
  {Hudson}},\ }\bibfield  {title} {\bibinfo {title} {An algorithm for
  discovering {Lagrangians} automatically from data},\ }\href
  {https://doi.org/10.7717/peerj-cs.31} {\bibfield  {journal} {\bibinfo
  {journal} {PeerJ Computer Science}\ }\textbf {\bibinfo {volume} {1}},\
  \bibinfo {pages} {e31} (\bibinfo {year} {2015})}\BibitemShut {NoStop}%
\bibitem [{\citenamefont {Iten}\ \emph {et~al.}(2018)\citenamefont {Iten},
  \citenamefont {Metger}, \citenamefont {Wilming}, \citenamefont {del Rio},\
  and\ \citenamefont {Renner}}]{iten_discovering_2018}%
  \BibitemOpen
  \bibfield  {author} {\bibinfo {author} {\bibfnamefont {R.}~\bibnamefont
  {Iten}}, \bibinfo {author} {\bibfnamefont {T.}~\bibnamefont {Metger}},
  \bibinfo {author} {\bibfnamefont {H.}~\bibnamefont {Wilming}}, \bibinfo
  {author} {\bibfnamefont {L.}~\bibnamefont {del Rio}},\ and\ \bibinfo {author}
  {\bibfnamefont {R.}~\bibnamefont {Renner}},\ }\bibfield  {title} {\bibinfo
  {title} {Discovering physical concepts with neural networks},\ }\href
  {http://arxiv.org/abs/1807.10300} {\bibfield  {journal} {\bibinfo  {journal}
  {arXiv:1807.10300 [physics, physics:quant-ph]}\ } (\bibinfo {year} {2018})},\
  \bibinfo {note} {arXiv: 1807.10300}\BibitemShut {NoStop}%
\bibitem [{\citenamefont {Schmidt}\ and\ \citenamefont
  {Lipson}(2009)}]{schmidt_distilling_2009}%
  \BibitemOpen
  \bibfield  {author} {\bibinfo {author} {\bibfnamefont {M.}~\bibnamefont
  {Schmidt}}\ and\ \bibinfo {author} {\bibfnamefont {H.}~\bibnamefont
  {Lipson}},\ }\bibfield  {title} {\bibinfo {title} {Distilling {Free}-{Form}
  {Natural} {Laws} from {Experimental} {Data}},\ }\href
  {https://doi.org/10.1126/science.1165893} {\bibfield  {journal} {\bibinfo
  {journal} {Science}\ }\textbf {\bibinfo {volume} {324}},\ \bibinfo {pages}
  {81} (\bibinfo {year} {2009})},\ \bibinfo {note} {publisher: American
  Association for the Advancement of Science \_eprint:
  https://science.sciencemag.org/content/324/5923/81.full.pdf}\BibitemShut
  {NoStop}%
\bibitem [{\citenamefont {de~Silva}\ \emph {et~al.}(2019)\citenamefont
  {de~Silva}, \citenamefont {Higdon}, \citenamefont {Brunton},\ and\
  \citenamefont {Kutz}}]{de_silva_discovery_2019}%
  \BibitemOpen
  \bibfield  {author} {\bibinfo {author} {\bibfnamefont {B.}~\bibnamefont
  {de~Silva}}, \bibinfo {author} {\bibfnamefont {D.~M.}\ \bibnamefont
  {Higdon}}, \bibinfo {author} {\bibfnamefont {S.~L.}\ \bibnamefont
  {Brunton}},\ and\ \bibinfo {author} {\bibfnamefont {J.~N.}\ \bibnamefont
  {Kutz}},\ }\bibfield  {title} {\bibinfo {title} {Discovery of {Physics} from
  {Data}: {Universal} {Laws} and {Discrepancy} {Models}},\ }\href
  {http://arxiv.org/abs/1906.07906} {\bibfield  {journal} {\bibinfo  {journal}
  {arXiv:1906.07906 [physics, stat]}\ } (\bibinfo {year} {2019})},\ \bibinfo
  {note} {arXiv: 1906.07906}\BibitemShut {NoStop}%
\bibitem [{\citenamefont {Toth}\ \emph {et~al.}(2019)\citenamefont {Toth},
  \citenamefont {Rezende}, \citenamefont {Jaegle}, \citenamefont {Racanière},
  \citenamefont {Botev},\ and\ \citenamefont
  {Higgins}}]{toth_hamiltonian_2019}%
  \BibitemOpen
  \bibfield  {author} {\bibinfo {author} {\bibfnamefont {P.}~\bibnamefont
  {Toth}}, \bibinfo {author} {\bibfnamefont {D.~J.}\ \bibnamefont {Rezende}},
  \bibinfo {author} {\bibfnamefont {A.}~\bibnamefont {Jaegle}}, \bibinfo
  {author} {\bibfnamefont {S.}~\bibnamefont {Racanière}}, \bibinfo {author}
  {\bibfnamefont {A.}~\bibnamefont {Botev}},\ and\ \bibinfo {author}
  {\bibfnamefont {I.}~\bibnamefont {Higgins}},\ }\bibfield  {title} {\bibinfo
  {title} {Hamiltonian {Generative} {Networks}},\ }\href
  {http://arxiv.org/abs/1909.13789} {\bibfield  {journal} {\bibinfo  {journal}
  {arXiv:1909.13789 [cs, stat]}\ } (\bibinfo {year} {2019})},\ \bibinfo {note}
  {arXiv: 1909.13789}\BibitemShut {NoStop}%
\bibitem [{\citenamefont {Choudhary}\ \emph {et~al.}(2019)\citenamefont
  {Choudhary}, \citenamefont {Lindner}, \citenamefont {Holliday}, \citenamefont
  {Miller}, \citenamefont {Sinha},\ and\ \citenamefont
  {Ditto}}]{choudhary_physics_2019}%
  \BibitemOpen
  \bibfield  {author} {\bibinfo {author} {\bibfnamefont {A.}~\bibnamefont
  {Choudhary}}, \bibinfo {author} {\bibfnamefont {J.~F.}\ \bibnamefont
  {Lindner}}, \bibinfo {author} {\bibfnamefont {E.~G.}\ \bibnamefont
  {Holliday}}, \bibinfo {author} {\bibfnamefont {S.~T.}\ \bibnamefont
  {Miller}}, \bibinfo {author} {\bibfnamefont {S.}~\bibnamefont {Sinha}},\ and\
  \bibinfo {author} {\bibfnamefont {W.~L.}\ \bibnamefont {Ditto}},\ }\bibfield
  {title} {\bibinfo {title} {Physics enhanced neural networks predict order and
  chaos},\ }\href {http://arxiv.org/abs/1912.01958} {\bibfield  {journal}
  {\bibinfo  {journal} {arXiv:1912.01958 [physics]}\ } (\bibinfo {year}
  {2019})},\ \bibinfo {note} {arXiv: 1912.01958}\BibitemShut {NoStop}%
\bibitem [{\citenamefont {Finzi}\ \emph {et~al.}(2020)\citenamefont {Finzi},
  \citenamefont {Wang},\ and\ \citenamefont {Wilson}}]{finzi_simplifying_2020}%
  \BibitemOpen
  \bibfield  {author} {\bibinfo {author} {\bibfnamefont {M.}~\bibnamefont
  {Finzi}}, \bibinfo {author} {\bibfnamefont {K.~A.}\ \bibnamefont {Wang}},\
  and\ \bibinfo {author} {\bibfnamefont {A.~G.}\ \bibnamefont {Wilson}},\
  }\bibfield  {title} {\bibinfo {title} {Simplifying {Hamiltonian} and
  {Lagrangian} {Neural} {Networks} via {Explicit} {Constraints}},\ }\href
  {http://arxiv.org/abs/2010.13581} {\bibfield  {journal} {\bibinfo  {journal}
  {arXiv:2010.13581 [physics, stat]}\ } (\bibinfo {year} {2020})},\ \bibinfo
  {note} {arXiv: 2010.13581}\BibitemShut {NoStop}%
\bibitem [{\citenamefont {Mattheakis}\ \emph {et~al.}(2020)\citenamefont
  {Mattheakis}, \citenamefont {Sondak}, \citenamefont {Dogra},\ and\
  \citenamefont {Protopapas}}]{mattheakis_hamiltonian_2020}%
  \BibitemOpen
  \bibfield  {author} {\bibinfo {author} {\bibfnamefont {M.}~\bibnamefont
  {Mattheakis}}, \bibinfo {author} {\bibfnamefont {D.}~\bibnamefont {Sondak}},
  \bibinfo {author} {\bibfnamefont {A.~S.}\ \bibnamefont {Dogra}},\ and\
  \bibinfo {author} {\bibfnamefont {P.}~\bibnamefont {Protopapas}},\ }\bibfield
   {title} {\bibinfo {title} {Hamiltonian {Neural} {Networks} for solving
  differential equations},\ }\href {http://arxiv.org/abs/2001.11107} {\bibfield
   {journal} {\bibinfo  {journal} {arXiv:2001.11107 [physics]}\ } (\bibinfo
  {year} {2020})},\ \bibinfo {note} {arXiv: 2001.11107}\BibitemShut {NoStop}%
\end{thebibliography}
\end{document}



\appendix
\section{Training Regime}
\begin{table}[htb]
  \centering
  \caption{Training and Testing Parameters}
  \resizebox{\textwidth}{!}{

    \begin{tabular}{c|c|c|c|c|c|c|c|c}
          & Tmax Train & Tmax Test & $\Delta t$ & Samples & Training Points & Total Samples & Hidden Layers & Nodes per Layer \\
    Mass Spring & 3     & 9     & 0.1   & 30    & 25    & 750   & 2     & 200 \\
    Pendulum & 3     & 9     & 0.1   & 30    & 25    & 750   & 2     & 200 \\
    2-Body Gravitational & 20    & 60    & 0.1   & 200   & 20    & 4000  & 2     & 300 \\
    3-Body Gravitational & 2     & 3     & 0.1   & 20    & 200   & 4000  & 2     & 300 \\
    5 Spring Particle & 4     & 12    & 0.1   & 40    & 100   & 4000  & 2     & 300 \\
    Heinon Heiles & 2     & 3     & 0.1   & 20    & 100   & 2000  & 2     & 300 \\
    \end{tabular}%
  \label{tab:addlabel}%
  }
\end{table}%

\begin{table}[htb]
  \centering
  \caption{Parameter Sweep}
  \resizebox{\textwidth}{!}{
    \begin{tabular}{c|c|c|c|c|c|c}
    System & Model & Embedded Graph & Embedded Integrator & Integrator Type & Range & Noise \\
    Mass Spring & Baseline & Yes   & Yes   & RK1   & 1-step & Yes \\
    Pendulum & Hamiltonian & No    & No    & RK2   & 5-step & No \\
    2-Body Gravitational & Potential &       &       & RK3   & 10-step &  \\
    3-Body Gravitational &       &       &       & RK4   &       &  \\
    5 Spring Particle &       &       &       & VI1   &       &  \\
    Heinon Heiles &       &       &       & VI2   &       &  \\
          &       &       &       & VI3   &       &  \\
          &       &       &       & VI4   &       &  \\
    \end{tabular}%
  \label{tab:addlabel}%
  }
\end{table}%

\clearpage
\section{Symplecticity}

\subsection{Symplectic Runge-Kutta}
A Runge-Kutta method can typically be represented by a Butcher Tableau which outlines the coefficients needed for an s-stage integrator.

\begin{table}[htb]
    \centering
    \begin{tabular}{c|ccc}
         $c_1$ & $a_{11}$ & \ldots & $a_{1s}$  \\
         \vdots &  \vdots &  & \vdots \\
         $c_s$ & $a_{s1}$ & \ldots & $a_{ss}$ \\
         \hline 
          & $b_1$ & \ldots & $b_s$ \\
    \end{tabular}
    \caption{Butcher Tableau}
    \label{table.butcher}
\end{table}

where:

$$ k_i = f(t_0 + c_i h, y_0 +h\sum_{j=1}^s a_{ij}k_j), ~~ i=1,...,s$$

$$ y_1 = y_0 + h\sum_{i=1}^s b_i k_i.$$

If the $a_{ij}$ consists entirely of non-zero coefficients then the method is implicit. However, if the method is only lower triangular then the method is considered explicit.

It can be shown that if a RK method preserves quadratic first integrals then it is symplectic. This translates to having coefficients s.t.:

$$ b_ia_{ij} + b_ja_{ji} = b_ib_j \forall i,j = 1,...s. $$

The constraint above enforces the integration scheme to be implicit.

\begin{figure*}[ht]
\centering
        \includegraphics[width=.31\textwidth]{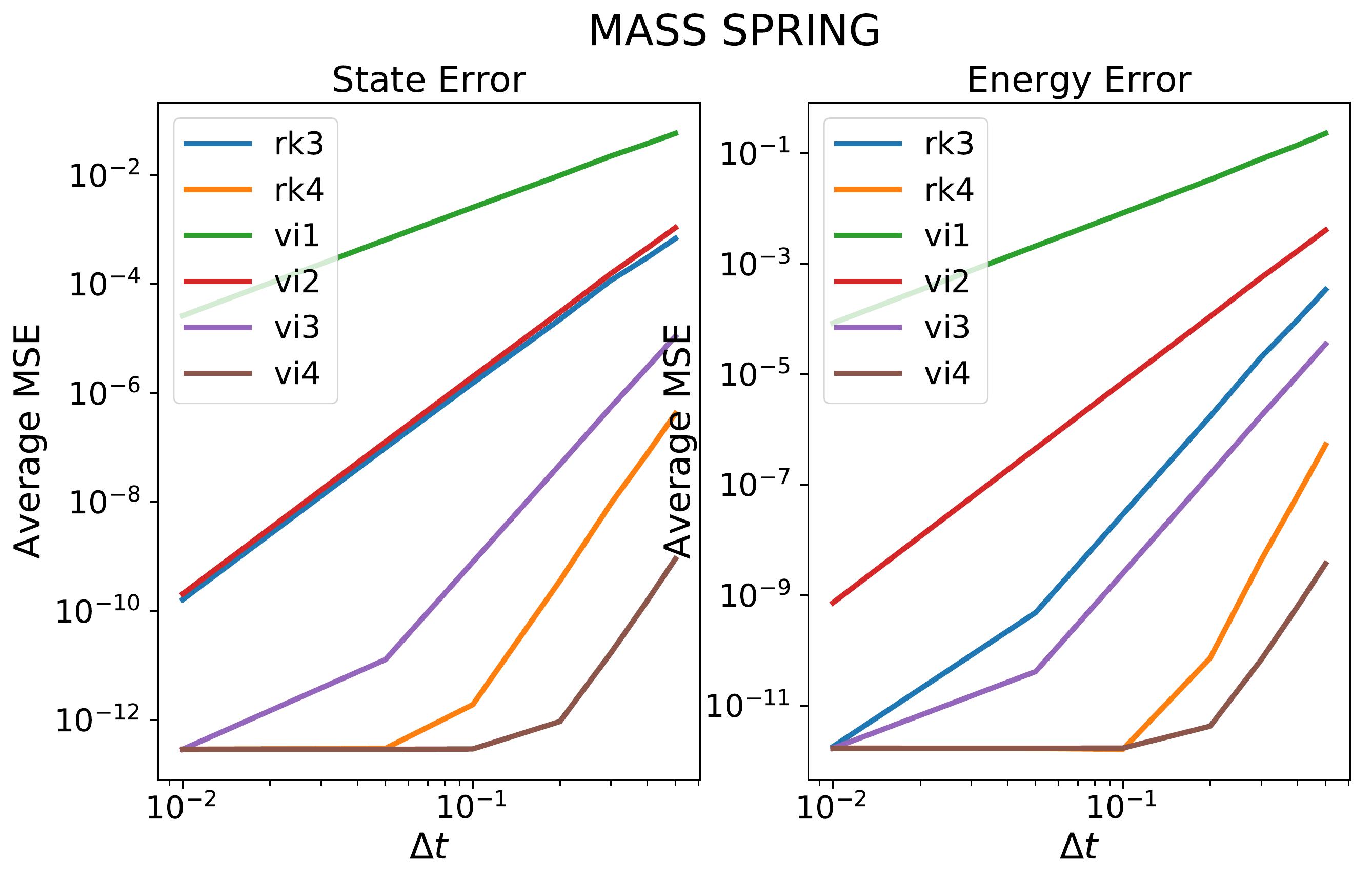}
        \includegraphics[width=.31\textwidth]{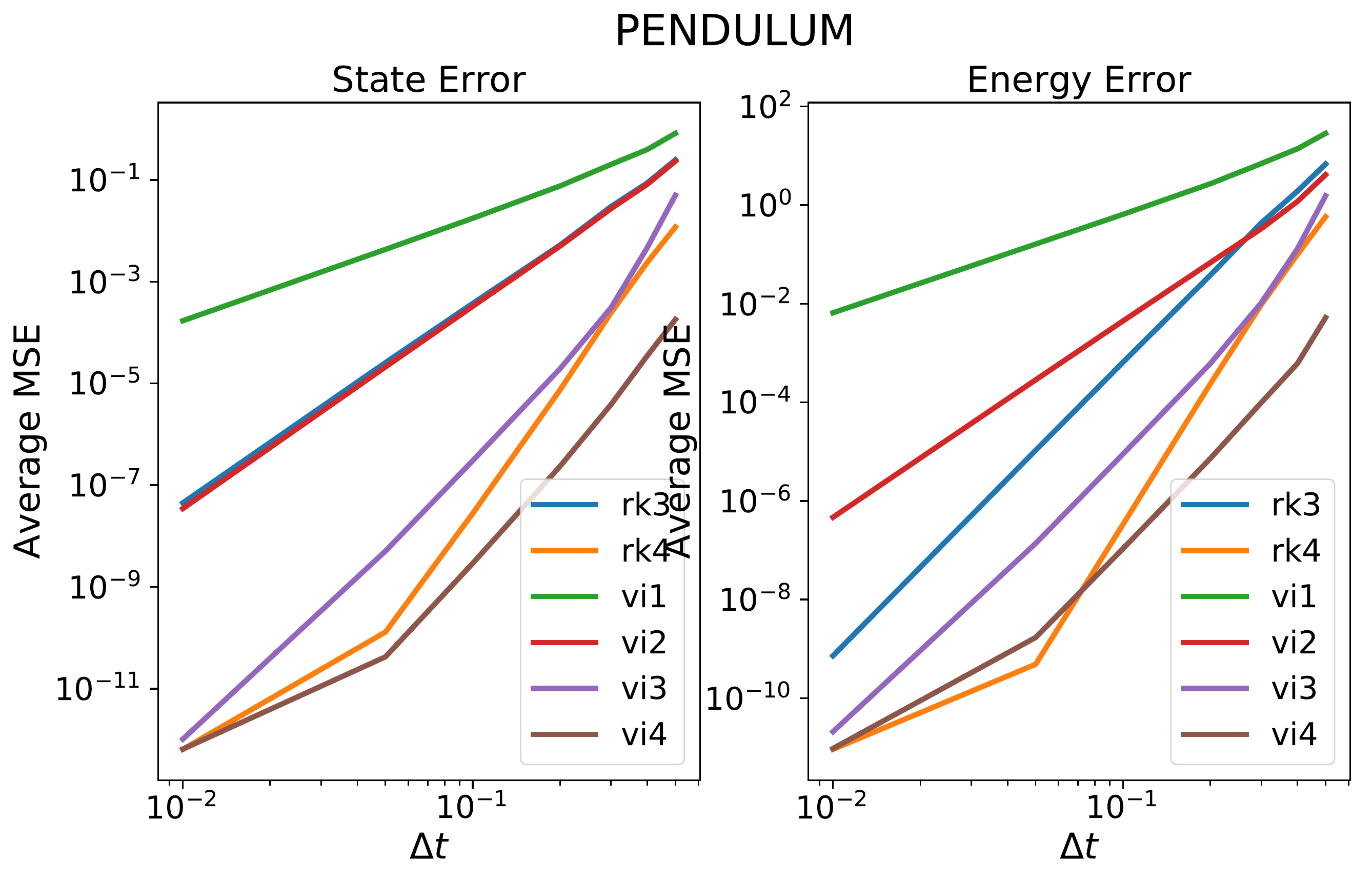}
        \includegraphics[width=.31\textwidth]{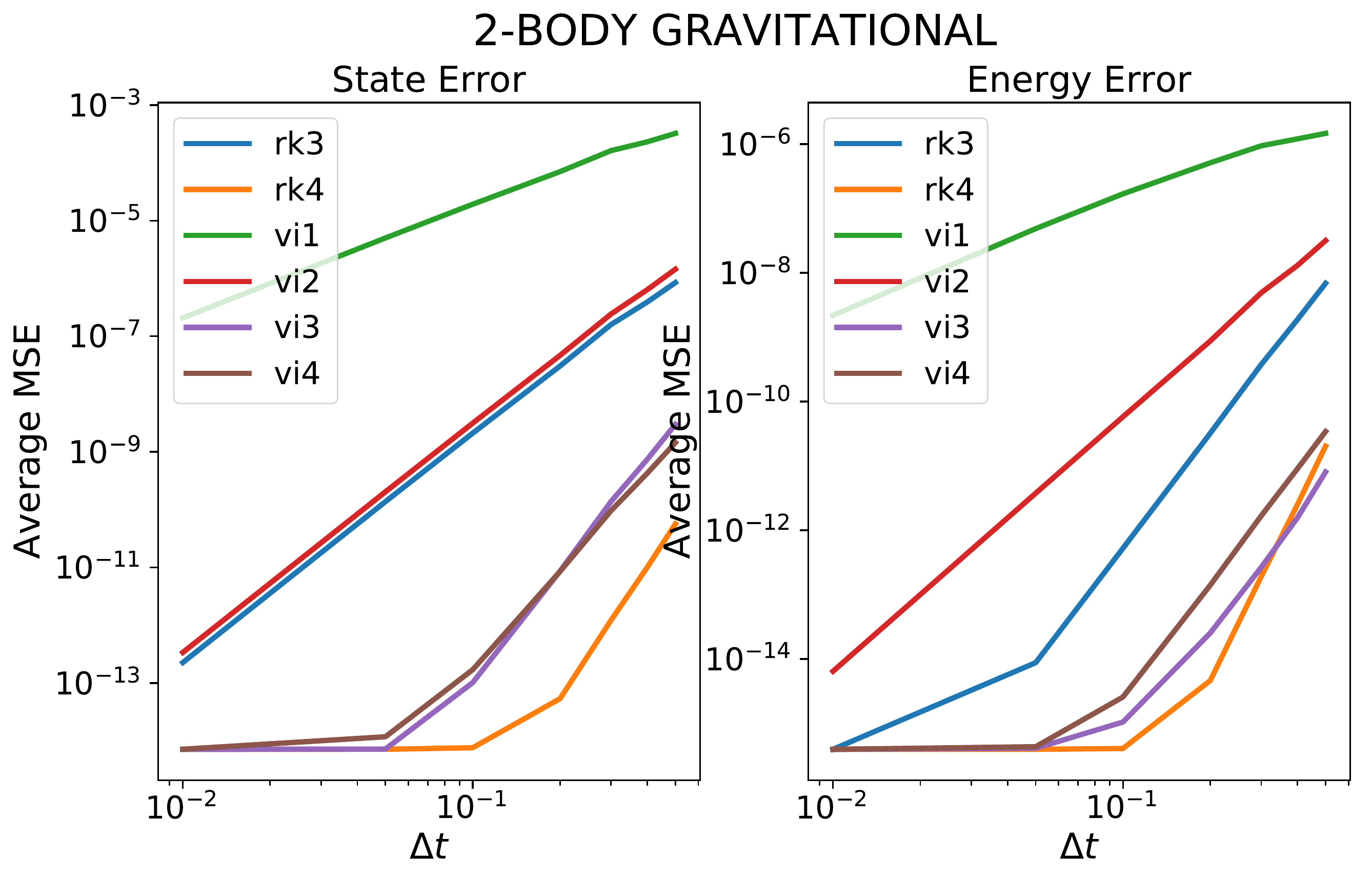}
        \includegraphics[width=.31\textwidth]{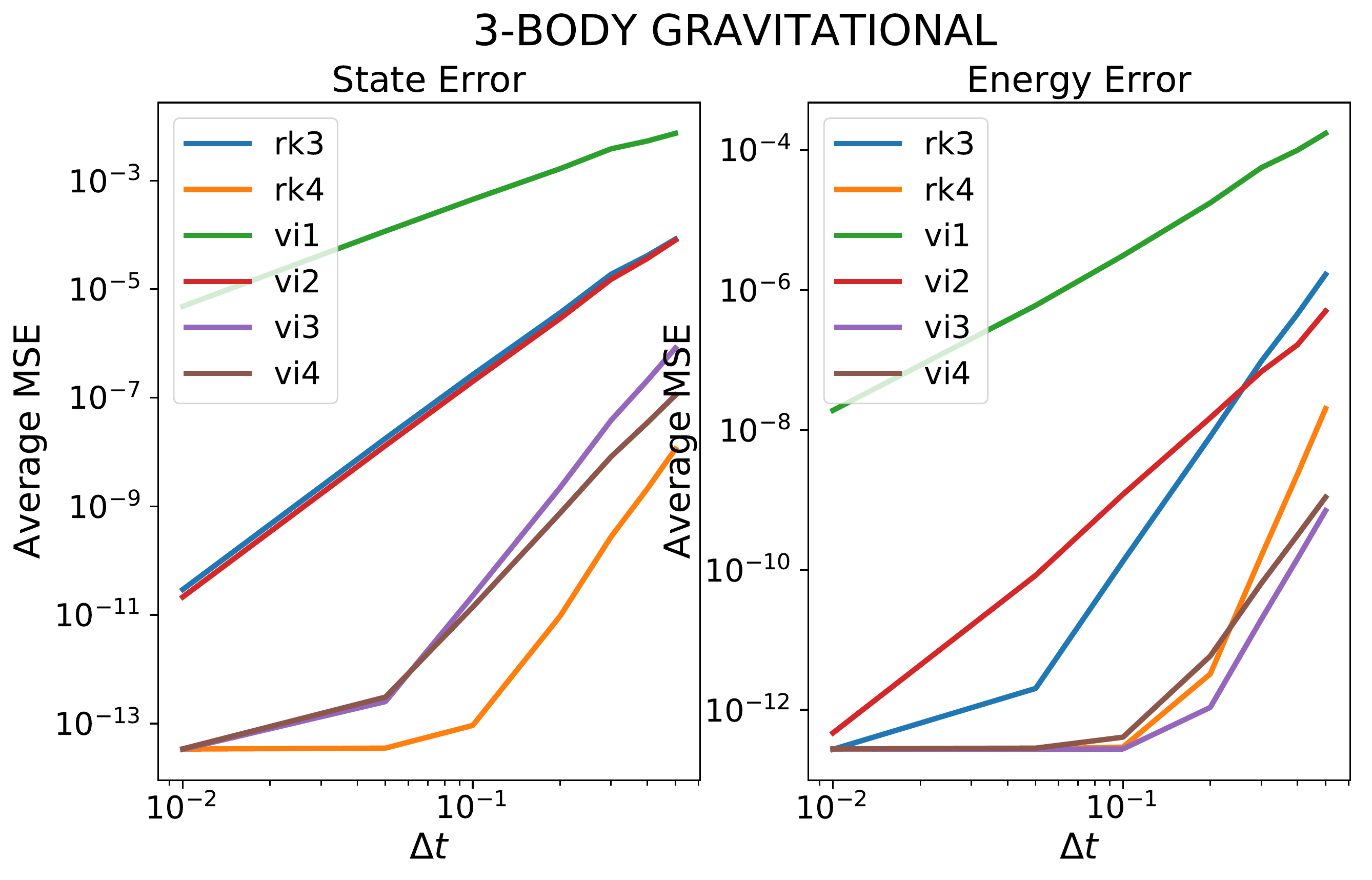}
        \includegraphics[width=.31\textwidth]{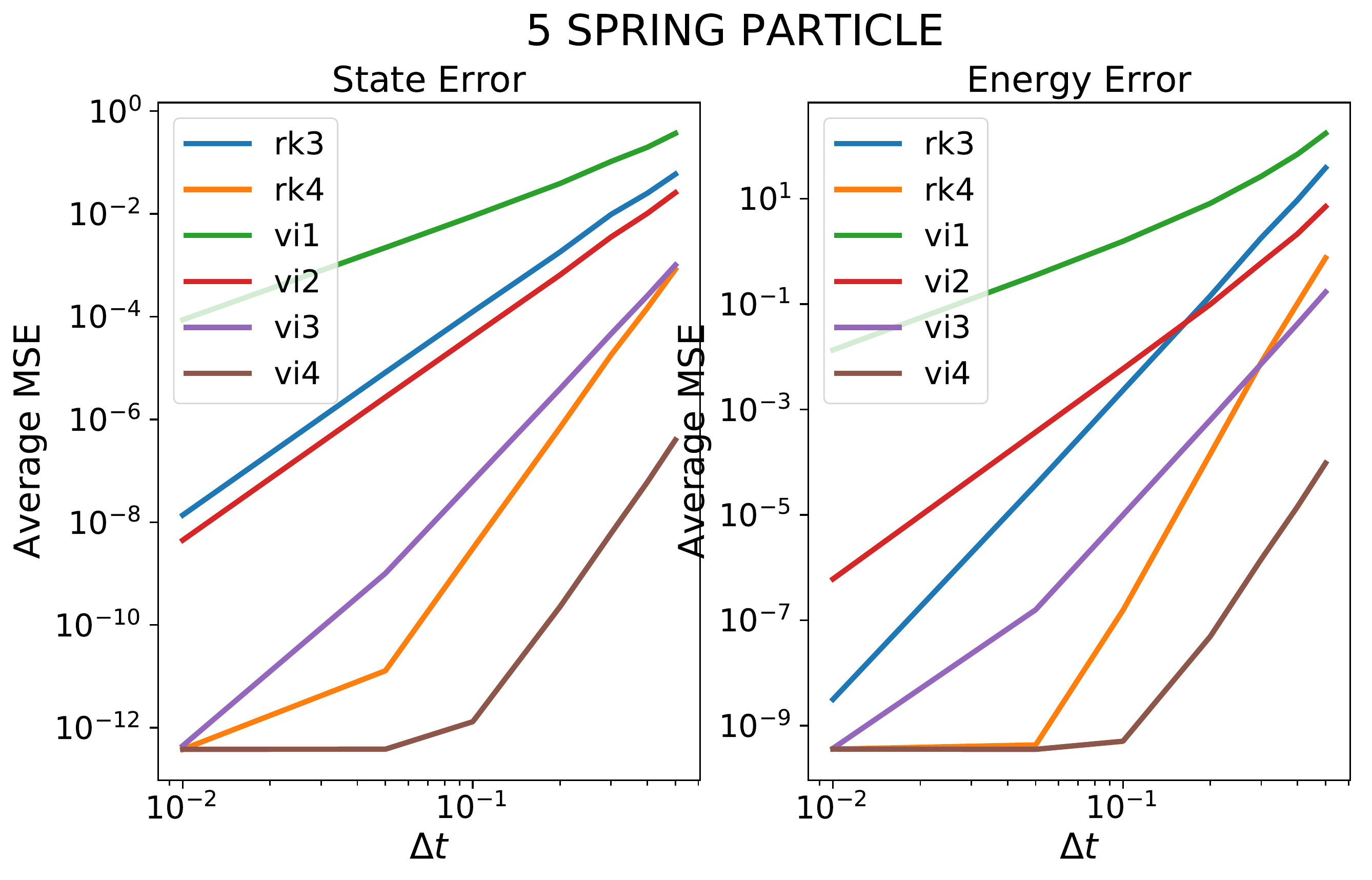}
        \includegraphics[width=.31\textwidth]{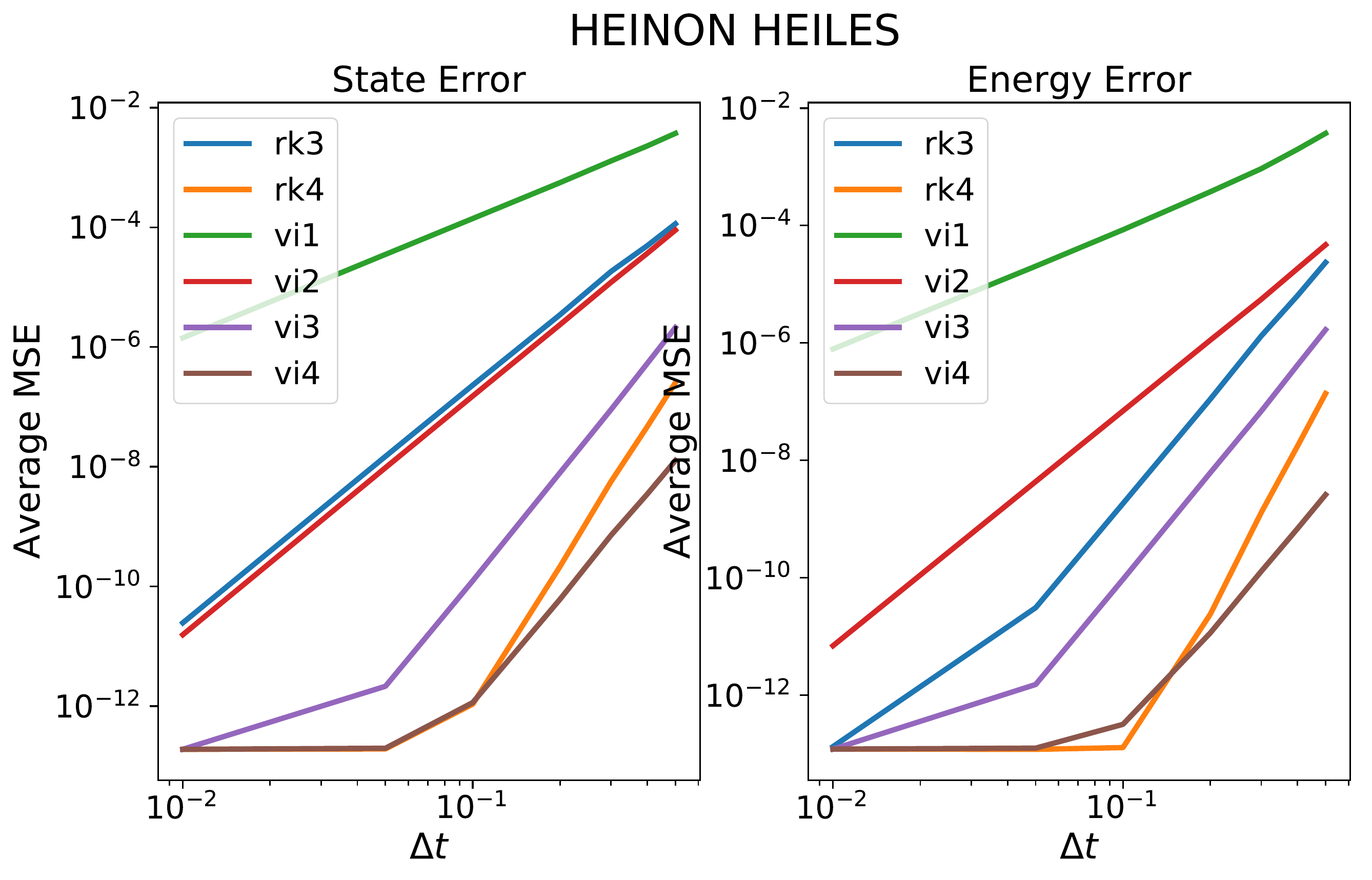}
\caption{We record the state and energy MSE of over 25 initial conditions for different $\Delta t$ values and average them. Each system is integrated to $T_{max}=2$. We see that for some systems a fourth order symplectic integrator performs much better than other integrators with larger $\Delta t$. Note, we exclude the RK1 and RK2 plots as they perform very poorly and distract from the performance differences of the higher order methods.}

\end{figure*}
\begin{figure*}[ht]
	    \includegraphics[width=.31\textwidth]{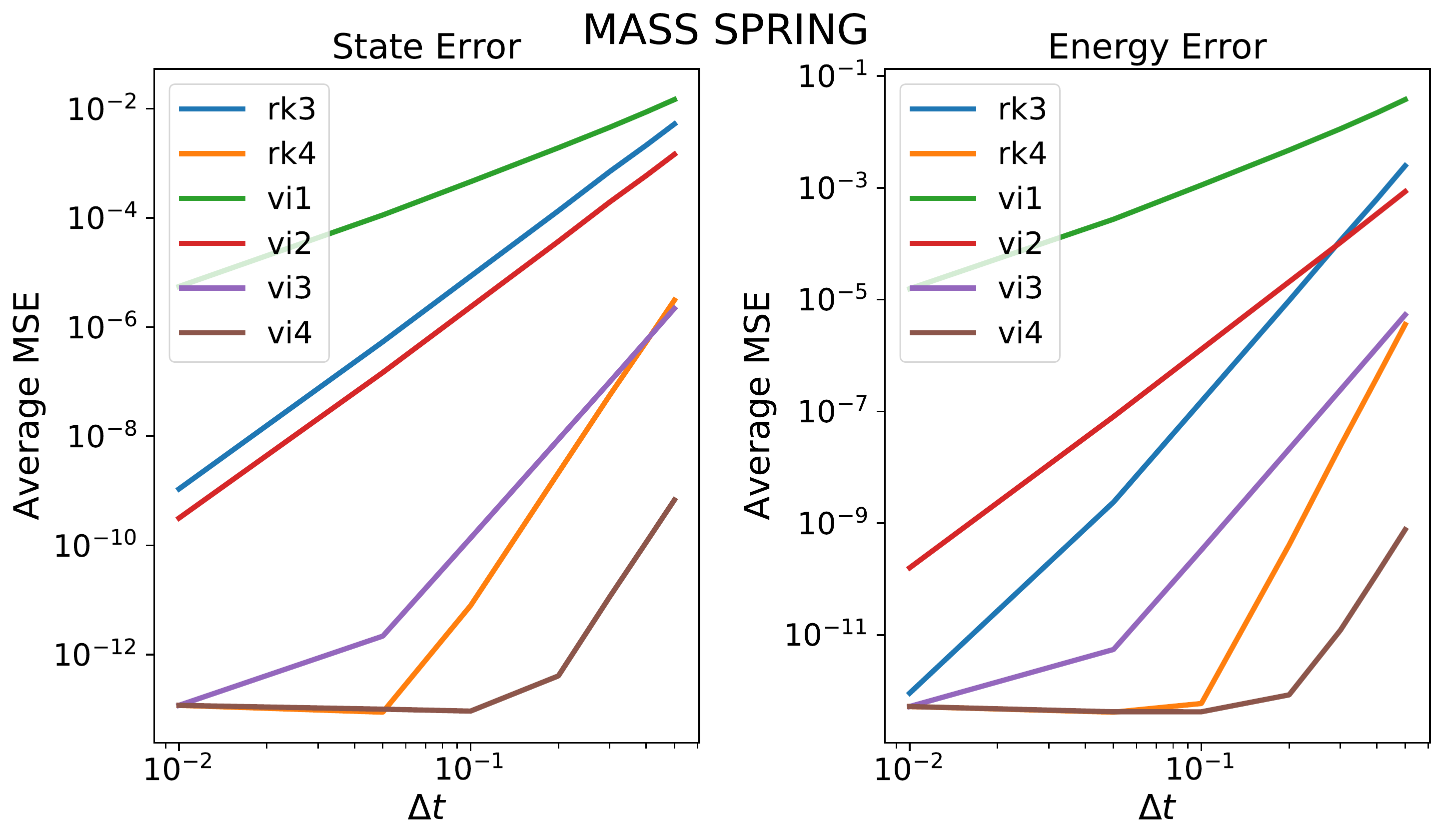}
        \includegraphics[width=.31\textwidth]{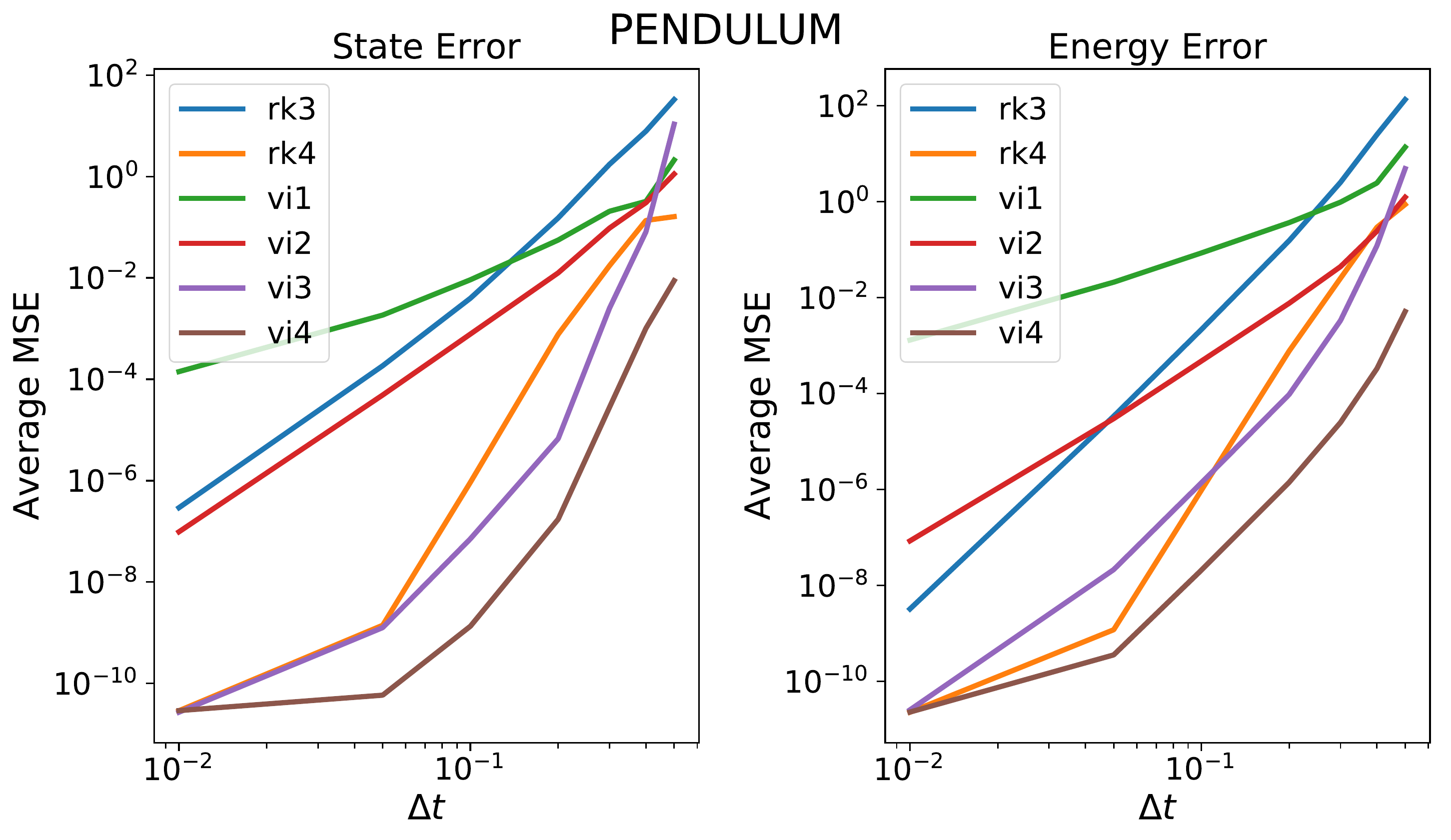}
        \includegraphics[width=.31\textwidth]{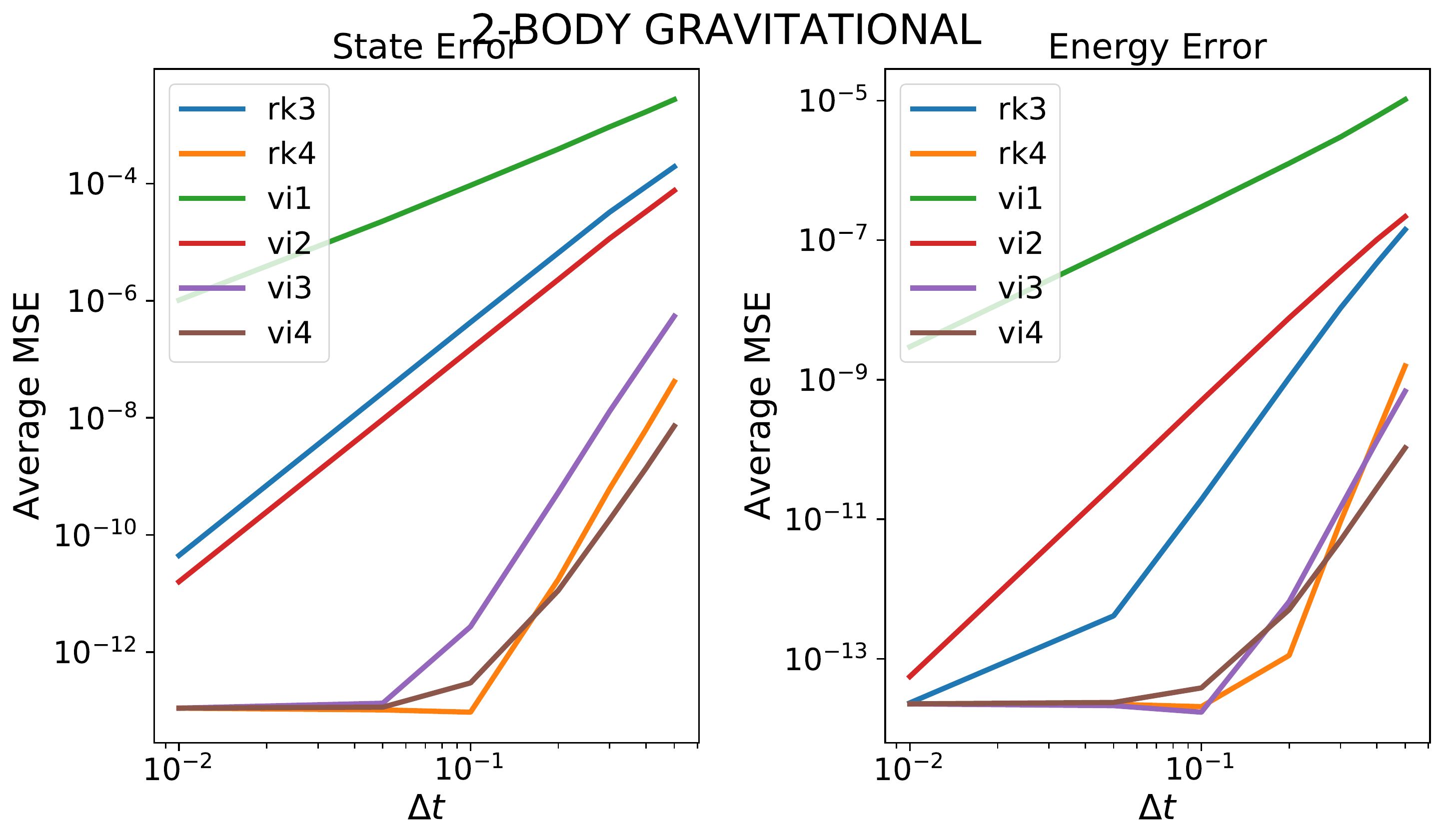}
        \includegraphics[width=.31\textwidth]{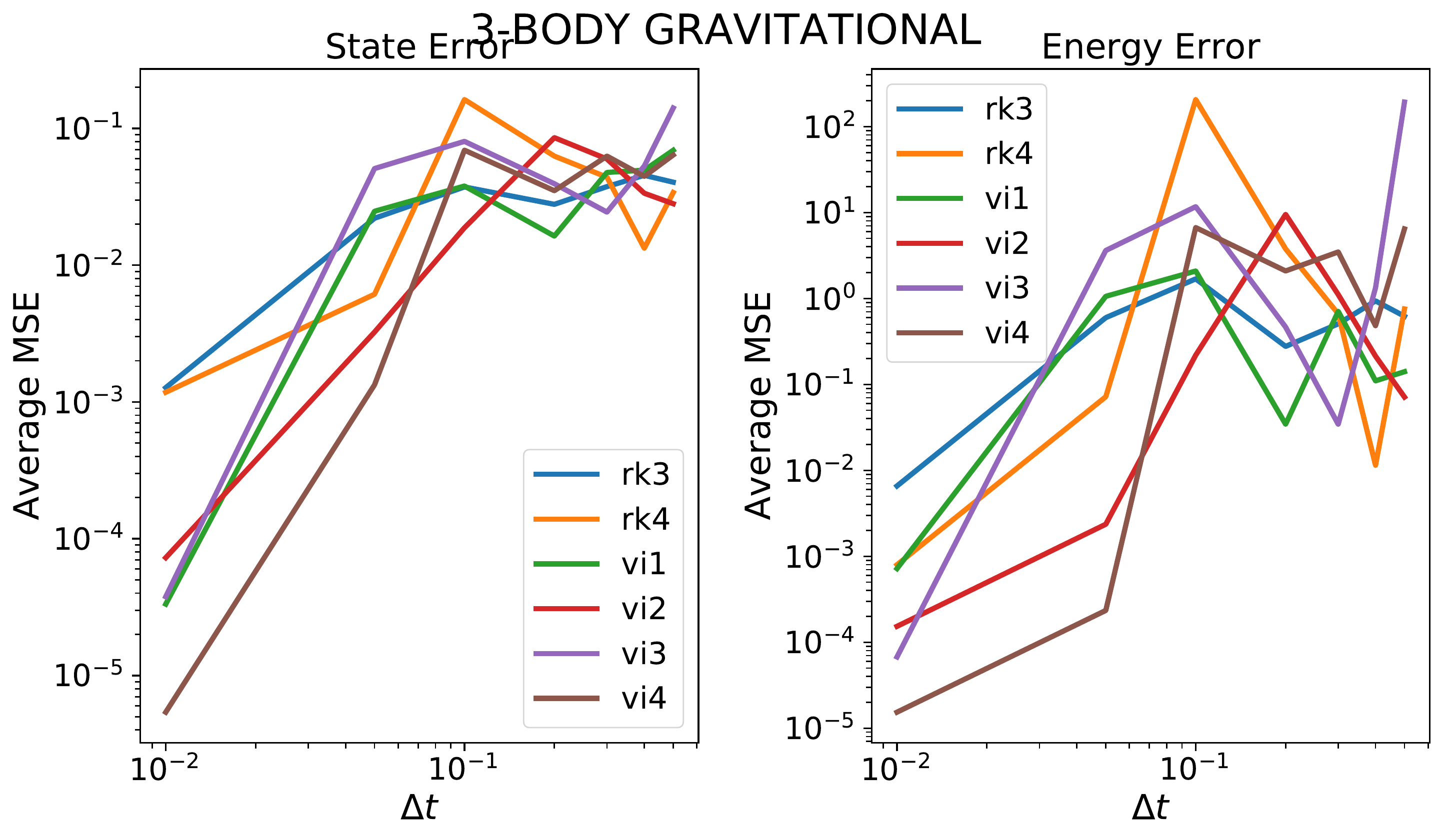}
        \includegraphics[width=.31\textwidth]{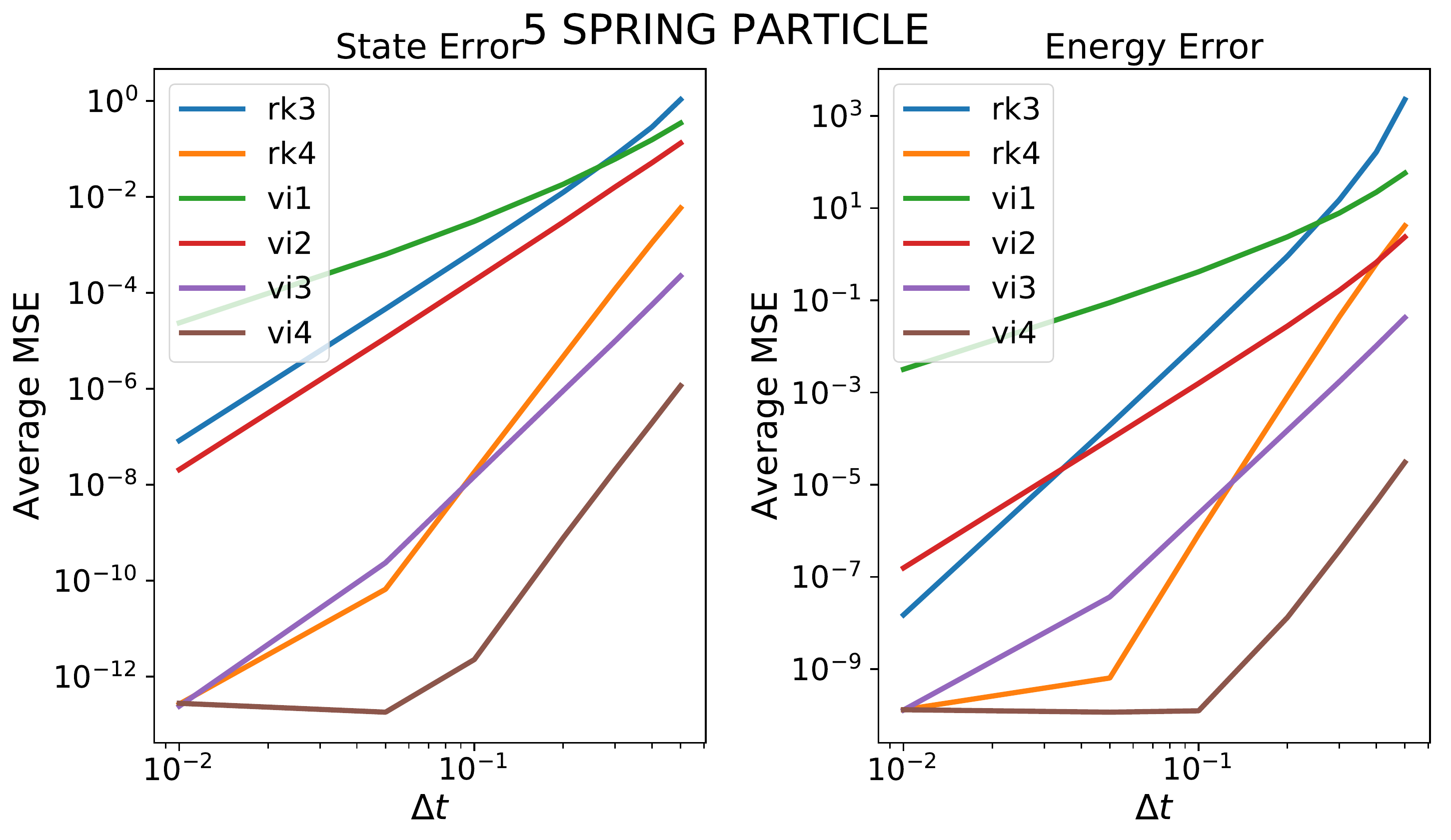}
        \includegraphics[width=.31\textwidth]{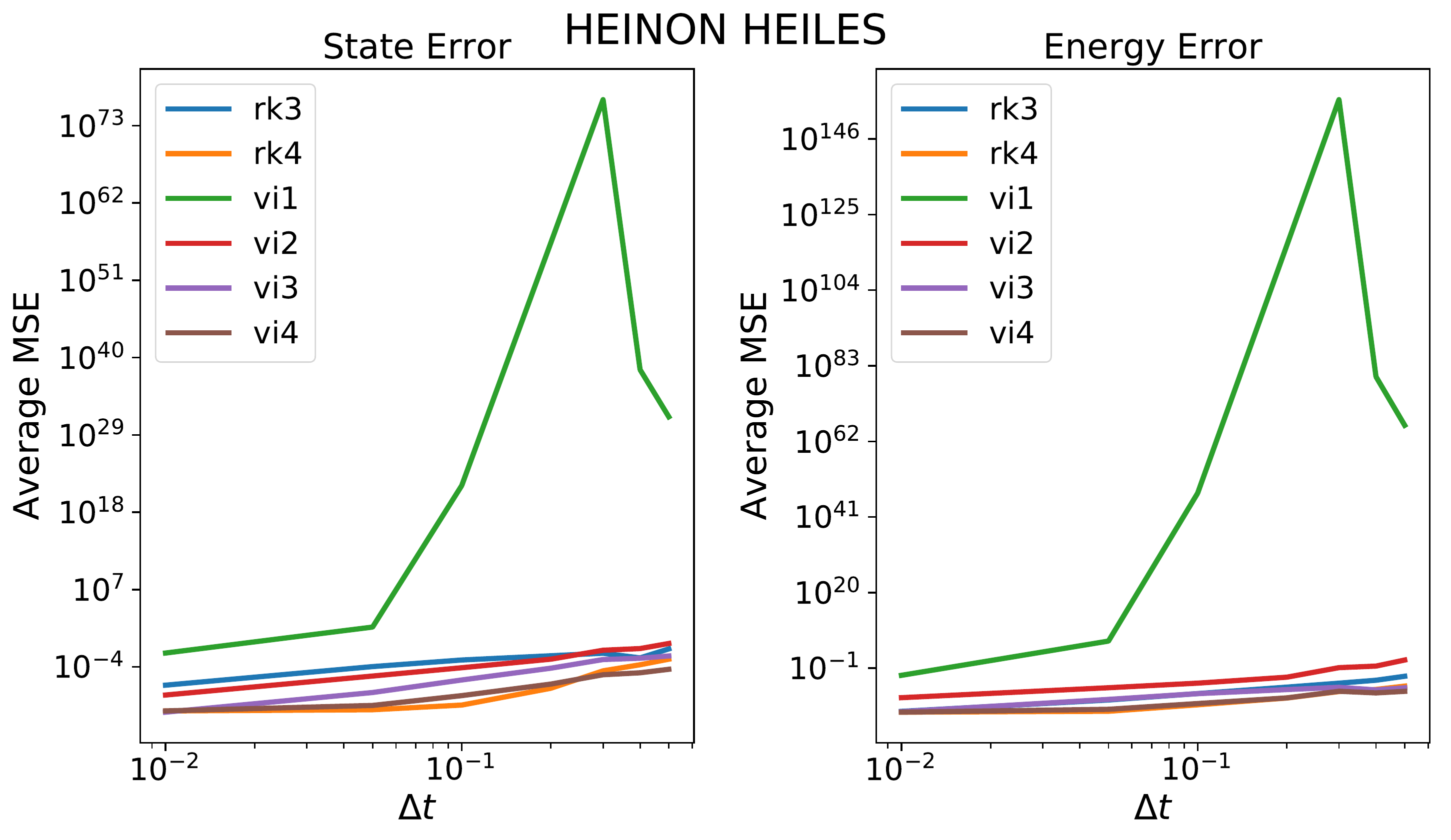}
\caption{We record the state and energy MSE of over 25 initial conditions for different $\Delta t$ values and average them. Each system is integrated to $T_{max}=10$. }
\end{figure*}

\subsection{Variational Integrators}

To achieve high order explicit symplectic integration, we have to use PRK methods. Variational calculus provides one pathway to illustrating the connection between high order symplectic integration and Lagrangians.

Lagrangian mechanics offers an alternative to the Hamiltonian in generalizing a dynamical system. Rather than position and momentum (canonical coordinates) defining the state space, Lagrangian mechanics is defined using a generalized coordinate state space $(\mathbf{q},\dot{\mathbf{q}})$. This is particularly useful in physical settings where the description and measurement of generalized coordinates may be easier to work with than canonical coordinates \cite{marsden_discrete_2001}. Given these coordinates, Joseph-Louis Lagrange showed that a scalar value $\mathcal{A}$, referred to as the action, can be defined as the integral of a Lagrangian, $\mathcal{L}(\mathbf{q},\dot{\mathbf{q}})$:
\begin{equation}
\mathcal{A} = \int_{t}^{t+1} \mathcal{L}(\mathbf{q},\dot{\mathbf{q}}) \mathrm{d}t
\label{eqn.action_integral}
\end{equation}
The integral can be thought as inducing multiple paths between points in state space i.e. multiple walks in the domain of $(\mathbf{q},\mathbf{\dot{q}})$. However, only one path is a stationary state of the action integral. This state lets us move from $t \rightarrow t+1$ with minimal energy. It can be shown, through variational calculus, that this stationary state must satisfy the Euler-Lagrange equation:
\begin{equation}
\frac{\mathrm{d} }{\mathrm{d}t} \left ( \frac{\partial \mathcal{L}}{\partial \dot{\mathbf{q}}} \right )= \frac{\partial \mathcal{L}}{\partial \mathbf{q}}
\label{eqn.euler_lagrange}
\end{equation}
Although complex in form, the action integral and the Euler-Lagrange equations can be discretized and collectively form the basis for variational integrators.

Variational Integrators discretize the action integral $\mathcal{A}$ of the Lagrangian $\mathcal{L}$:
\begin{equation}
\mathcal{L}^d(\mathbf{q}_t,\mathbf{q}_{t+1},h) \approx \int_t^{t+h} \mathcal{L}(\mathbf{q},\mathbf{\dot{q}}) \mathrm{d}\tau
\label{eqn.disc_vi}
\end{equation}

Once discretized, the Euler-Lagrange equations, coupled with the separable Newtonian Lagrangian (Eqn. \ref{eqn.newton}), can be used to obtain the St\"{o}rmer-Verlet equation (Eqn. \ref{stormer_verlet}):
\begin{equation}
\mathcal{L}(\mathbf{q},\mathbf{\dot{q}}) = \mathcal{T}(\mathbf{\dot{q}}) - \mathcal{V}(\mathbf{q}) = \frac{1}{2} \dot{\mathbf{q}}^T M \dot{\mathbf{q}} - \mathcal{V}(\mathbf{q})
\label{eqn.newton}
\end{equation}
\begin{equation}
\mathbf{q}_{t+1} = 2\mathbf{q}_t - \mathbf{q}_{t-1} -h^2 M^{-1} \frac{\partial \mathcal{V}(\mathbf{q})}{\partial \mathbf{q}}
\label{stormer_verlet}
\end{equation}
Notice, the St\"{o}rmer-Verlet equation looks like a discretized second-order differential equation and has a truncation error of $O(h^2)$. The key distinction between this approach and the Hamiltonian is that it allows us to represent information in terms of generalized coordinates and naturally couples the dynamics of momentum with position.

To obtain higher order variational integrators, the work in \cite{marsden_discrete_2001} considers discretizing the Lagrangian by setting the elements of the Lagrangian to polynomials of degree $s$:
\begin{equation}
\mathcal{L}^d(\mathbf{q}_0,\mathbf{q}_{1}) = h \sum_{i=1}^s b_i \mathcal{L}(\mathbf{Q}_i,\dot{\mathbf{Q}_i})
\label{eqn.disc_lag}
\end{equation}
where:
\begin{equation}
\mathbf{Q}_i = \mathbf{q}_0 + h\sum_{j=1}^s a_{ij} \dot{\mathbf{Q_j}} , ~~ 
\mathbf{q}_1 = \mathbf{q}_0 + h\sum_{i=1}^s b_i\dot{\mathbf{Q}_i}
\label{eqn.disc_lag2}
\end{equation}
If we extremize this Lagrangian with respect to $ \dot{\mathbf{Q}}$, \cite{marsden_discrete_2001} show that we obtain a variational integrator which allows us to update position and momentum through the following equations:
\begin{equation}
\mathbf{Q}_i = \mathbf{q}_0 + h\sum_{j=1}^s a_{ij} \dot{\mathbf{Q_j}} , ~~
\mathbf{P}_i = \mathbf{p}_0 + h\sum_{j=1}^s \hat{a}_{ij} \dot{\mathbf{P}_j}
\label{eqn.lag_updat}
\end{equation}
\begin{equation}
\mathbf{q}_1 = \mathbf{q}_0 + h\sum_{i=1}^s b_i\dot{\mathbf{Q}_i}, ~~
\mathbf{p}_1 = \mathbf{p}_0 + h\sum_{i=1}^s b_i \dot{\mathbf{P}_i}
\label{eqn.lag_updat1}
\end{equation}
The result of extremizing the integral in this way results in a set of update rules that can be described by a Partitioned Runge-Kutta (PRK) method. It can also be shown that the equations provide a generalized higher-order form of the St\"{o}rmer-Verlet equation which allows us to couple momentum and position updates with increased accuracy.

\begin{table}[htb]
    \centering
    \begin{tabular}{c|ccc}
         $c_1$ & $a_{11}$ & \ldots & $a_{1s}$  \\
         \vdots &  \vdots &  & \vdots \\
         $c_s$ & $a_{s1}$ & \ldots & $a_{ss}$ \\
         \hline 
          & $b_1$ & \ldots & $b_s$ \\
    \end{tabular}
    ,
    \centering
    \begin{tabular}{c|ccc}
         $\hat{c}_1$ & $\hat{a}_{11}$ & \ldots & $\hat{a}_{1s}$  \\
         \vdots &  \vdots &  & \vdots \\
         $\hat{c}_s$ & $\hat{a}_{s1}$ & \ldots & $\hat{a}_{ss}$ \\
         \hline 
          & $\hat{b}_1$ & \ldots & $\hat{b}_s$ \\
    \end{tabular}
    \caption{PRK Butcher Tableau}
    \label{table.butcher}
\end{table}

\begin{table}[htb]
    \centering
    \begin{tabular}{c|cccc}
         $c_1$ &a1 & 0 & 0 & 0  \\
         $c_2$ &a1 & a2 & 0 & 0  \\
         $c_3$ &a1 & a2 & a3 & 0  \\
         $c_4$ &a1 & a2 & a3 & a4  \\
         \hline 
          & a1 &a2&a3&a4\\
    \end{tabular}
    ,
    \centering
    \begin{tabular}{c|cccc}
        $c_1$ &0 & 0 & 0 & 0  \\
         $c_2$ &d1 &0 & 0 & 0  \\
         $c_3$ &d1 & d2 & 0 & 0  \\
         $c_4$ &d1 & d2 & d3 & 0  \\
         \hline 
          & d1 &d2&d3&d4\\
    \end{tabular}
    \caption{Yoshida 4th order}
    \label{table.butcher_vi4}
\end{table}

The fourth order Yoshida method can be described by PRK coefficients as:

$w1 =\frac{1}{2-2^{1/3}}$ and $w0= \frac{-2^{1/3}}{2-2^{1/3}}$ then, $a1=a4= w1/2$, $a2=a3=(w0+w1)/2$ and$d1=d3=w1$,$d2=w0$ and $d4=0$.

The fourth order Mcate method can be described by PRK coefficients as:
\begin{table}
\begin{tabular}{c|c}
coefficient & value \\
\hline
d1 & 0.515352837431122936\\
d2 & -0.085782019412973646\\
d3 & 0.441583023616466524\\
d4 & 0.128846158365384185\\
a1 &0.134496199277431089\\
a2& -0.224819803079420806\\
a3 & 0.756320000515668291\\
a4 & 0.334003603286321425\\
\end{tabular}
\caption{Mcate 4th order coefficients}
\end{table}

We should clarify that by conserving the 2-form, symplectic integrators are more consistent with the underlying dynamics. However, enforcing the 2-form constraint doesn't always correlate to better state MSE. In other words, the constraint to stay on the 2-form is strong enough to distort the trajectory since the symplectic integrators conserve a perturbed Hamiltonian. Furthermore, the integration time horizon also affects the reported MSE. We find that for small $T_{\max}$ values, high order ($n>4$) RK and VI methods are quite comparable. This re-emphasises the point that symplectic high order methods are good for long range trajectories.

\section{Ablation Results}

\begin{figure*}[h]
\centering
	\begin{subfigure}[b]{0.3\textwidth}
        \includegraphics[width=\textwidth]{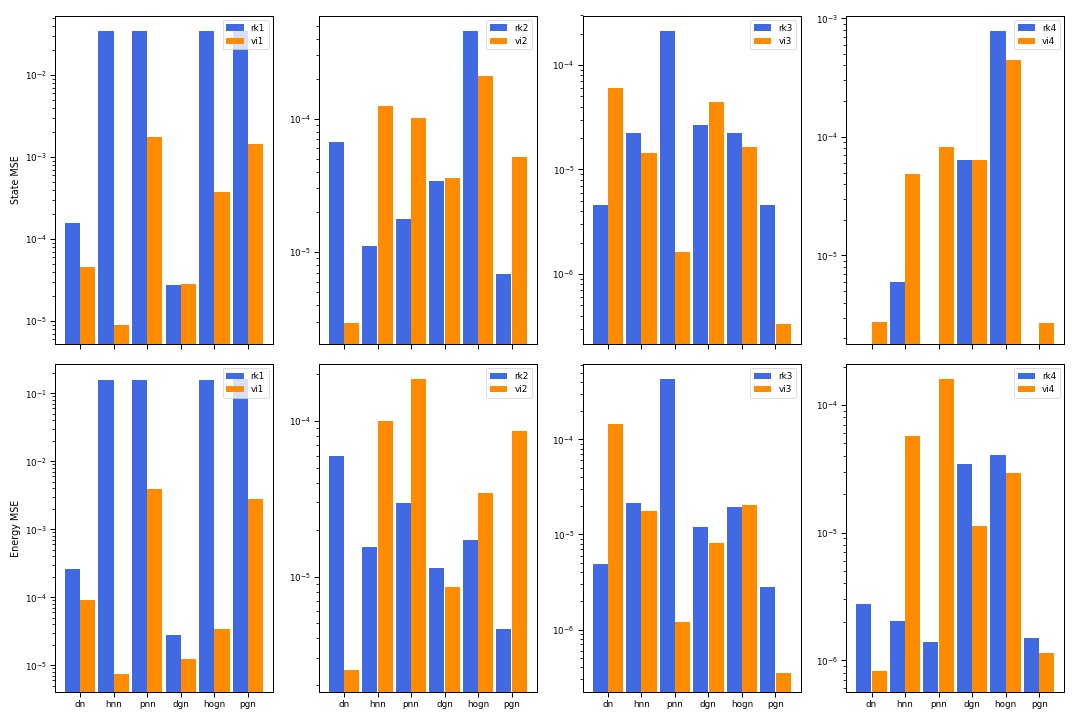}
        \caption{2-step integration}
    \end{subfigure}
	\begin{subfigure}[b]{0.3\textwidth}
        \includegraphics[width=\textwidth]{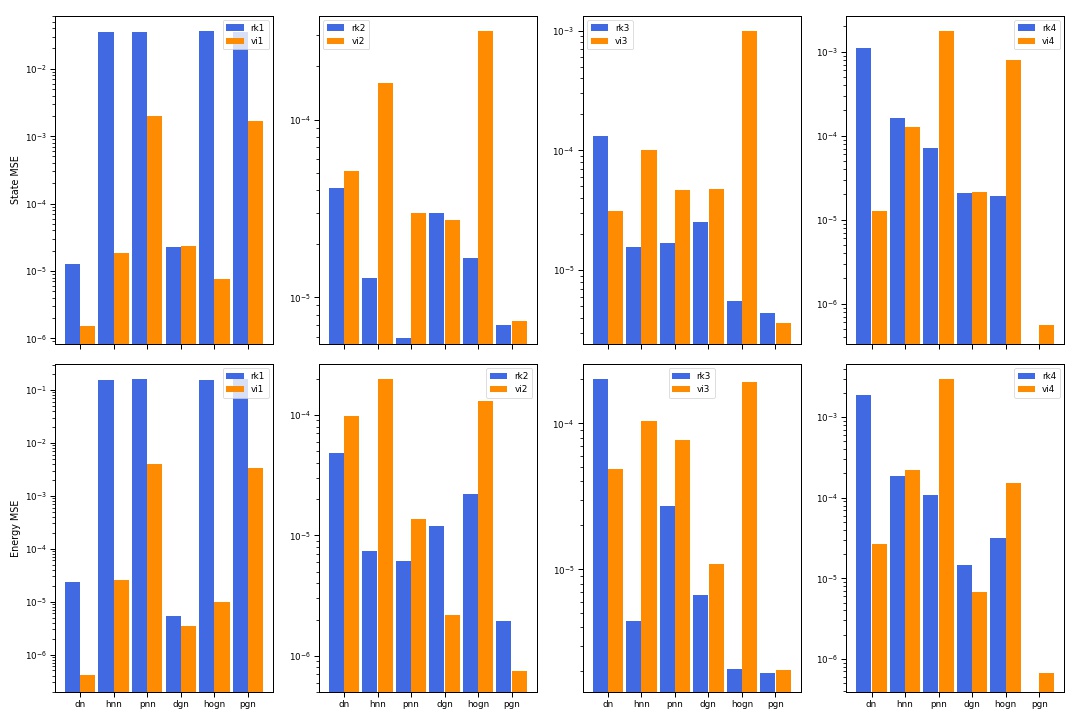}
        \caption{5-step integration}
    \end{subfigure}
	\begin{subfigure}[b]{0.3\textwidth}
        \includegraphics[width=\textwidth]{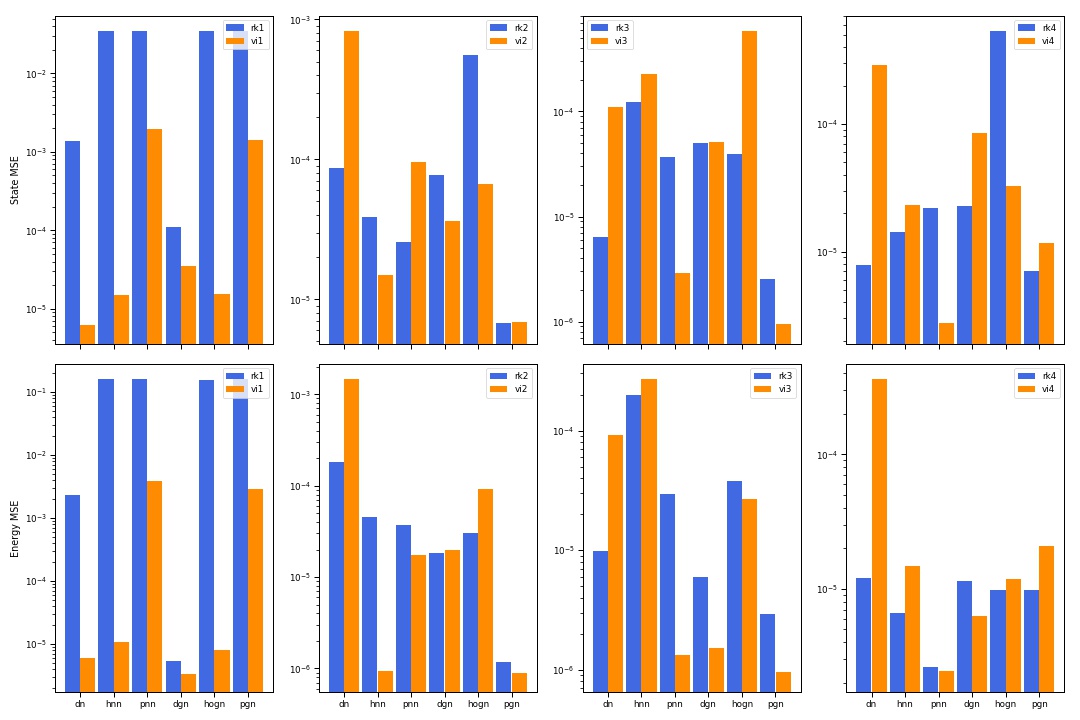}
        \caption{10-step integration}
    \end{subfigure}
\caption{Mass Spring System with noiseless training data. Each bar represents the geometric mean of the MSE of 25 test initial conditions.}
\end{figure*}
\begin{figure*}[htb]
\centering
	\begin{subfigure}[b]{0.3\textwidth}
        \includegraphics[width=\textwidth]{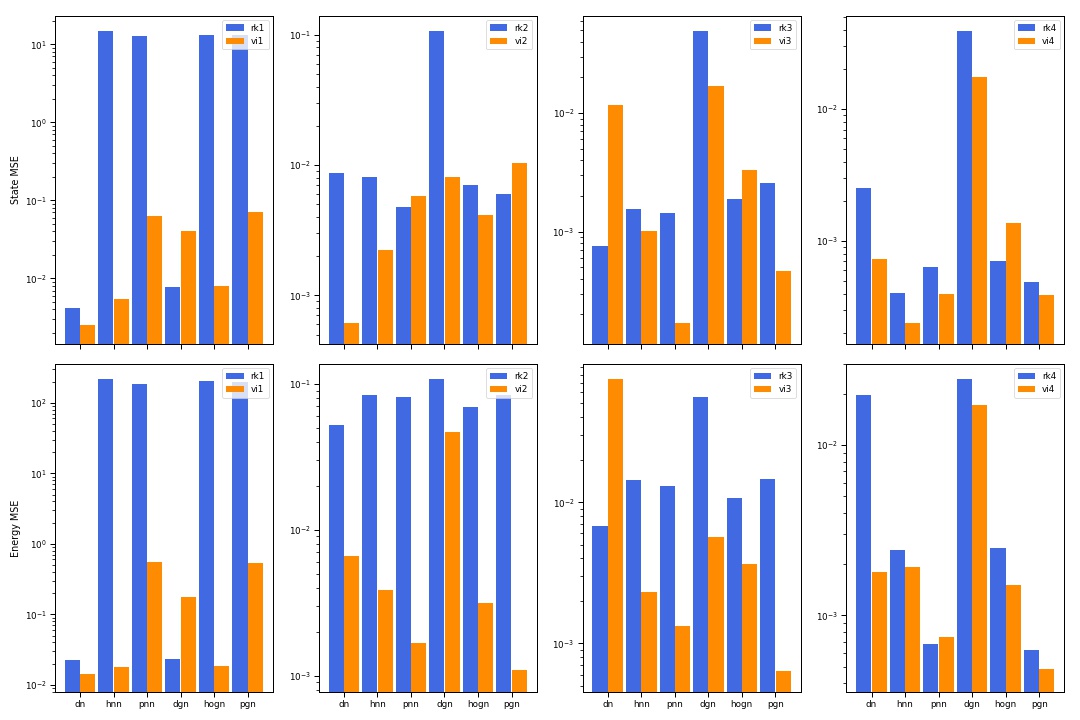}
        \caption{2-step integration}
    \end{subfigure}
	\begin{subfigure}[b]{0.3\textwidth}
        \includegraphics[width=\textwidth]{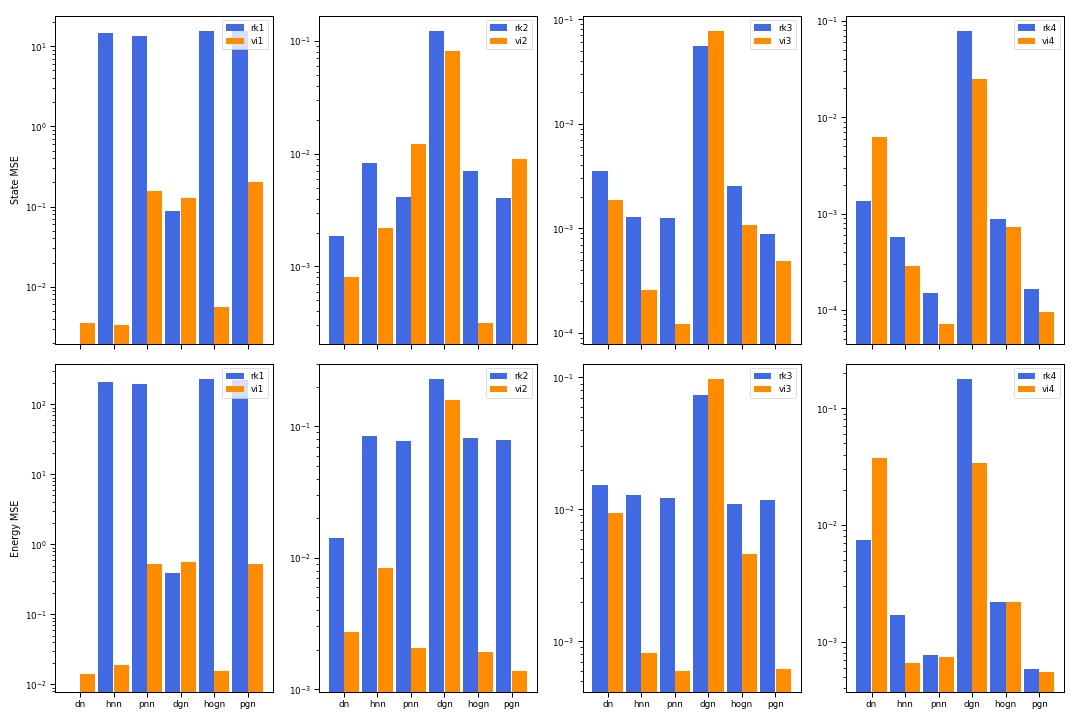}
        \caption{5-step integration}
    \end{subfigure}
	\begin{subfigure}[b]{0.3\textwidth}
        \includegraphics[width=\textwidth]{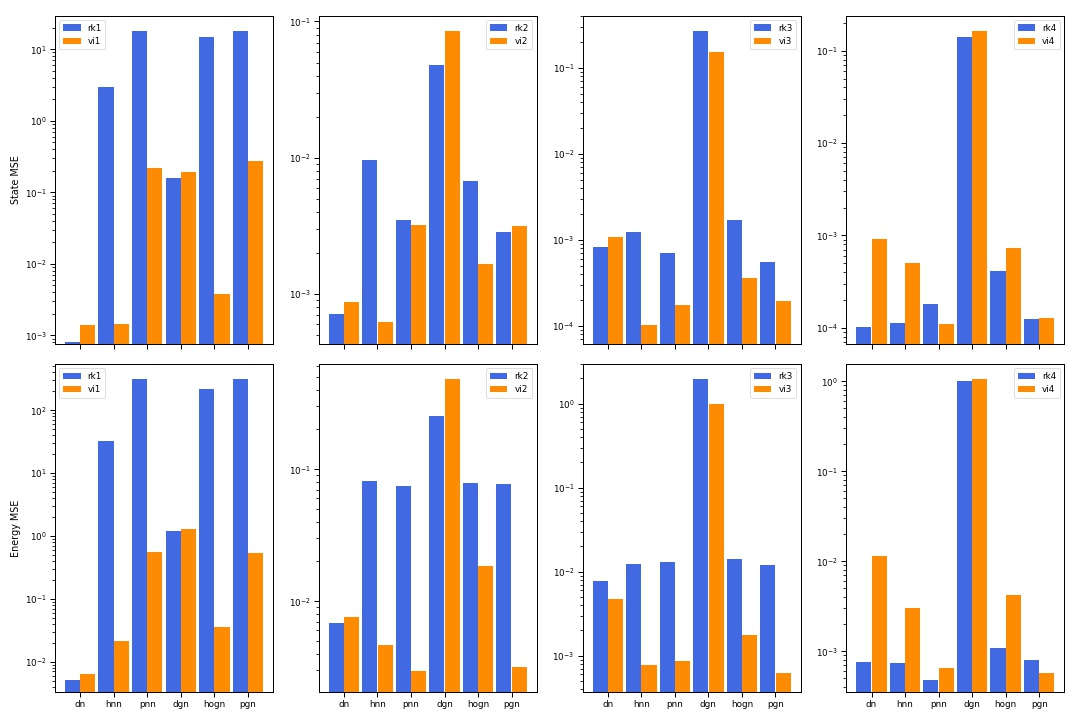}
        \caption{10-step integration}
    \end{subfigure}
\caption{Pendulum with noiseless training data. Each bar represents the geometric mean of the MSE of 25 test initial conditions.}
\end{figure*}
\begin{figure*}[htb]
\centering
	\begin{subfigure}[b]{0.3\textwidth}
        \includegraphics[width=\textwidth]{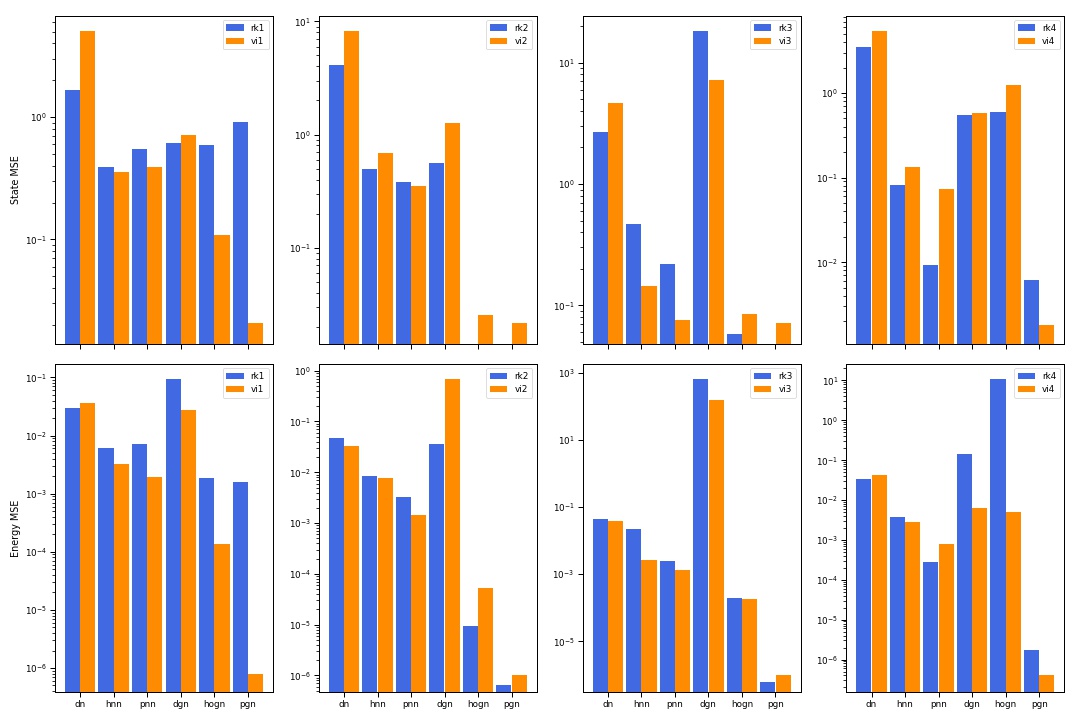}
        \caption{2-step integration}
    \end{subfigure}
	\begin{subfigure}[b]{0.3\textwidth}
        \includegraphics[width=\textwidth]{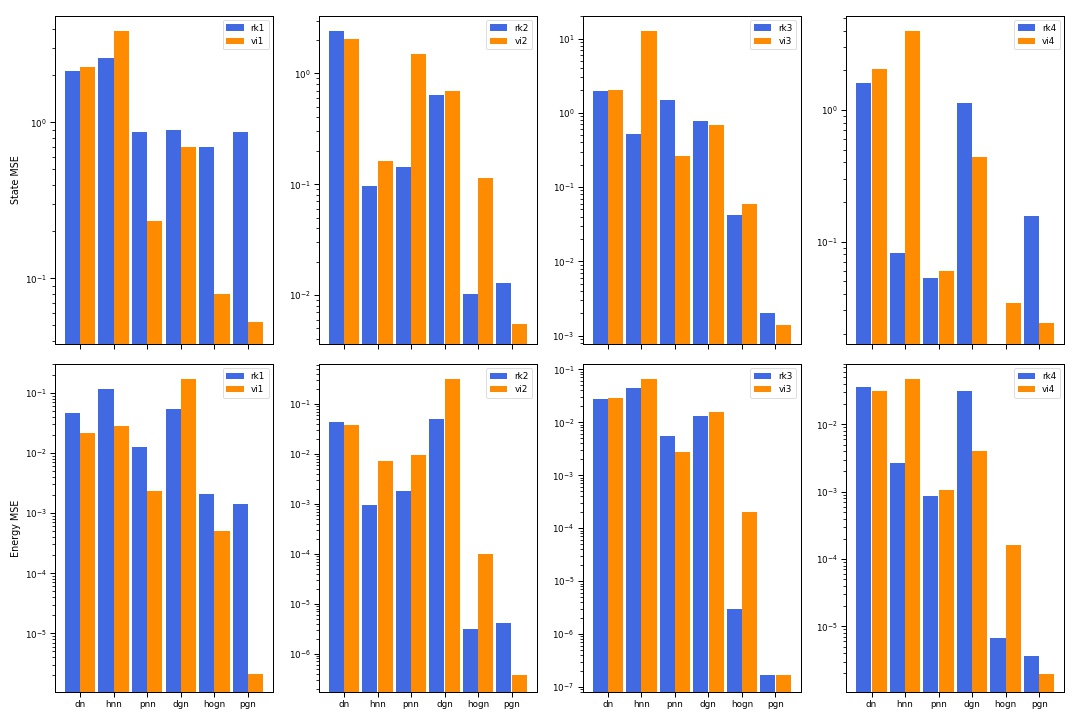}
        \caption{5-step integration}
    \end{subfigure}
	\begin{subfigure}[b]{0.3\textwidth}
        \includegraphics[width=\textwidth]{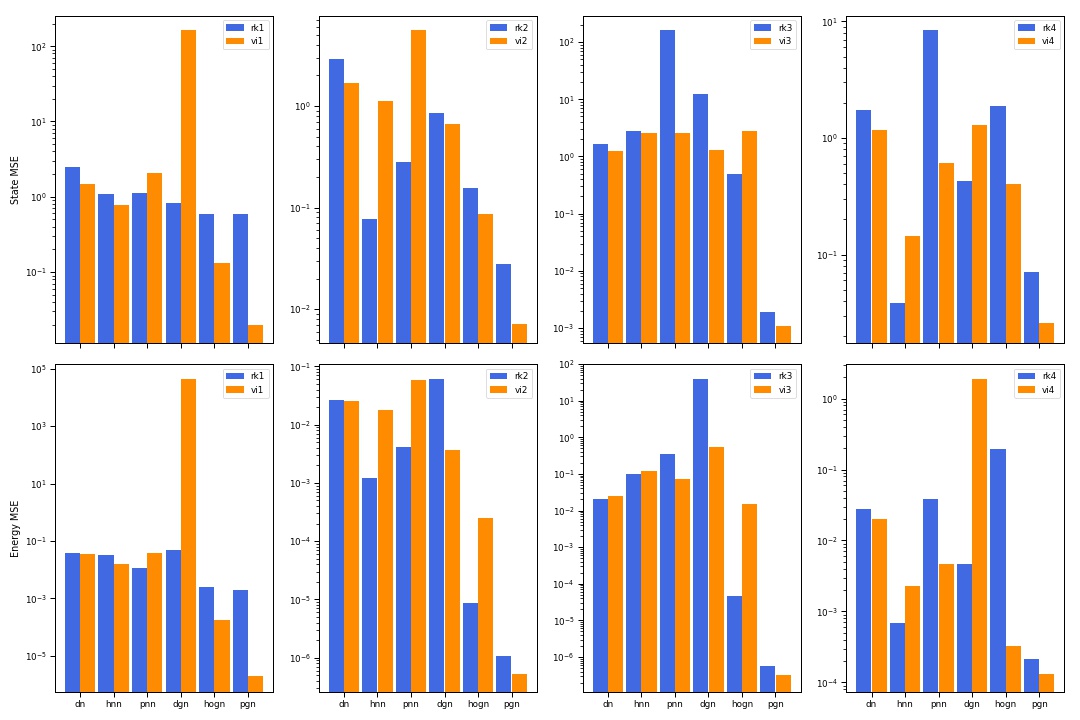}
        \caption{10-step integration}
    \end{subfigure}
\caption{2-Body gravitational system with noiseless training data. Each bar represents the geometric mean of the MSE of 25 test initial conditions.}
\end{figure*}
\begin{figure*}[htb]
\centering
	\begin{subfigure}[b]{0.3\textwidth}
        \includegraphics[width=\textwidth]{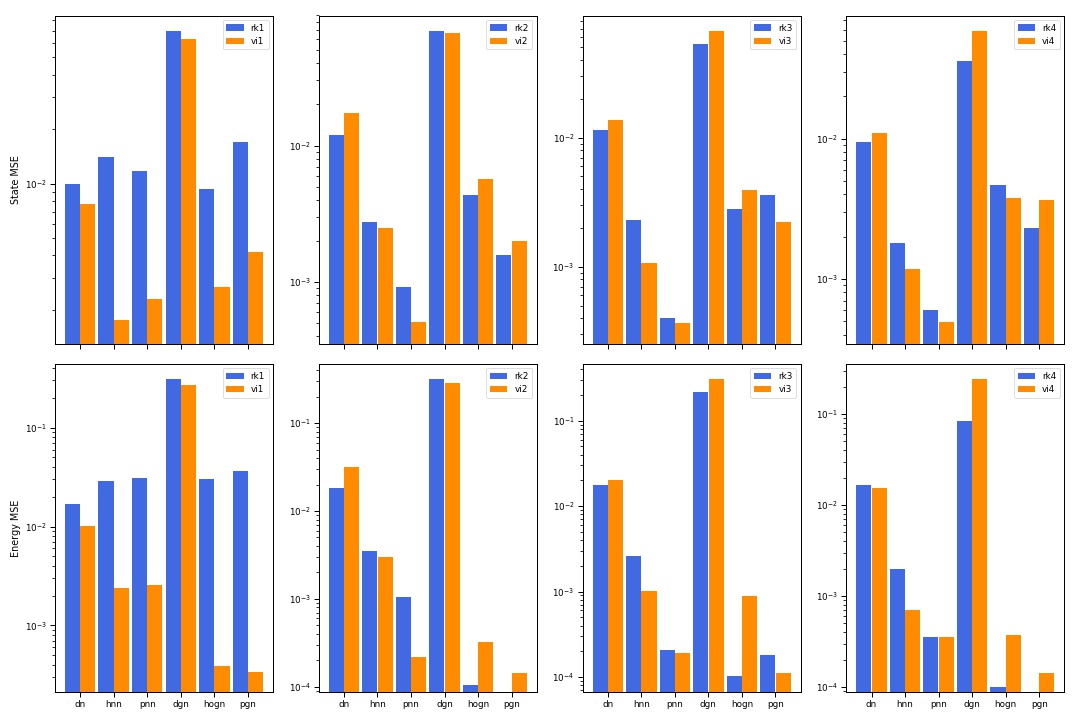}
        \caption{2-step integration}
    \end{subfigure}
	\begin{subfigure}[b]{0.3\textwidth}
        \includegraphics[width=\textwidth]{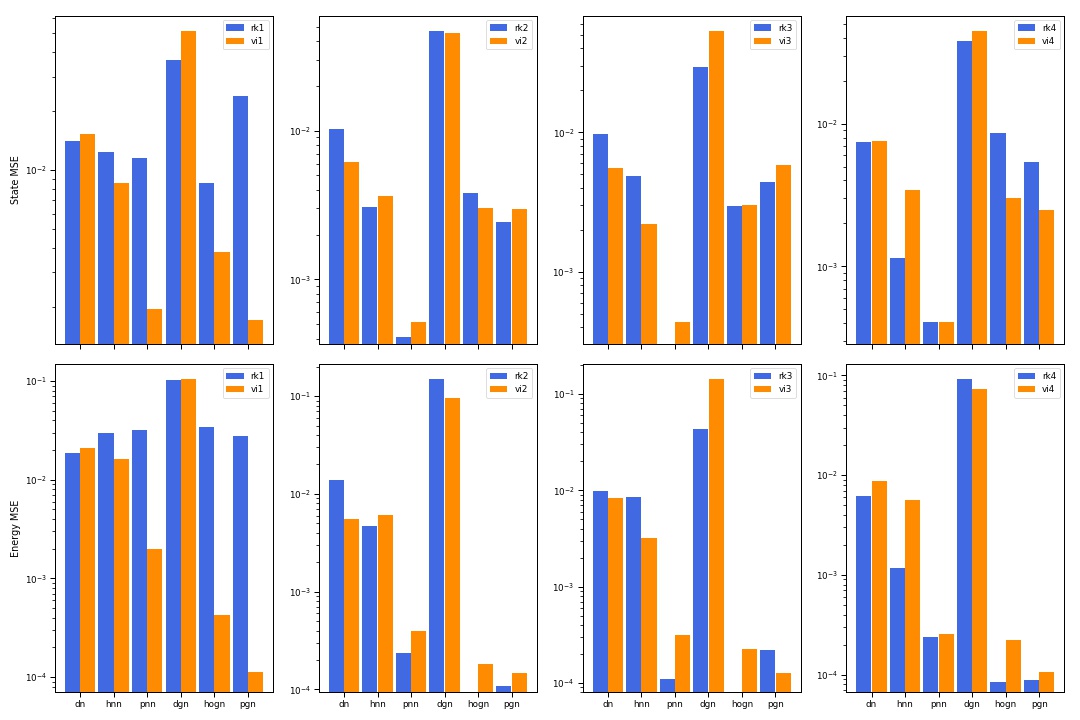}
        \caption{5-step integration}
    \end{subfigure}
	\begin{subfigure}[b]{0.3\textwidth}
        \includegraphics[width=\textwidth]{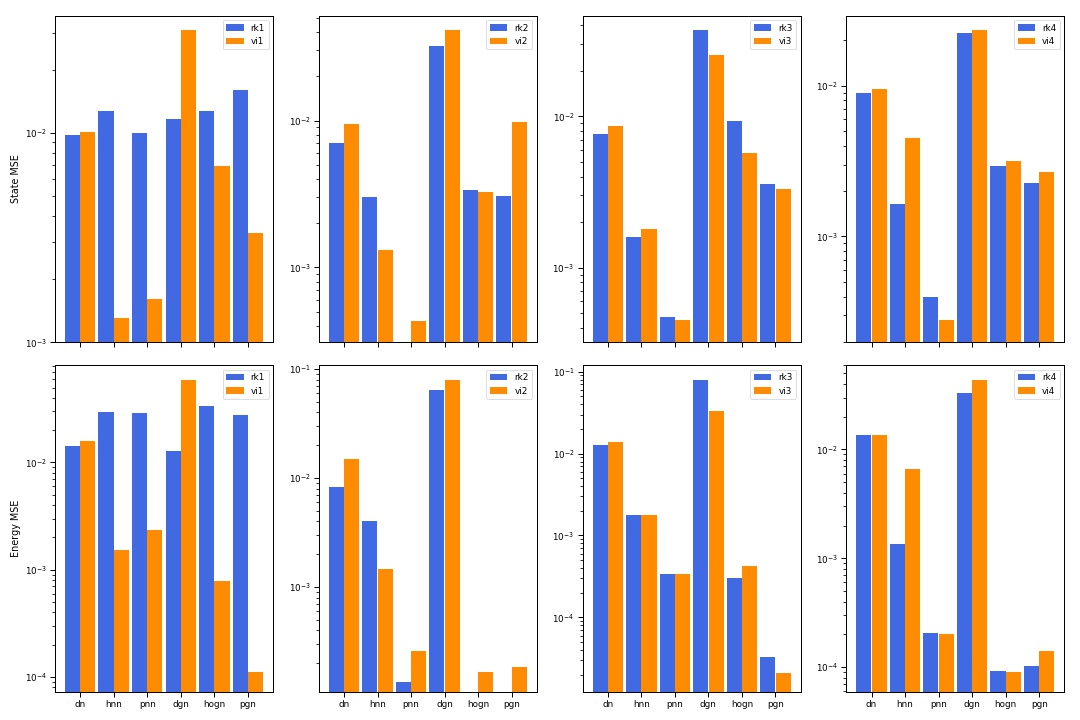}
        \caption{10-step integration}
    \end{subfigure}
\caption{3-Body gravitational system with noiseless training data. Each bar represents the geometric mean of the MSE of 25 test initial conditions.}
\end{figure*}

\begin{figure*}[htb]
\centering
	\begin{subfigure}[b]{0.3\textwidth}
        \includegraphics[width=\textwidth]{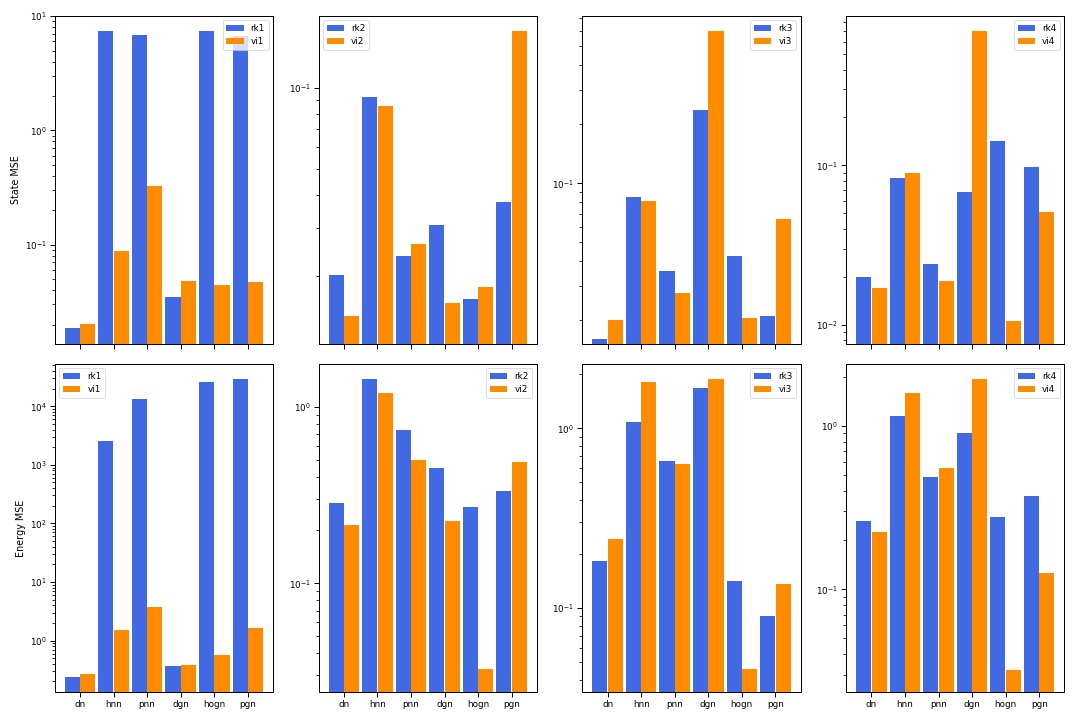}
        \caption{2-step integration}
    \end{subfigure}
	\begin{subfigure}[b]{0.3\textwidth}
        \includegraphics[width=\textwidth]{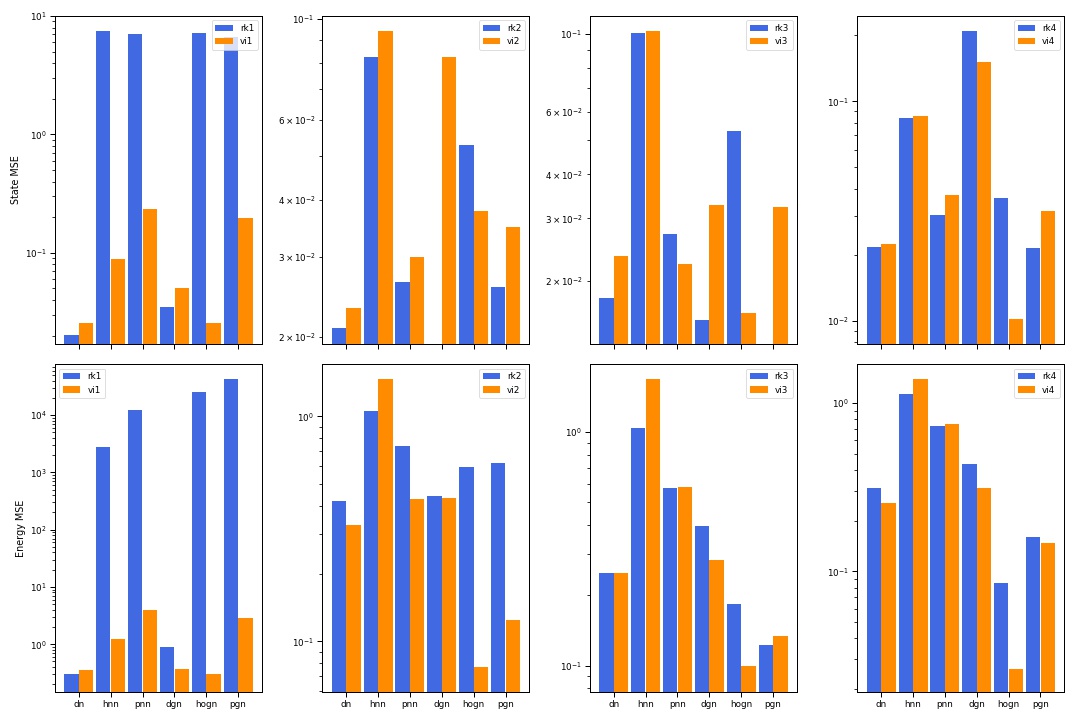}
        \caption{5-step integration}
    \end{subfigure}
	\begin{subfigure}[b]{0.3\textwidth}
        \includegraphics[width=\textwidth]{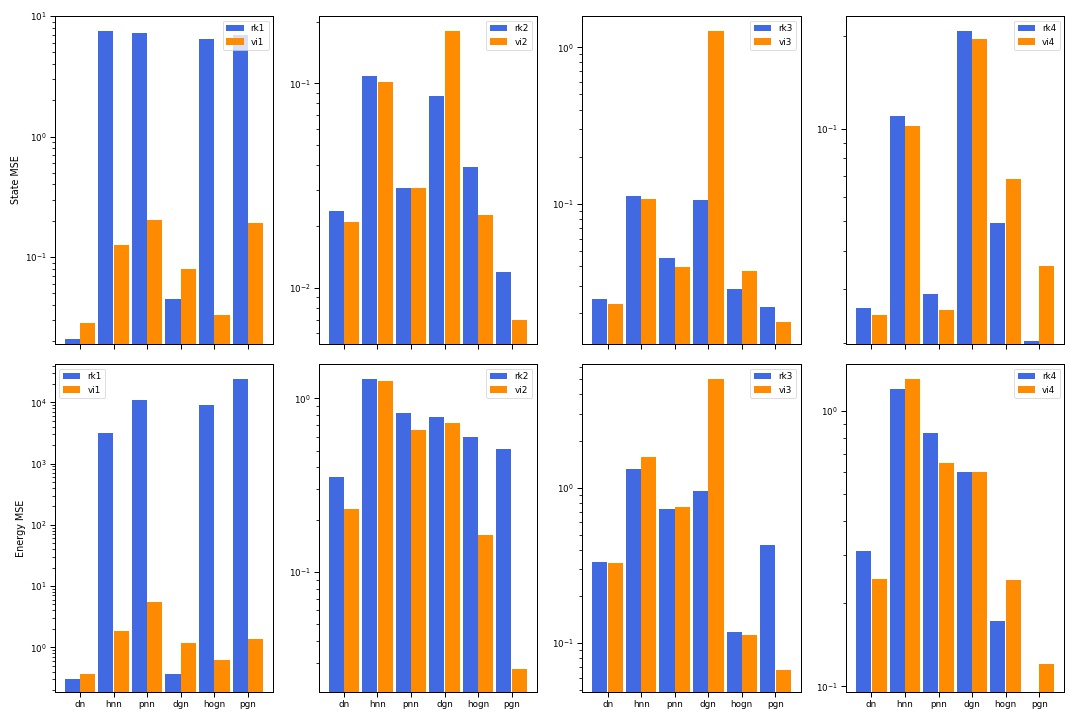}
        \caption{10-step integration}
    \end{subfigure}
\caption{5-spring particle system with noiseless training data.  Each bar represents the geometric mean of the MSE of 25 test initial conditions.}
\end{figure*}

\begin{figure*}[htb]
\centering
	\begin{subfigure}[b]{0.3\textwidth}
        \includegraphics[width=\textwidth]{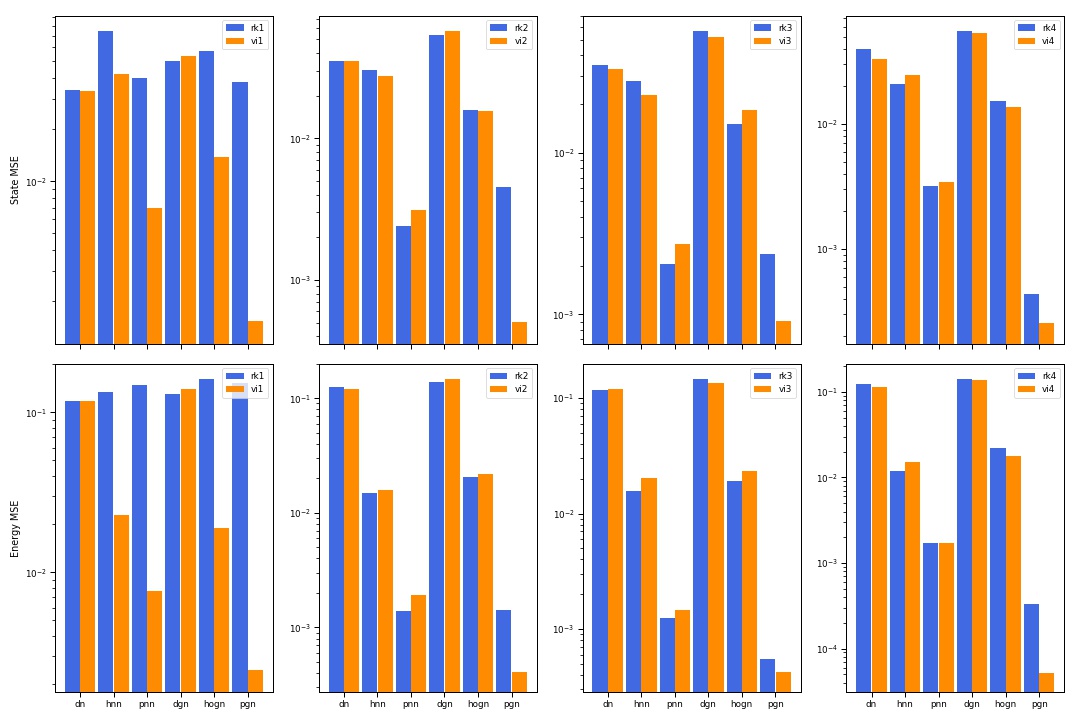}
        \caption{2-step integration}
    \end{subfigure}
	\begin{subfigure}[b]{0.3\textwidth}
        \includegraphics[width=\textwidth]{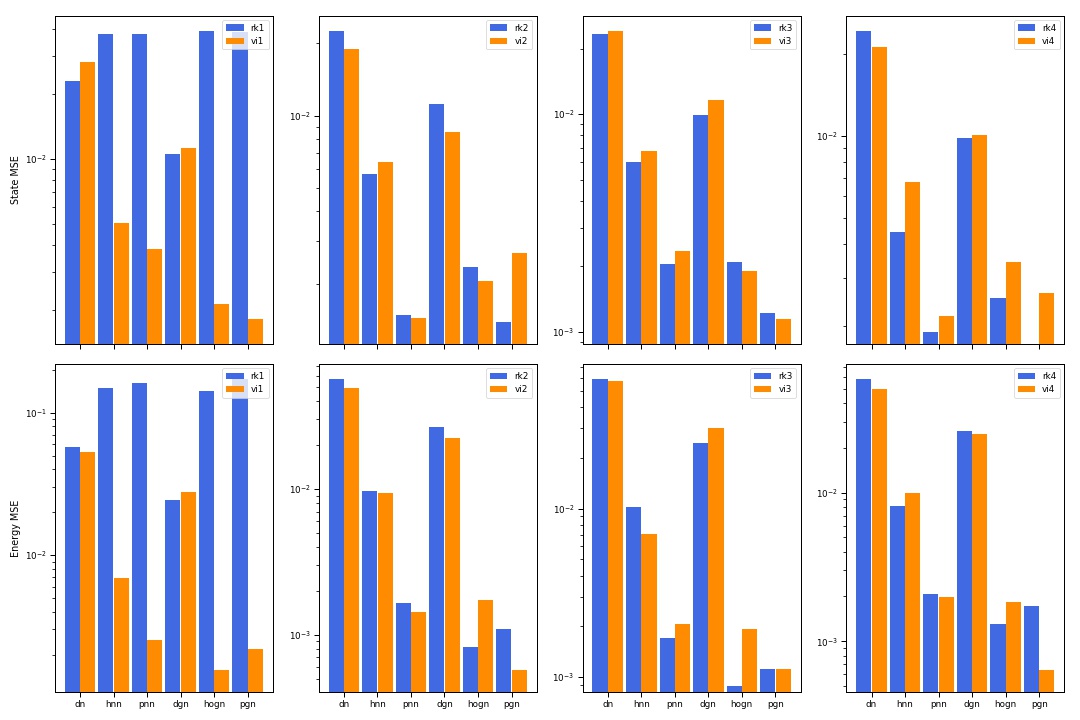}
        \caption{5-step integration}
    \end{subfigure}
	\begin{subfigure}[b]{0.3\textwidth}
        \includegraphics[width=\textwidth]{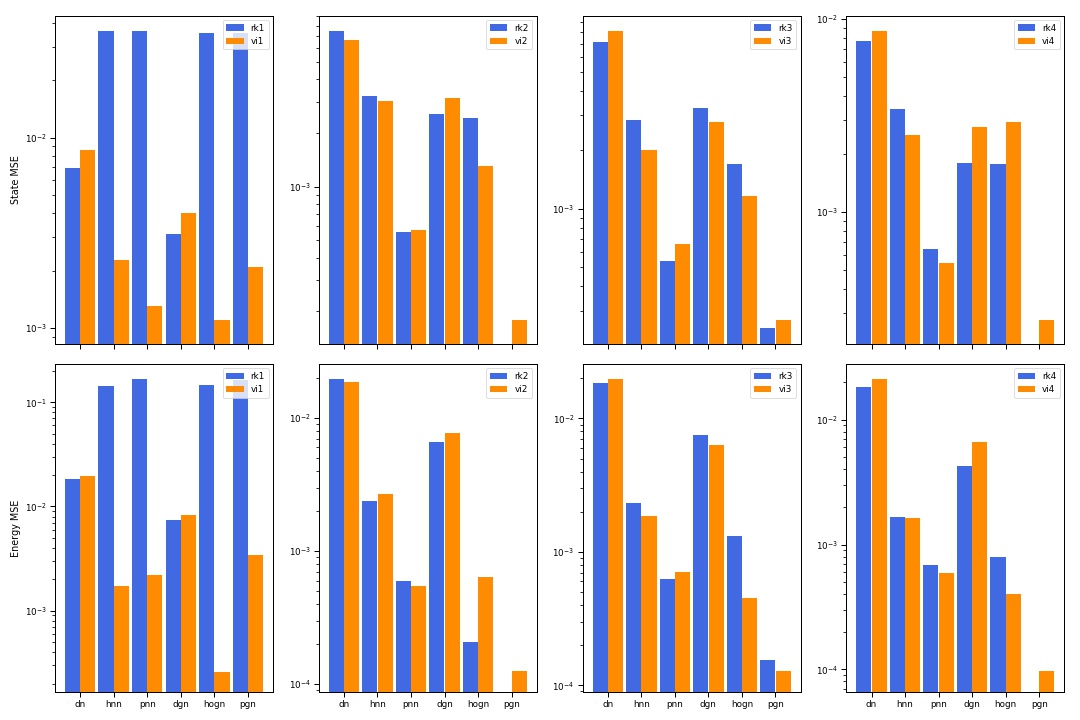}
        \caption{10-step integration}
    \end{subfigure}
\caption{Mass Spring System with noisy training data.  Each bar represents the geometric mean of the MSE of 25 test initial conditions.}
\end{figure*}
\begin{figure*}[htb]
\centering
	\begin{subfigure}[b]{0.3\textwidth}
        \includegraphics[width=\textwidth]{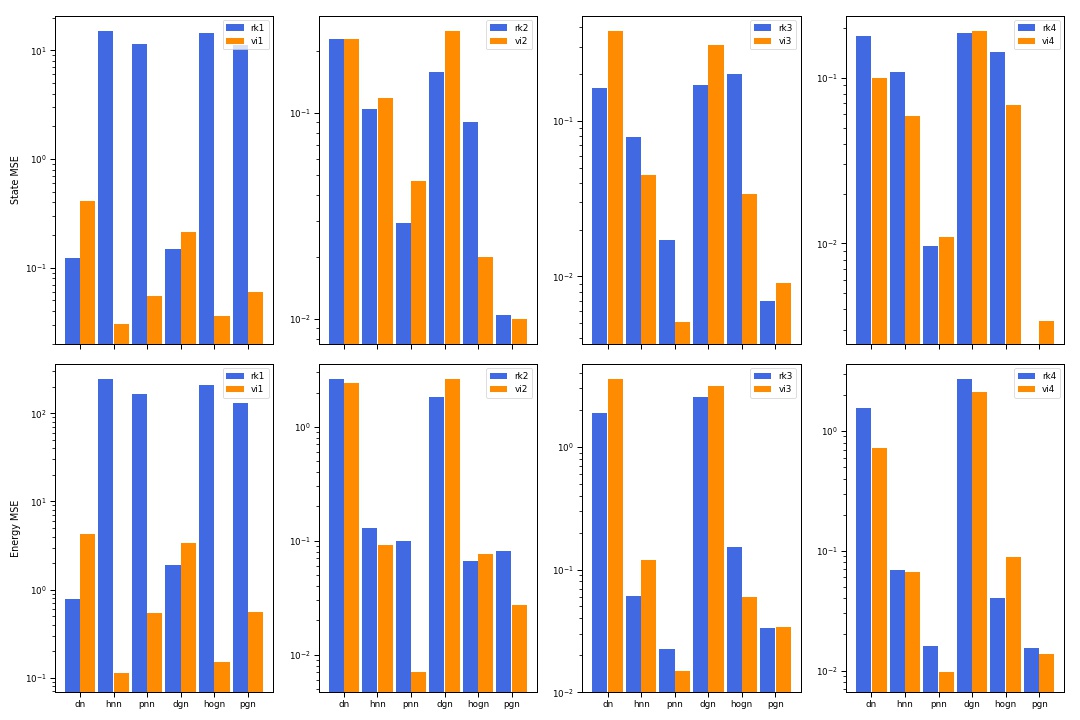}
        \caption{2-step integration}
    \end{subfigure}
	\begin{subfigure}[b]{0.3\textwidth}
        \includegraphics[width=\textwidth]{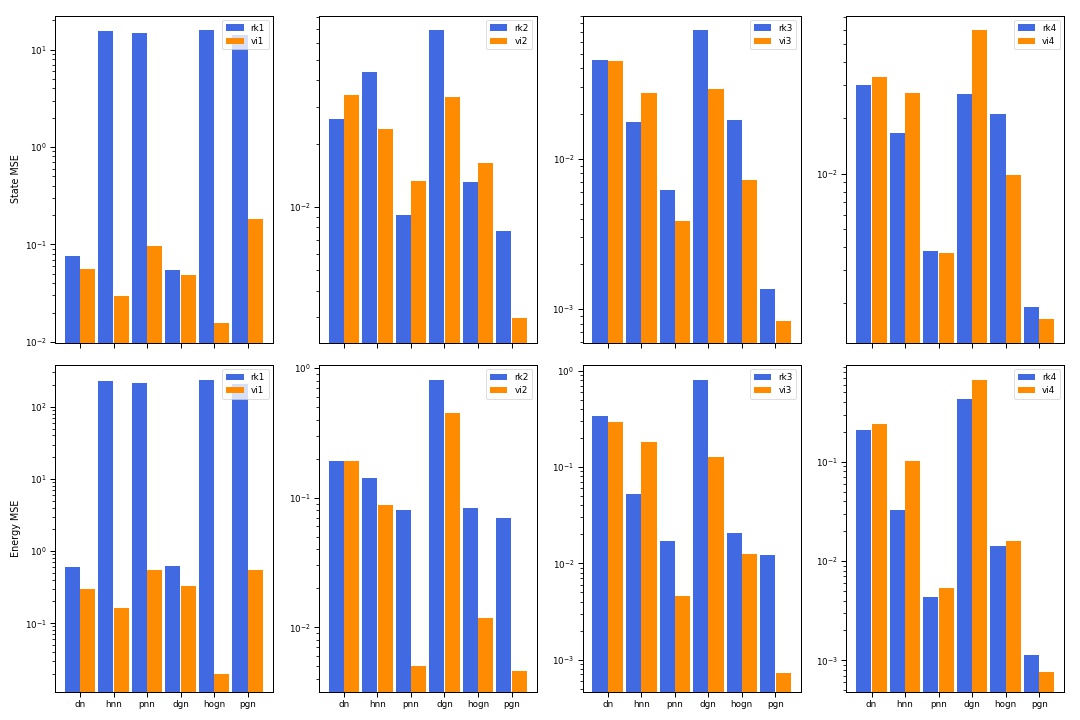}
        \caption{5-step integration}
    \end{subfigure}
	\begin{subfigure}[b]{0.3\textwidth}
        \includegraphics[width=\textwidth]{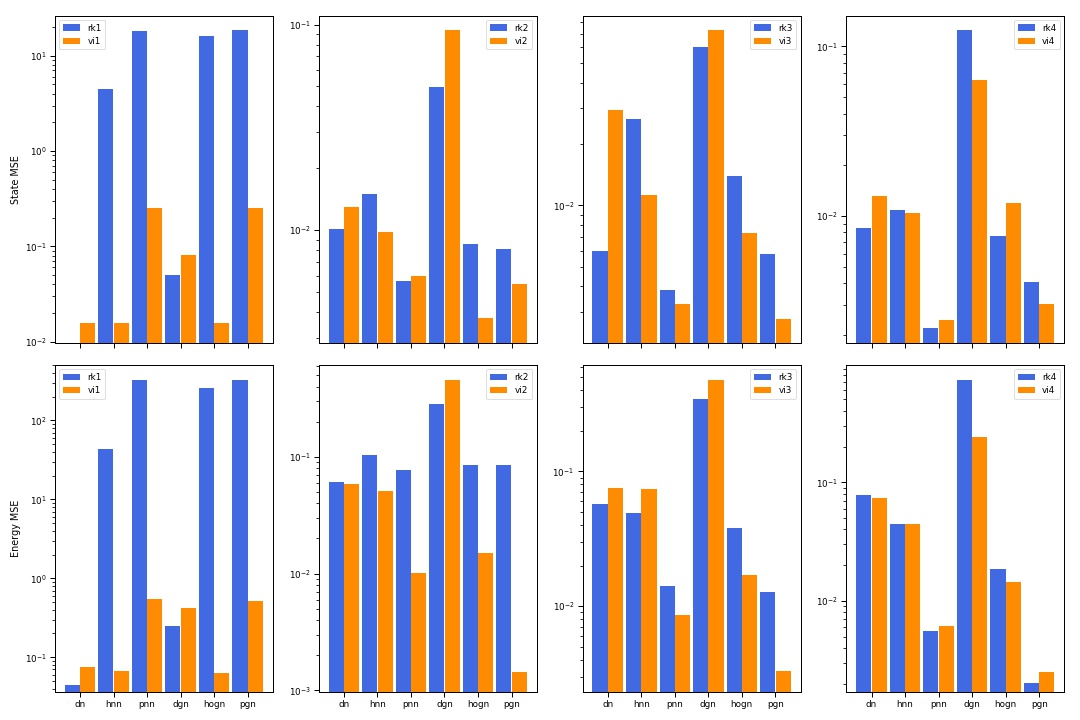}
        \caption{10-step integration}
    \end{subfigure}
\caption{Pendulum with noisy training data.  Each bar represents the geometric mean of the MSE of 25 test initial conditions.}
\end{figure*}
\begin{figure*}[htb]
\centering
	\begin{subfigure}[b]{0.3\textwidth}
        \includegraphics[width=\textwidth]{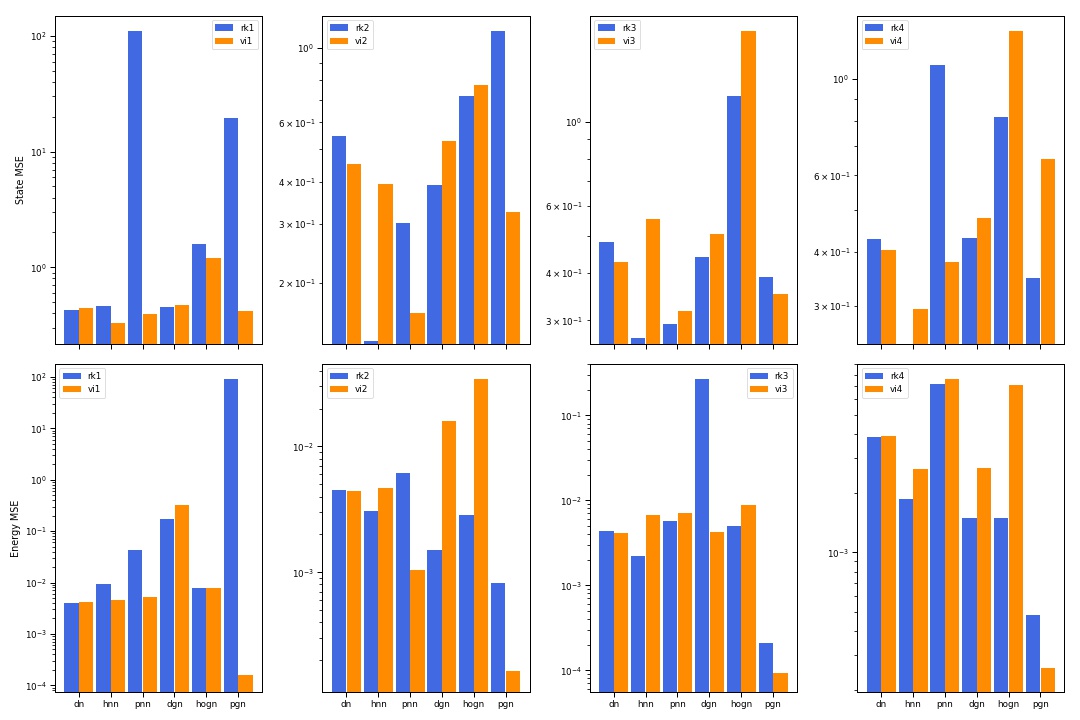}
        \caption{2-step integration}
    \end{subfigure}
	\begin{subfigure}[b]{0.3\textwidth}
        \includegraphics[width=\textwidth]{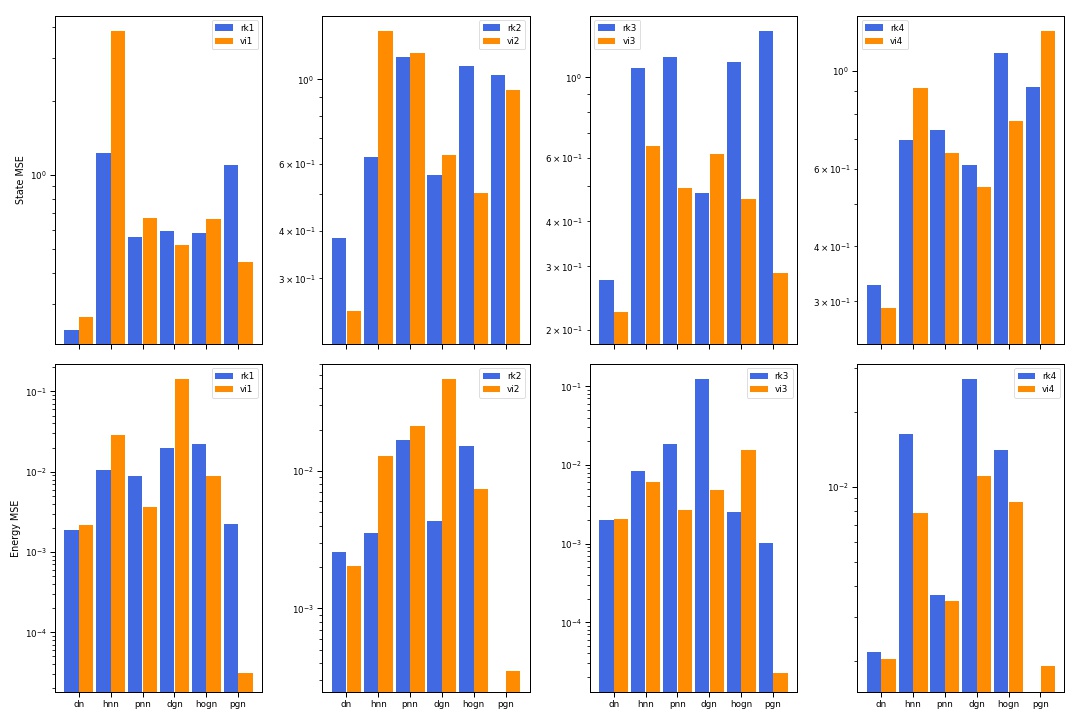}
        \caption{5-step integration}
    \end{subfigure}
	\begin{subfigure}[b]{0.3\textwidth}
        \includegraphics[width=\textwidth]{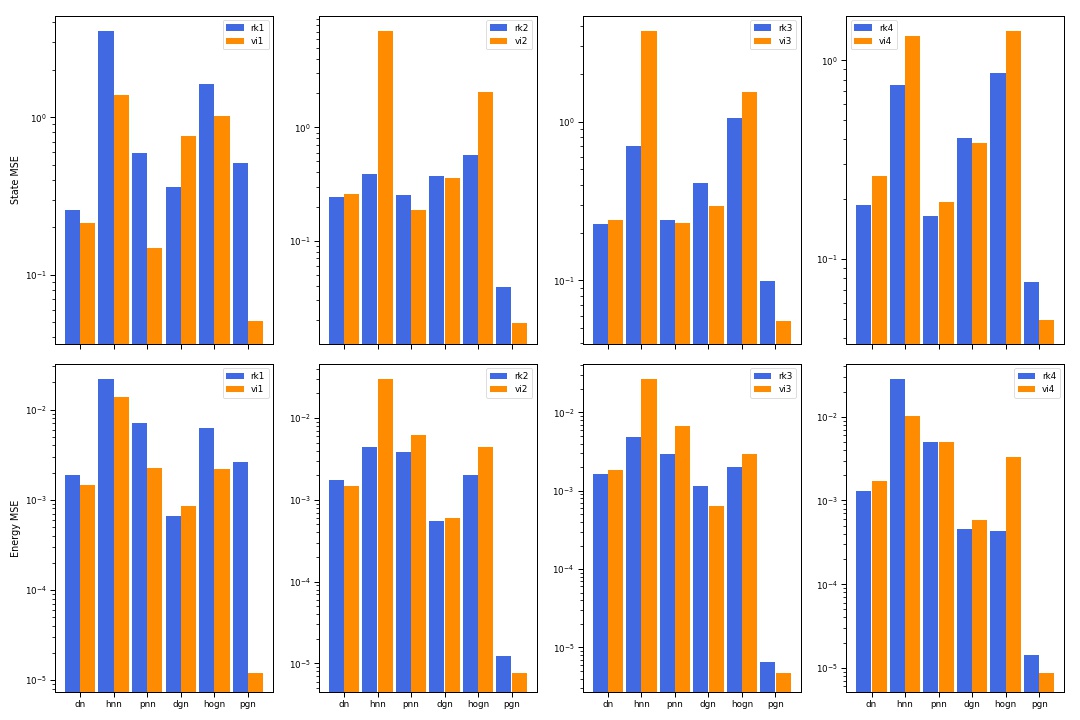}
        \caption{10-step integration}
    \end{subfigure}
\caption{2-Body gravitational system with noiseless training data.  Each bar represents the geometric mean of the MSE of 25 test initial conditions.}
\end{figure*}
\begin{figure*}[htb]
\centering
	\begin{subfigure}[b]{0.3\textwidth}
        \includegraphics[width=\textwidth]{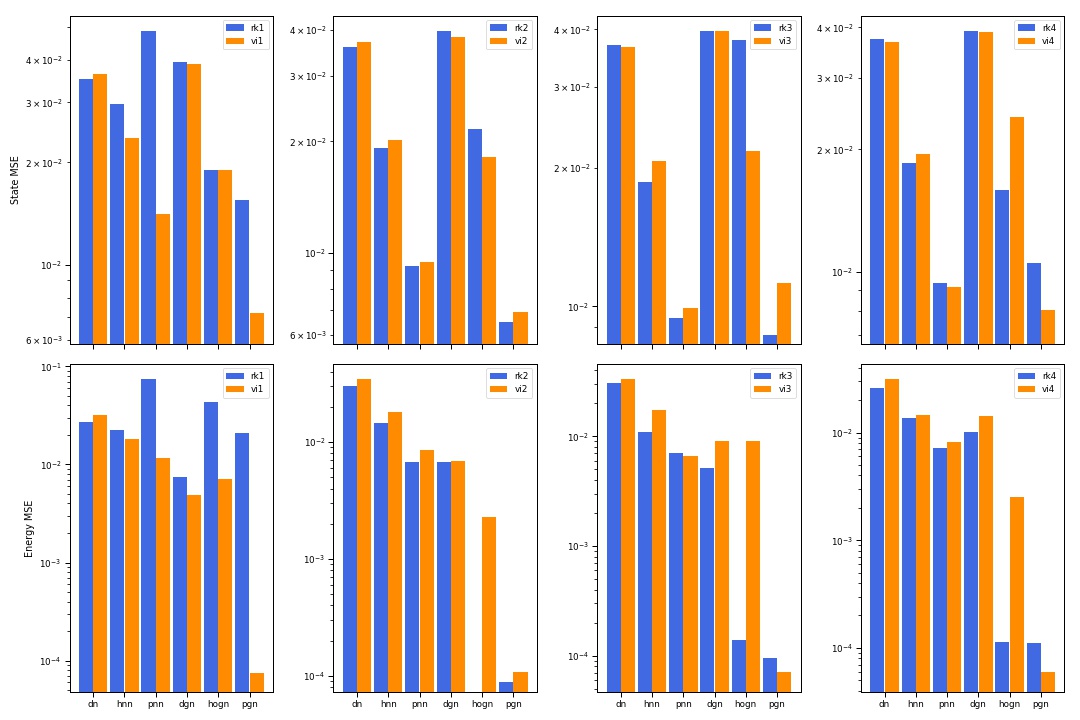}
        \caption{2-step integration}
    \end{subfigure}
	\begin{subfigure}[b]{0.3\textwidth}
        \includegraphics[width=\textwidth]{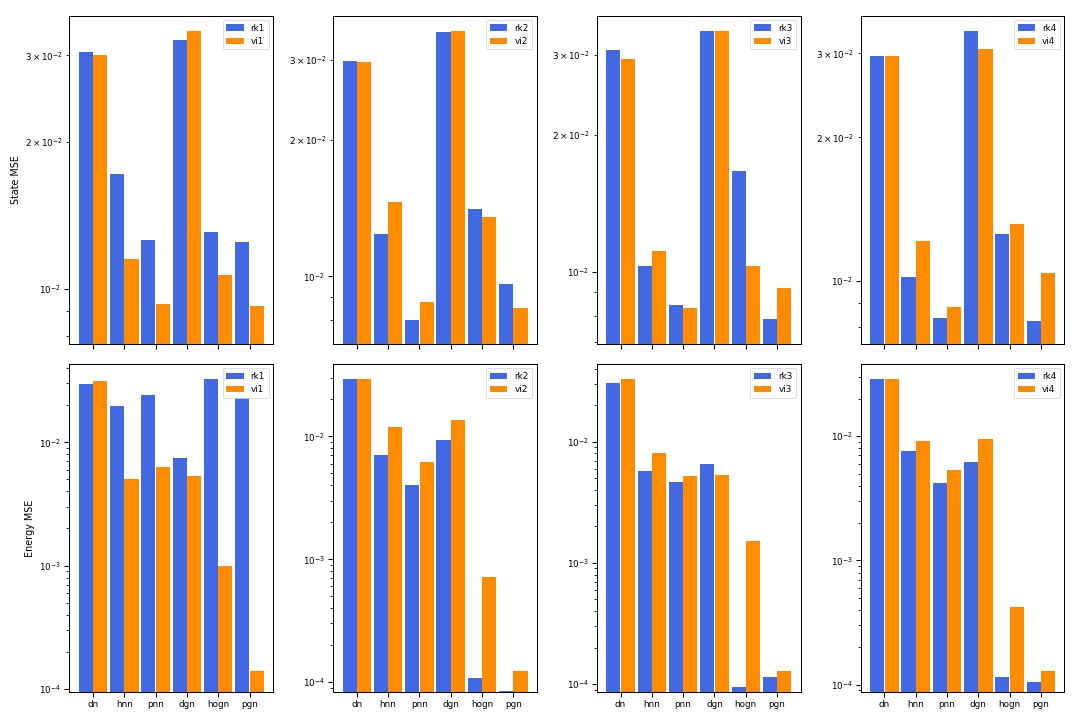}
        \caption{5-step integration}
    \end{subfigure}
	\begin{subfigure}[b]{0.3\textwidth}
        \includegraphics[width=\textwidth]{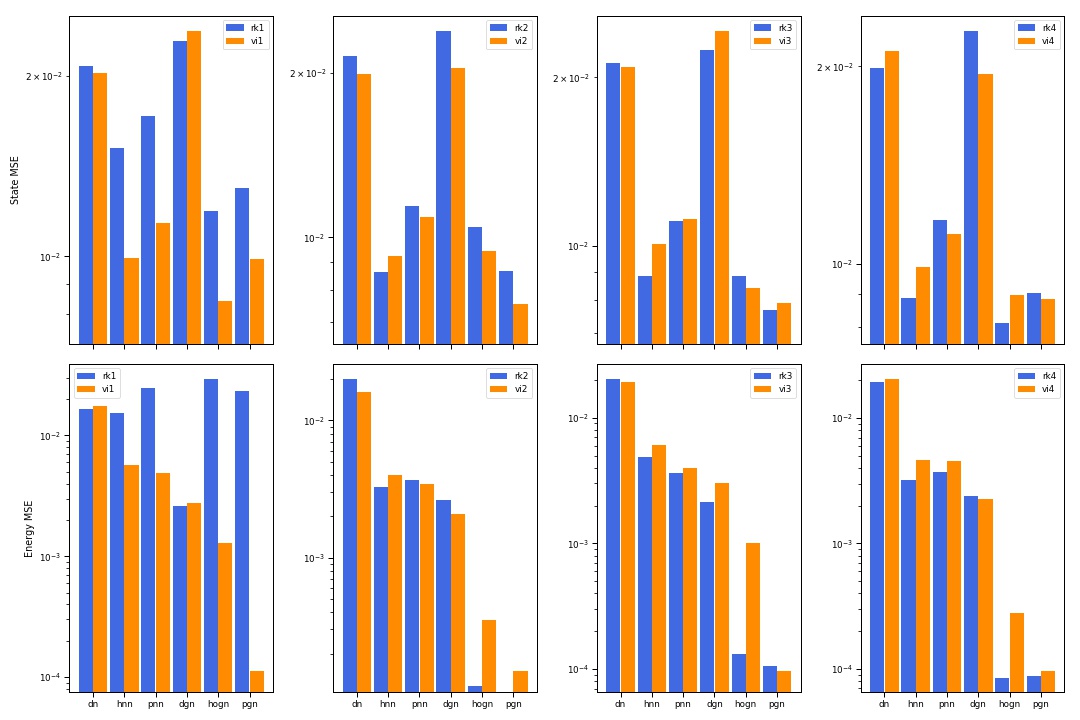}
        \caption{10-step integration}
    \end{subfigure}
\caption{3-Body gravitational system with noisy training data.  Each bar represents the geometric mean of the MSE of 25 test initial conditions.}
\end{figure*}

\begin{figure*}[htb]
\centering
	\begin{subfigure}[b]{0.3\textwidth}
        \includegraphics[width=\textwidth]{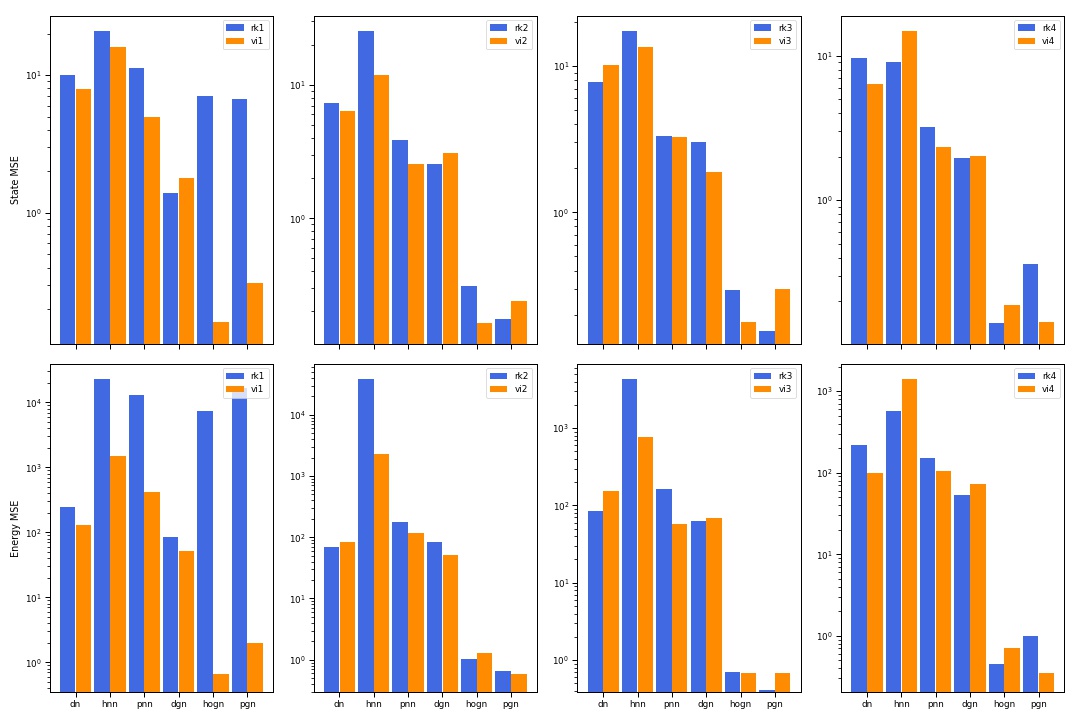}
        \caption{2-step integration}
    \end{subfigure}
	\begin{subfigure}[b]{0.3\textwidth}
        \includegraphics[width=\textwidth]{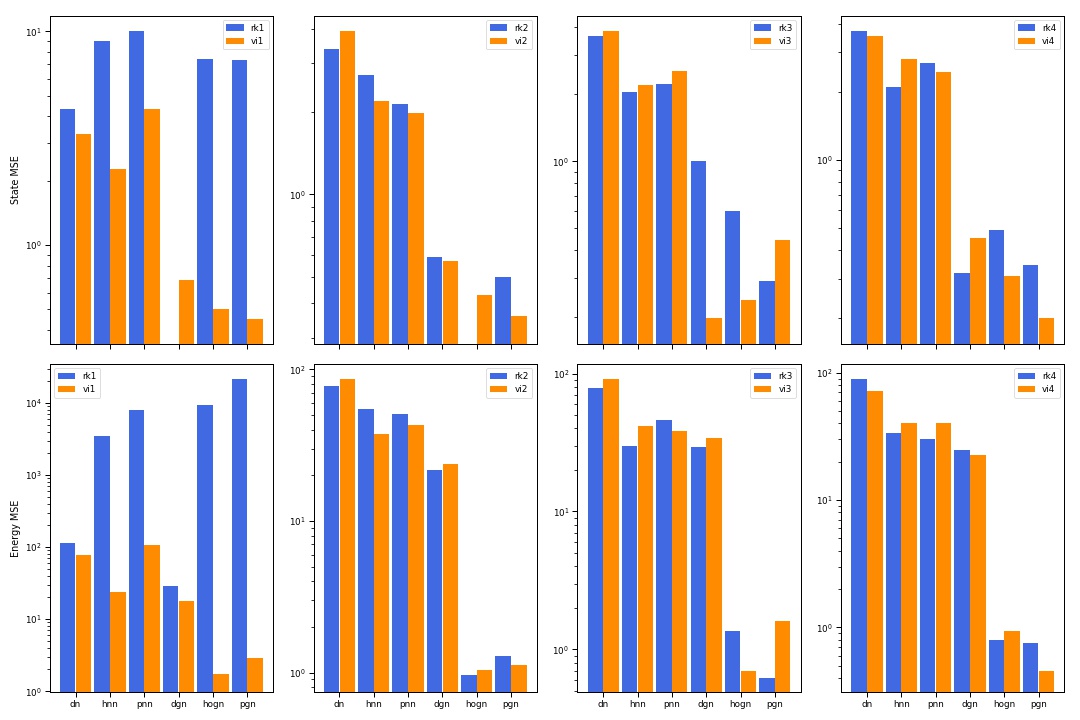}
        \caption{5-step integration}
    \end{subfigure}
	\begin{subfigure}[b]{0.3\textwidth}
        \includegraphics[width=\textwidth]{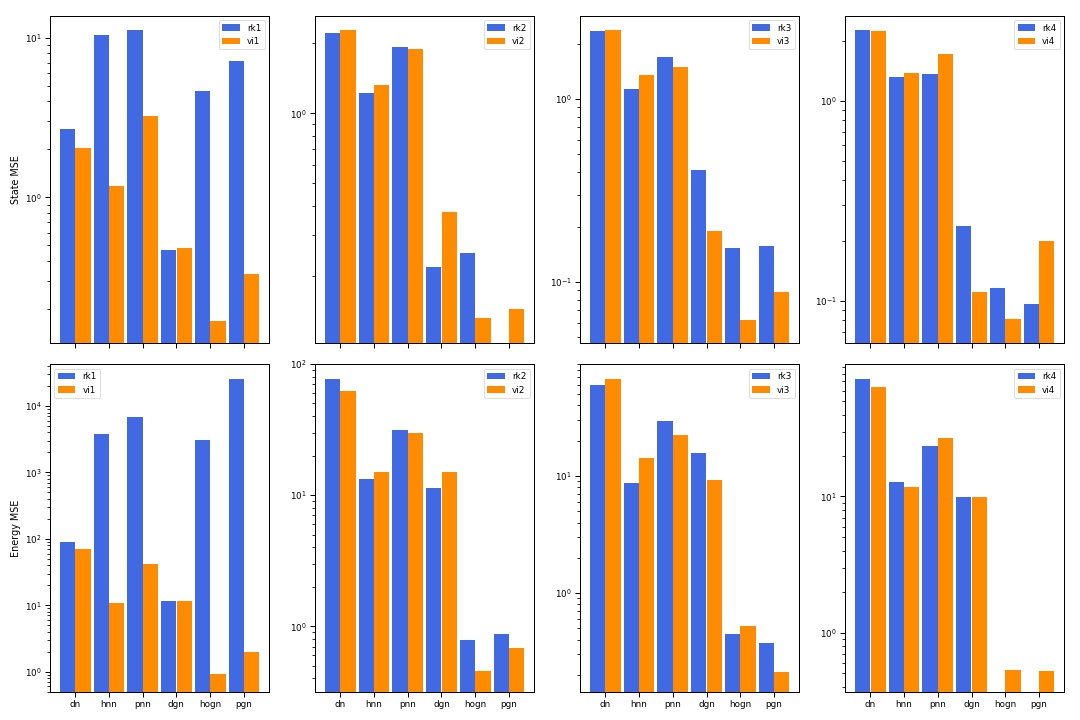}
        \caption{10-step integration}
    \end{subfigure}
\caption{5-spring particle system with noisy training data.  Each bar represents the geometric mean of the MSE of 25 test initial conditions.}
\end{figure*}

\newpage
\pagebreak

\section{Rollout Errors}

\begin{figure*}[htb]
\begin{subfigure}[b]{\textwidth}
\includegraphics[width=\textwidth]{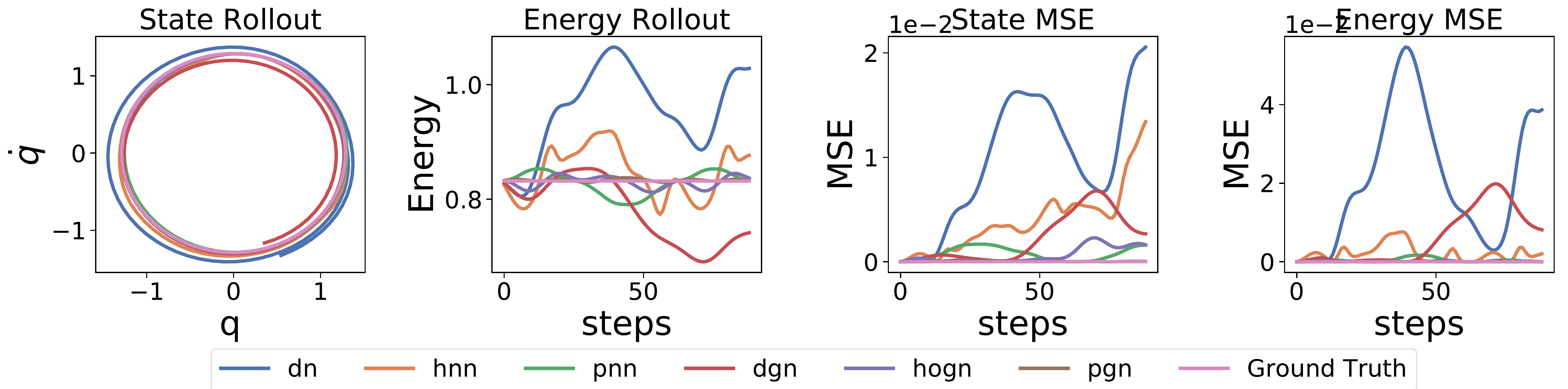}
\caption{RK4 rollout}
\end{subfigure}
\begin{subfigure}[b]{\textwidth}
\includegraphics[width=\textwidth]{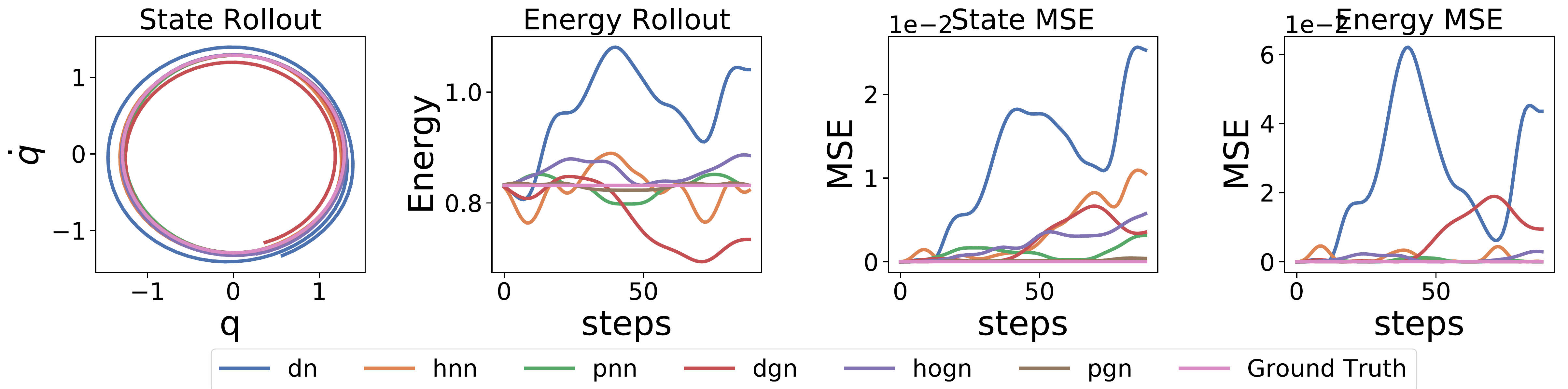}
\caption{VI4 rollout}
\end{subfigure}
\caption{Rollout of mass-spring system of a single point in the test set. The methods are pretrained with noisy data.}
\end{figure*}

\begin{figure*}[htb]
\begin{subfigure}[b]{\textwidth}
\includegraphics[width=\textwidth]{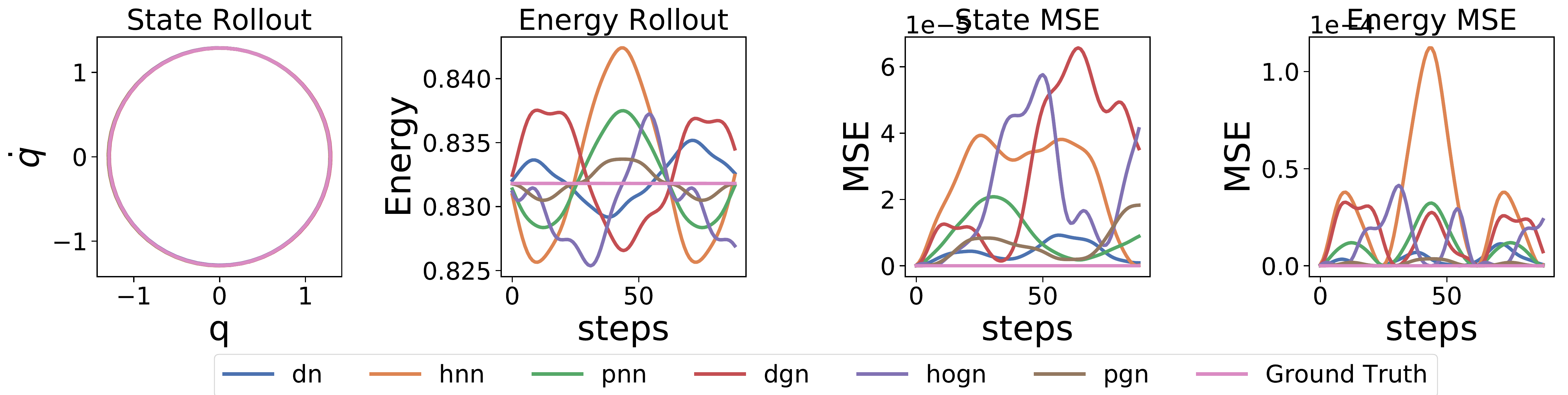}
\caption{RK4 rollout}
\end{subfigure}
\begin{subfigure}[b]{\textwidth}
\includegraphics[width=\textwidth]{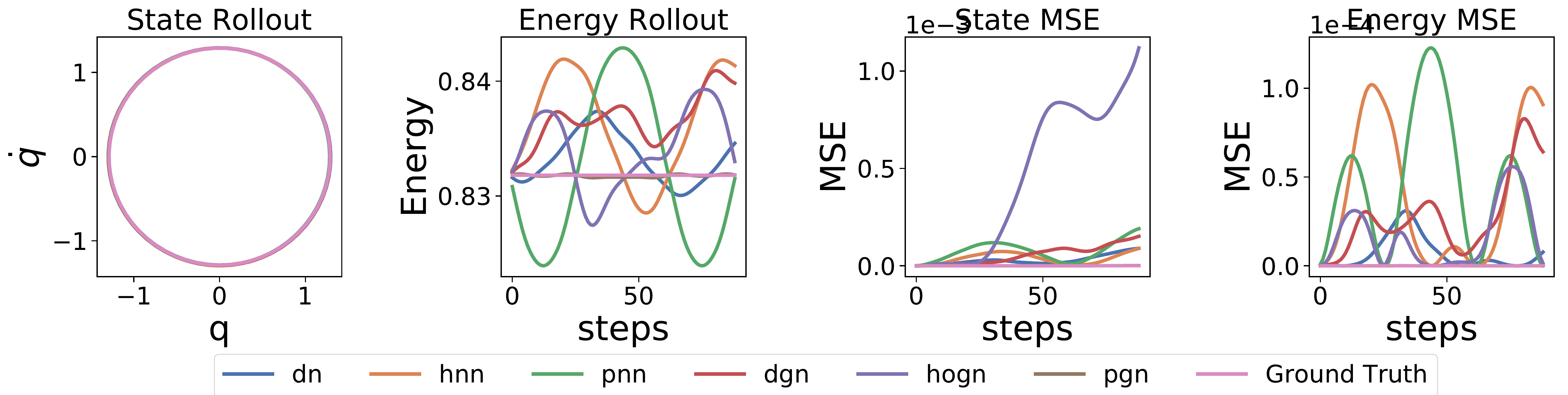}
\caption{VI4 rollout}
\end{subfigure}
\caption{Rollout of mass-spring system of a single point in the test set. The methods are pretrained with noiseless data.}
\end{figure*}


\begin{figure*}[htb]
\begin{subfigure}[b]{\textwidth}
\includegraphics[width=\textwidth]{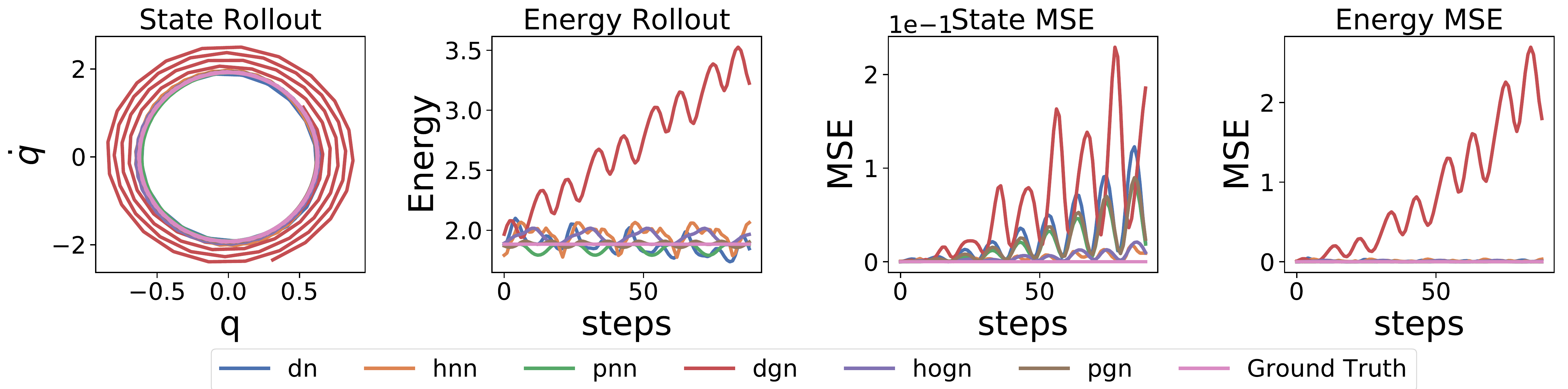}
\caption{RK4 rollout}
\end{subfigure}
\begin{subfigure}[b]{\textwidth}
\includegraphics[width=\textwidth]{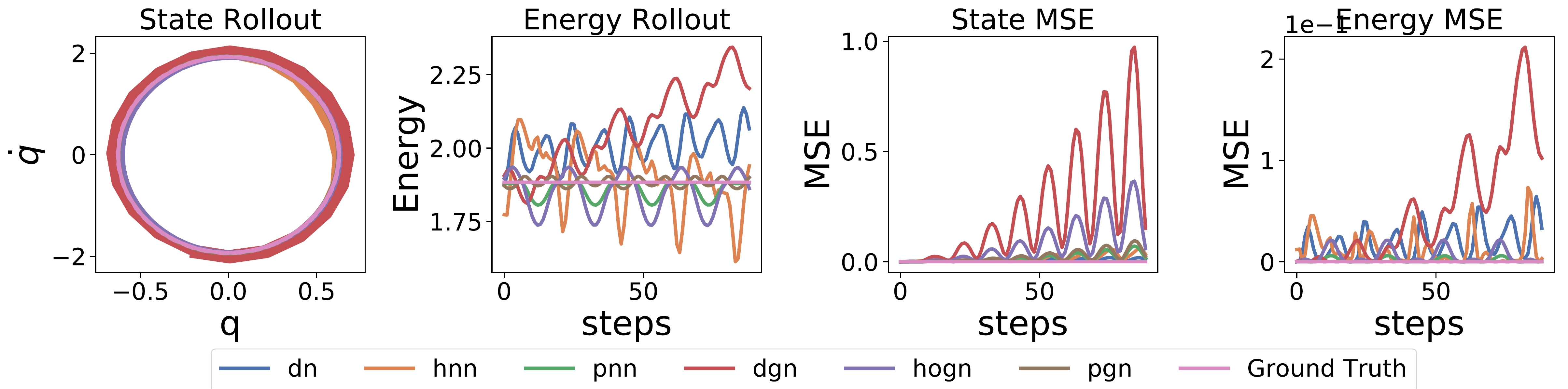}
\caption{VI4 rollout}
\end{subfigure}
\caption{Rollout of pendulum system of a single point in the test set. The methods are pretrained with noisy data.}
\end{figure*}

\begin{figure*}[htb]
\begin{subfigure}[b]{\textwidth}
\includegraphics[width=\textwidth]{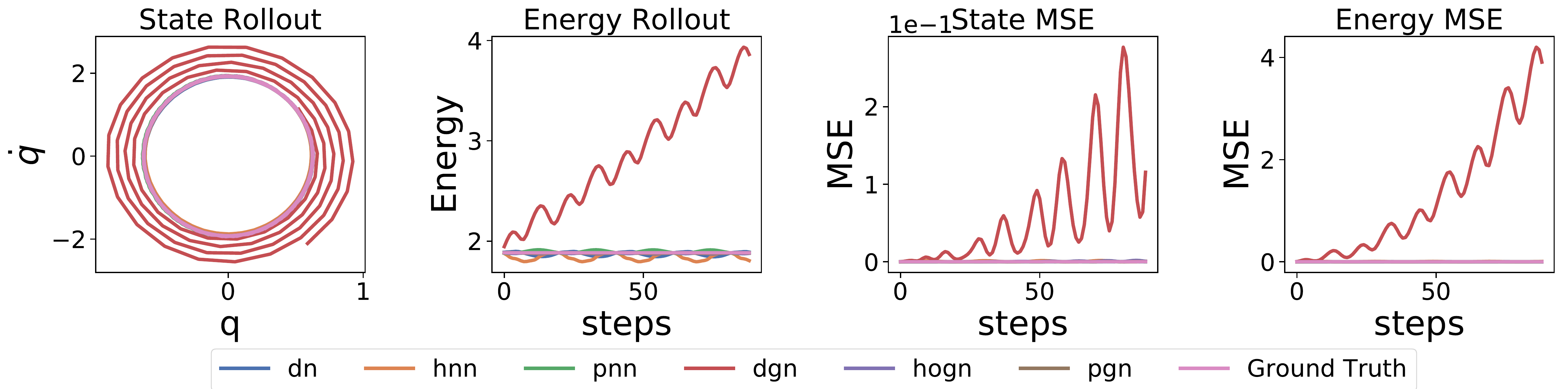}
\caption{RK4 rollout}
\end{subfigure}
\begin{subfigure}[b]{\textwidth}
\includegraphics[width=\textwidth]{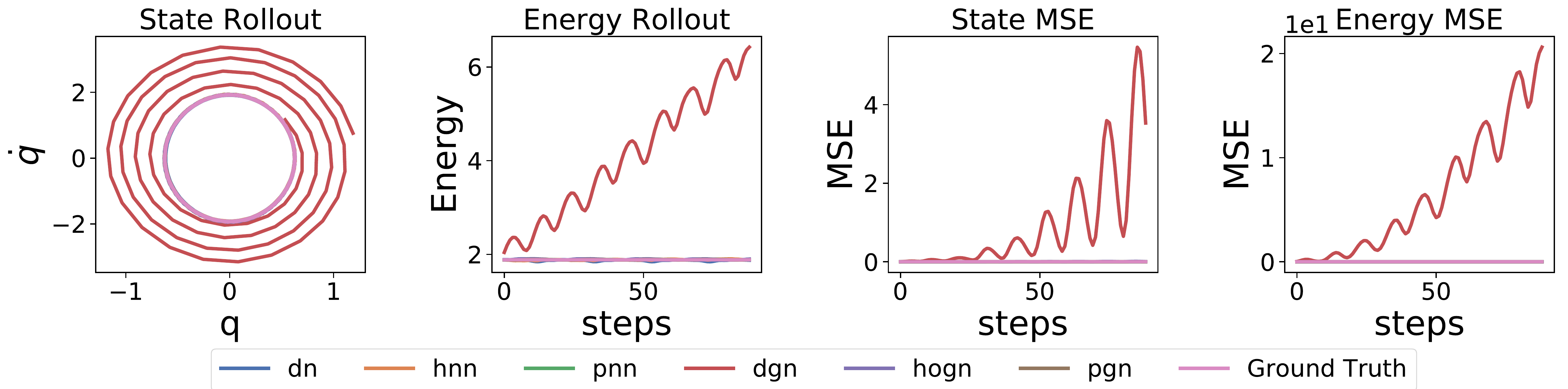}
\caption{VI4 rollout}
\end{subfigure}
\caption{Rollout of pendulum system of a single point in the test set. The methods are pretrained with noiseless data.}
\end{figure*}


\begin{figure*}[htb]
\begin{subfigure}[b]{\textwidth}
\includegraphics[width=\textwidth]{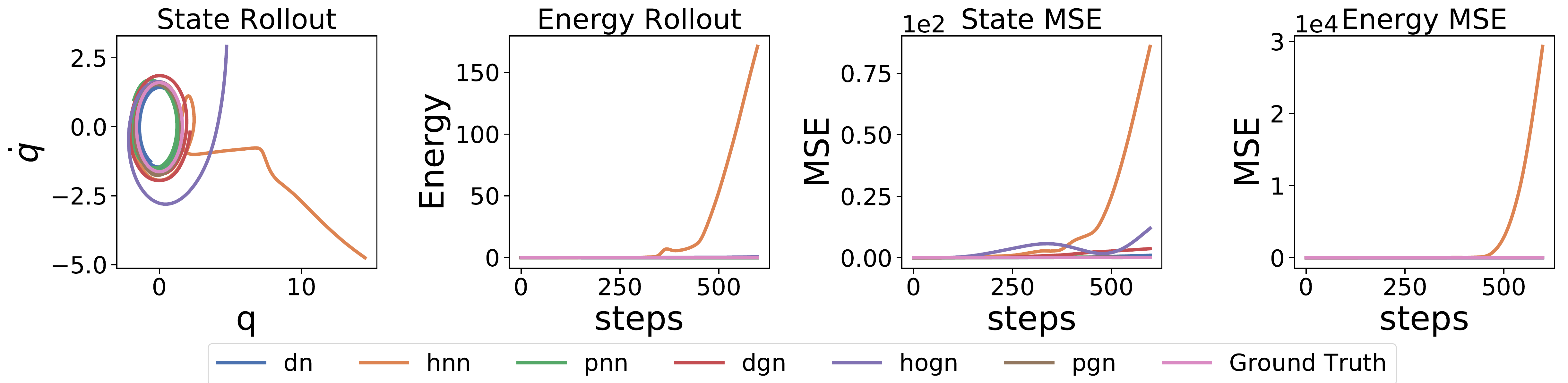}
\caption{RK4 rollout}
\end{subfigure}
\begin{subfigure}[b]{\textwidth}
\includegraphics[width=\textwidth]{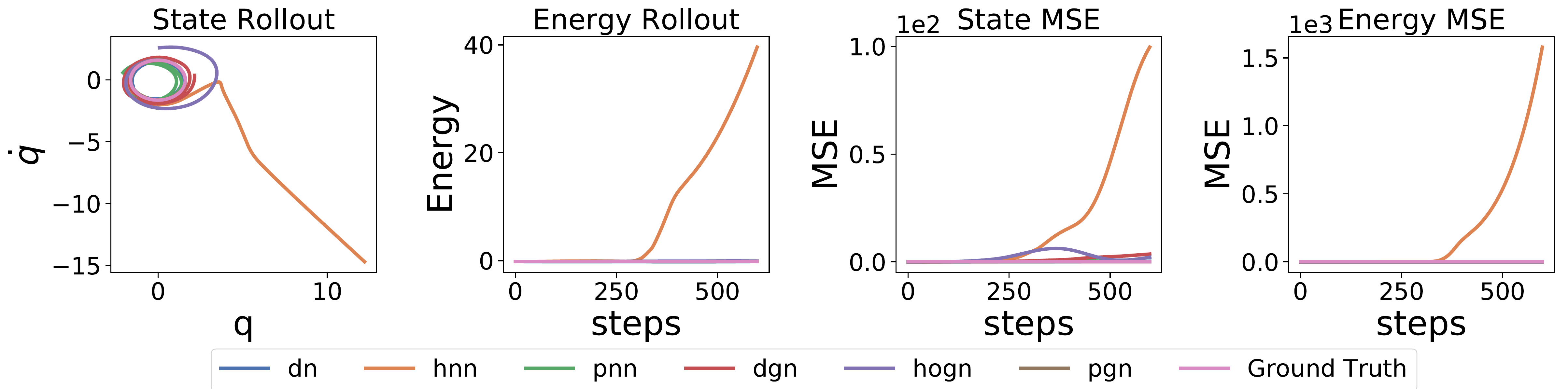}
\caption{VI4 rollout}
\end{subfigure}
\caption{Rollout of 2-Body gravitational system of a single point in the test set. The methods are pretrained with noisy data.}
\end{figure*}

\begin{figure*}[htb]
\begin{subfigure}[b]{\textwidth}
\includegraphics[width=\textwidth]{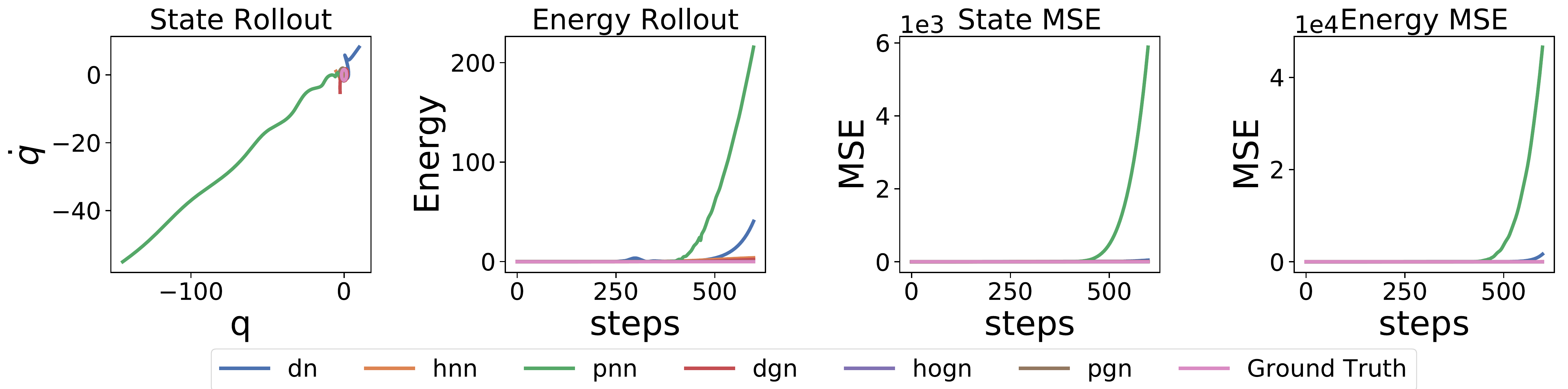}
\caption{RK4 rollout}
\end{subfigure}
\begin{subfigure}[b]{\textwidth}
\includegraphics[width=\textwidth]{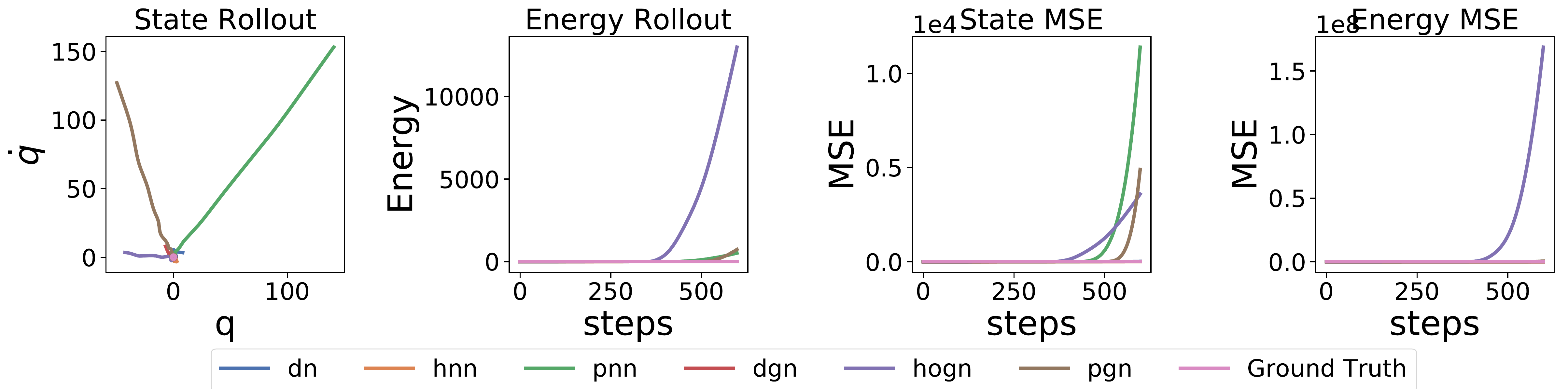}
\caption{VI4 rollout}
\end{subfigure}
\caption{Rollout of 2-Body gravitational system of a single point in the test set. The methods are pretrained with noiseless data.}
\end{figure*}


\begin{figure*}[htb]
\begin{subfigure}[b]{\textwidth}
\includegraphics[width=\textwidth]{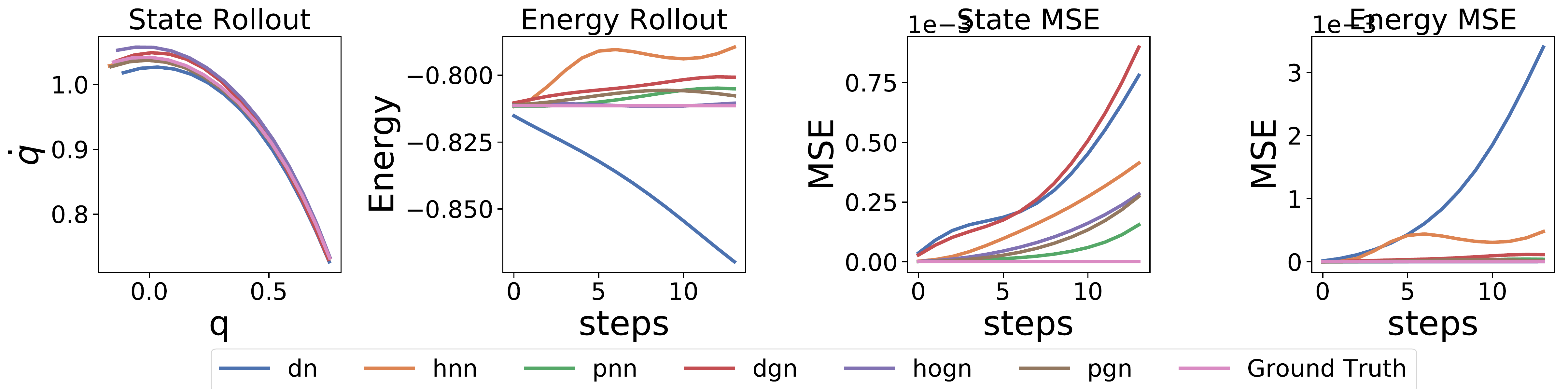}
\caption{RK4 rollout}
\end{subfigure}
\begin{subfigure}[b]{\textwidth}
\includegraphics[width=\textwidth]{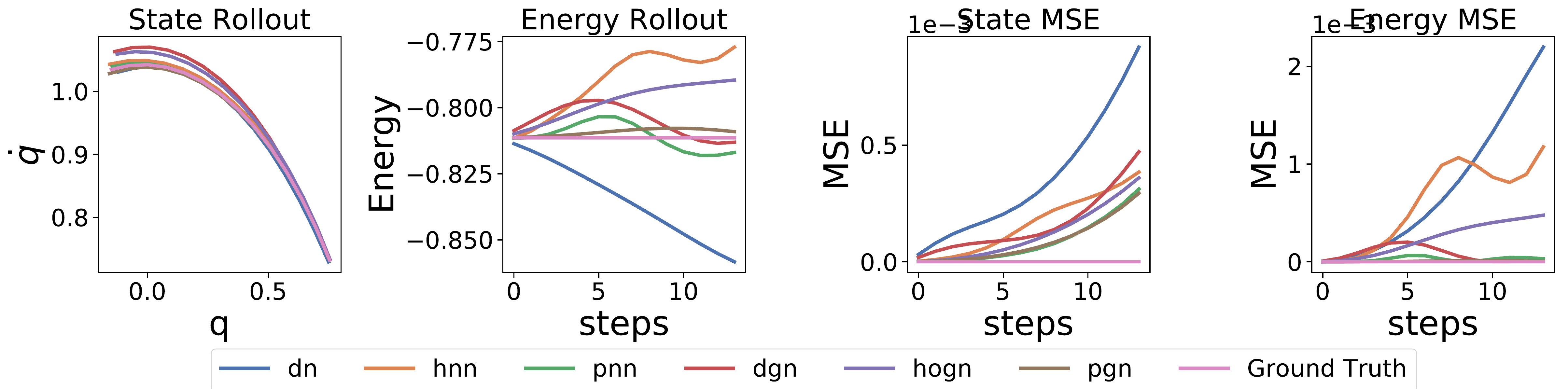}
\caption{VI4 rollout}
\end{subfigure}
\caption{Rollout of three body gravitational system of a single point in the test set. The methods are pretrained with noisy data.}
\end{figure*}

\begin{figure*}[htb]
\begin{subfigure}[b]{\textwidth}
\includegraphics[width=\textwidth]{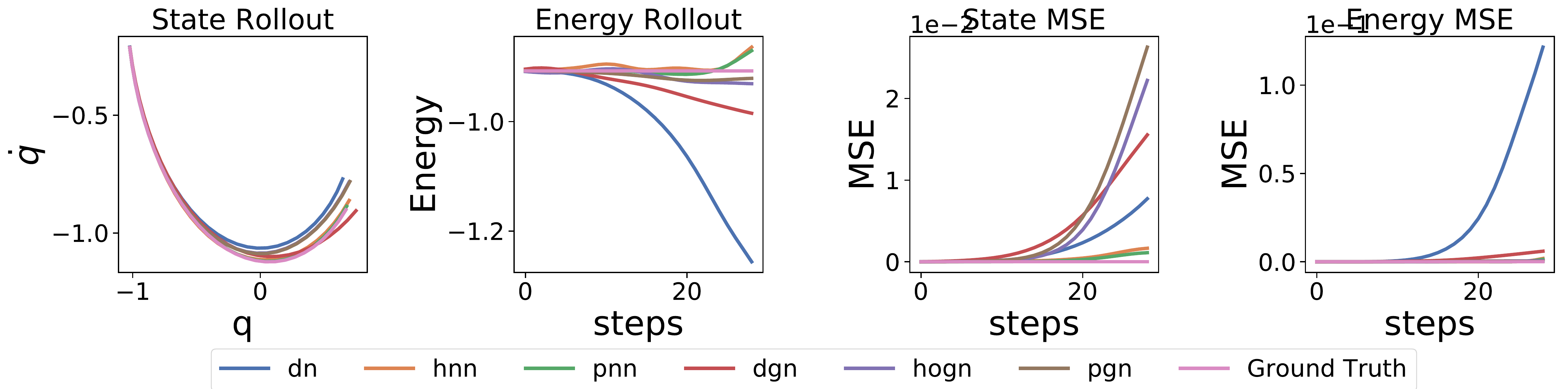}
\caption{RK4 rollout}
\end{subfigure}
\begin{subfigure}[b]{\textwidth}
\includegraphics[width=\textwidth]{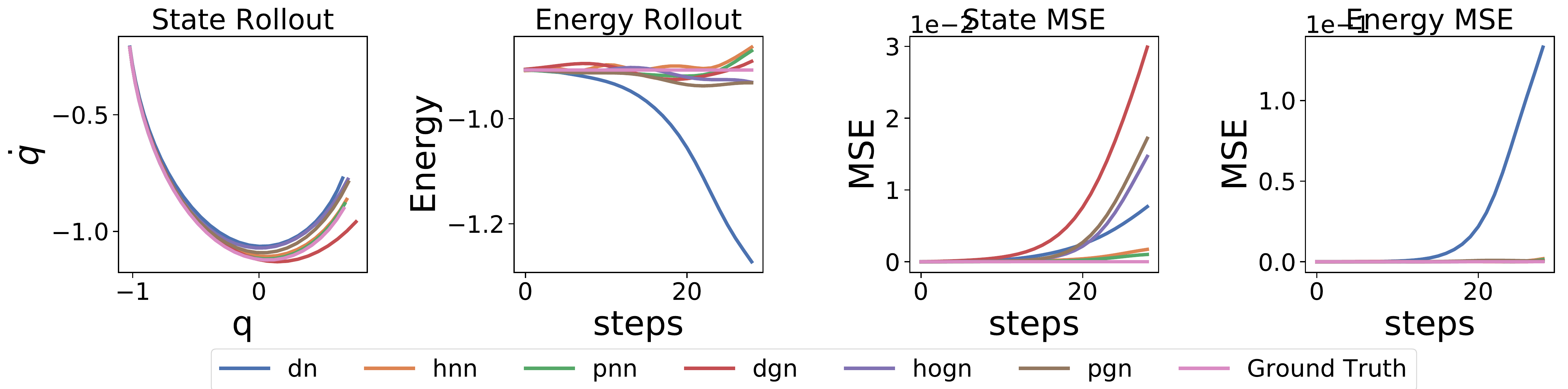}
\caption{VI4 rollout}
\end{subfigure}
\caption{Rollout of three body gravitational system of a single point in the test set. The methods are pretrained with noiseless data.}
\end{figure*}


\begin{figure*}[htb]
\begin{subfigure}[b]{\textwidth}
\includegraphics[width=\textwidth]{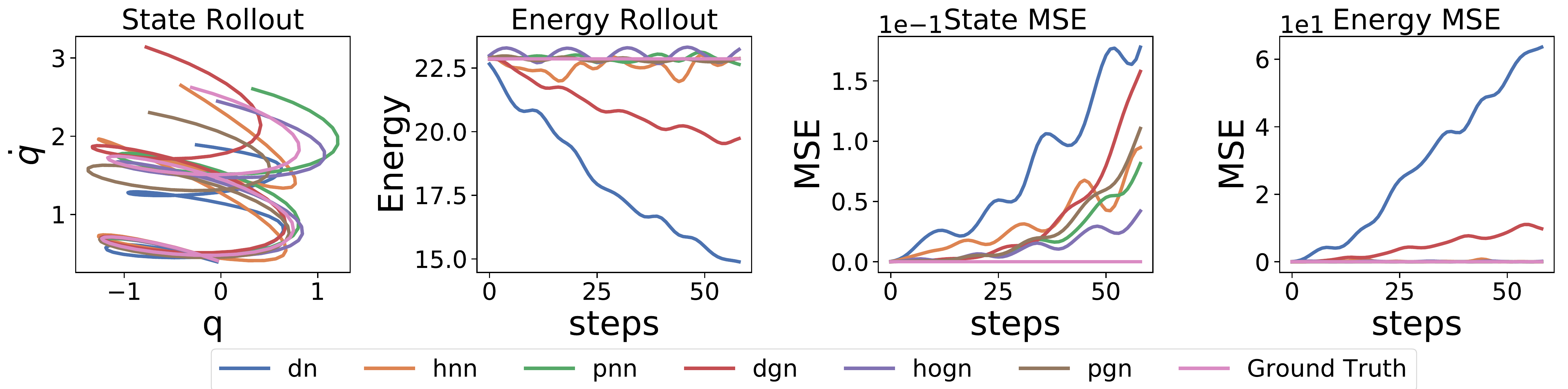}
\caption{RK4 rollout}
\end{subfigure}
\begin{subfigure}[b]{\textwidth}
\includegraphics[width=\textwidth]{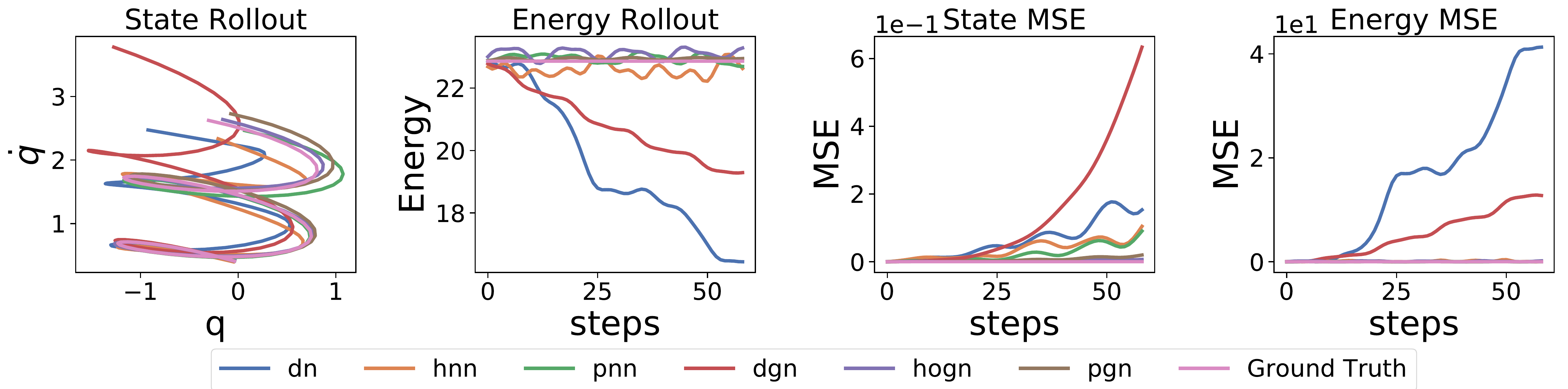}
\caption{VI4 rollout}
\end{subfigure}
\caption{Rollout of 5 body particle spring system of a single point in the test set. The methods are pretrained with noisy data.}
\end{figure*}

\begin{figure*}[htb]
\begin{subfigure}[b]{\textwidth}
\includegraphics[width=\textwidth]{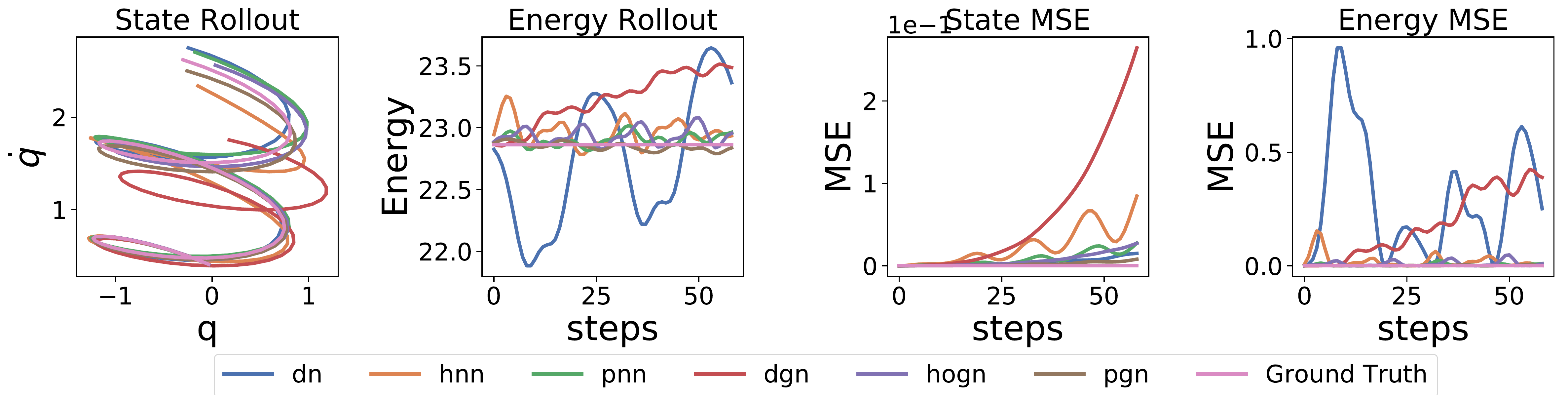}
\caption{RK4 rollout}
\end{subfigure}
\begin{subfigure}[b]{\textwidth}
\includegraphics[width=\textwidth]{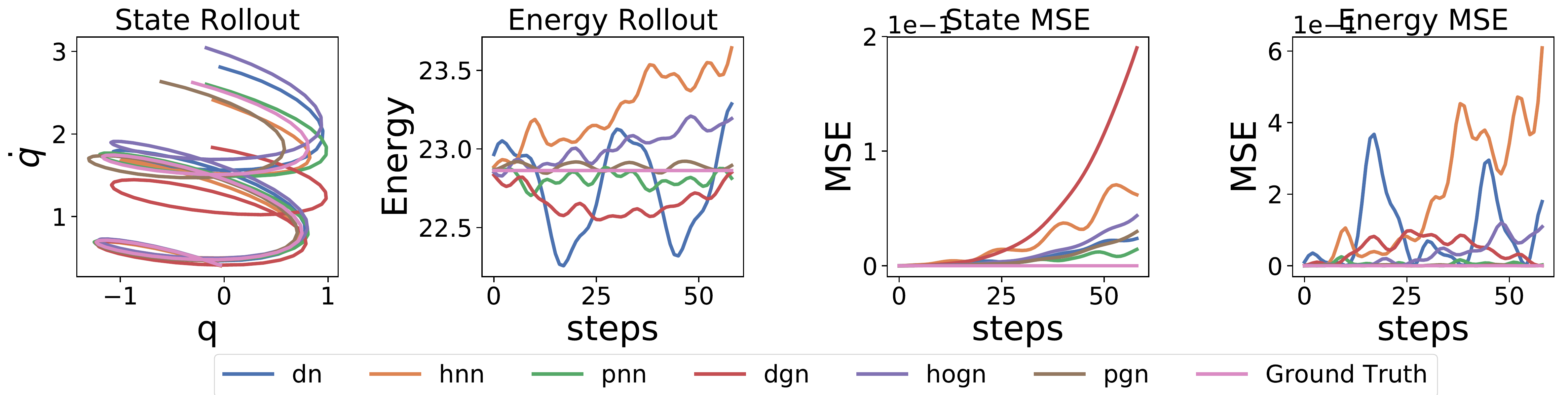}
\caption{VI4 rollout}
\end{subfigure}
\caption{Rollout of 5 body particle spring system of a single point in the test set. The methods are pretrained with noiseless data.}
\end{figure*}